\documentclass{article}

\usepackage{hyperref}

\usepackage[accepted]{icml2023}

\usepackage[utf8]{inputenc} 
\usepackage[T1]{fontenc}    
\usepackage{xr-hyper} 
\usepackage{url}            
\usepackage{booktabs}       
\usepackage{amsfonts}       
\usepackage{nicefrac}       
\usepackage{microtype}      

\usepackage{float}
\usepackage{amsmath}
\usepackage{amsthm}
\usepackage{dsfont}
\usepackage{amssymb}
\usepackage{bm}
\usepackage{nicematrix}
\usepackage{hyperref}
\hypersetup{
    colorlinks=true,
    linkcolor=blue,
    filecolor=magenta,      
    urlcolor=cyan,
    citecolor=blue,
}
\usepackage{url}
\usepackage[section]{placeins}
\usepackage{graphicx}
\usepackage{wrapfig}
\usepackage{mathtools} 
\usepackage{caption}
\usepackage{subcaption}

\usepackage[toc,page,header]{appendix}
\usepackage{minitoc}

\usepackage{enumitem,lipsum}
\usepackage[most]{tcolorbox}
\tcbset{boxsep=0mm,boxrule=0pt,colframe=white,arc=0mm,left=0.5mm,right=0.5mm}
\usepackage{todonotes}

\usepackage[font=small,labelfont=bf]{caption}

\definecolor{upperbound}{HTML}{CC034D}
\definecolor{rank}{HTML}{F15A33}
\definecolor{params}{HTML}{5A33F1}
\definecolor{rankhf}{HTML}{FFC333}
\definecolor{ranklcn}{HTML}{61AD61}

\usepackage{setspace}
\newcommand{\mb}{\mathbf}

\newcommand{\wMsup}[1]{\mb W^{(#1)}} 

\newcommand{\qtil}{\widehat{q}}

\newcommand{\btheta}{{\pmb \theta}}
\newcommand{\bTheta}{{\pmb \Theta}}

\renewcommand{\a}{\mathbf a}
\renewcommand{\b}{\mathbf b}
\newcommand{\x}{\mathbf x}
\newcommand{\y}{\mathbf y}
\newcommand{\z}{\mathbf z}
\newcommand{\funcsym}{\mathrm{F}}
\newcommand{\outersym}{\mathrm{O}}
\newcommand{\Fn}{\funcsym_{\btheta}}

\newcommand{\Loss}{{\mathrm{L}}}
\newcommand{\Om}{{\mb{\Omega}}}

\newcommand{\covx}{{\bm{\Sigma}_{\x \x}}}

\newcommand{\up}[1]{^{(#1)}}
\newcommand{\E}{\mathbf E \,}

\newcommand{\tens}[1]{\mathcal{#1}}
\newcommand{\conv}[2]{{\tens{#1}}^{(#2)}}
\newcommand{\convfibre}[3]{{\tens{#1}^{(#2)}_{(#3)\,\bullet}}}

\makeatletter
\newtheorem*{rep@theorem}{\rep@title}
\newcommand{\newreptheorem}[2]{%
\newenvironment{rep#1}[1]{%
 \def\rep@title{#2 \ref{##1}}%
 \begin{rep@theorem}}%
 {\end{rep@theorem}}}
\makeatother

\newtheorem{theorem}{Theorem}
\usepackage{mdframed}

\newtheorem{lemma}[theorem]{Lemma}

\newtheorem{proposition}[theorem]{Proposition}
\newtheorem{corollary}[theorem]{Corollary}

\newtheorem{faact}[theorem]{Fact}

\DeclareMathOperator*{\argmin}{argmin}

\DeclareMathOperator{\rank}{rk}

\DeclareMathOperator{\diag}{diag}
\DeclareMathOperator{\vect}{vec}
\DeclareMathOperator{\mat}{mat}

\newcommand{\HO}{\mb{H}_{\outersym}}
\newcommand{\HF}{{\mb{H}_{\funcsym}}}

\newcommand{\HFhat}{{\widehat{\mb{H}}_{\funcsym}}}

\newcommand{\HL}{{\mb{H}_{\Loss}}}
\newcommand{\HOsup}[1]{{\mb{H}^{#1}_{\outersym}}} 
\newcommand{\HFsup}[1]{{\mb{H}_{\funcsym}^{#1}}} 
\newcommand{\HFhatsup}[1]{{\widehat{\mb{H}}_{\funcsym}^{#1}}} 

\newcommand{\HLsup}[1]{{\mb{H}_{\Loss}^{#1}}} 

\newcommand{\opt}{{\btheta^\star}}

\DeclareMathOperator{\toep}{toep}

\newcommand{\Reals}[1]{{\mathbb{R}^{#1}}} 
\newcommand{\matcol}[2]{{#1}_{\bullet\, #2}}
\newcommand{\matrow}[2]{{#1}_{#2 \, \bullet}}

\newcommand{\toepMat}[1]{{\Tm}^{#1}}
\newcommand{\tmat}[1]{{\Tm}^{(#1)}}
\newcommand{\qmat}[1]{{\Qm}^{(#1)}}
\newcommand{\wmat}[1]{{\Wm}^{(#1)}}
\newcommand{\wmatt}[1]{{\Wm}^{(#1)}{}^\top}
\newcommand{\tmatt}[1]{{\Tm}^{(#1)}{}^\top}
\newcommand{\qmatt}[1]{{\Qm}^{(#1)}{}^\top}

\newcommand{\vmat}[1]{{{\Vm}^{(#1)}}}

\newcommand{\tildeqmat}[1]{{{\widetilde{\Qm}}^{(#1)}}}
\newcommand{\tildeqmatt}[1]{{\widetilde{\Qm}}^{(#1)}{}^\top}

\newcommand{\qtilmat}[1]{\widetilde{\Qm}^{(#1)}}
\newcommand{\toepSupRow}[2]{{\Tm}^{{#1}_{#2\, \bullet}}}

\newcommand{\e}{{\mb e}}

\newcommand{\kro}{%
  \mathbin{\mathop{\otimes}}%
}

\usepackage[scr]{rsfso}
\newcommand{\inpspace}{{\mathbb{X}}}
\newcommand{\outspace}{{\mathbb{Y}}}

\def\Am{{\bf A}}
\def\Bm{{\bf B}}
\def\Cm{{\bf C}}
\def\Dm{{\bf D}}

\def\Im{{\bf I}}
\def\Km{{\bf K}}

\def\Pm{{\bf P}}
\def\Qm{{\bf Q}}

\def\Tm{{\bf T}}
\def\Wm{{\bf W}}
\def\Vm{{\bf V}}

\def\Ym{{\bf Y}}
\def\Xm{{\bf X}}
\usepackage{xcolor}
\usepackage{ifmtarg}
\usepackage{xifthen}
\usepackage{environ}
\usepackage{multido}
\usepackage{lipsum}%
\usepackage{mdframed}
\theoremstyle{plain}

\usepackage{thmtools}
\usepackage{cleveref}

\declaretheoremstyle[%
  spaceabove=-2pt,%
  spacebelow=6pt,%
  headfont=\normalfont\itshape,%
  postheadspace=1em,%
  qed=\qedsymbol%
]{mystyle}



\newtheorem{assump}{Assumption}
\newcommand*{\colorboxed}{}
\def\colorboxed#1#{%
  \colorboxedAux{#1}%
}
\newcommand*{\colorboxedAux}[3]{%
  \begingroup
    \colorlet{cb@saved}{.}%
    \color#1{#2}%
    \boxed{%
      \color{cb@saved}%
      #3%
    }%
  \endgroup
}

\newreptheorem{proposition}{Proposition}
\newreptheorem{lemma}{Lemma}
\newreptheorem{theorem}{Theorem}
\newreptheorem{corollary}{Corollary}
\newreptheorem{assump}{Assumption}

\makeatletter
\newcommand{\neutralize}[1]{\expandafter\let\csname c@#1\endcsname\count@}
\makeatother

\icmltitlerunning{The Hessian perspective into the Nature of CNNs}

\begin{document}

\twocolumn[
\icmltitle{The Hessian perspective into the Nature of Convolutional Neural Networks}



\icmlsetsymbol{equal}{*}

\begin{icmlauthorlist}
\icmlauthor{Sidak Pal Singh}{eth}
\icmlauthor{Thomas Hofmann}{eth}
\icmlauthor{Bernhard Schölkopf}{mpi}
\end{icmlauthorlist}

\icmlaffiliation{eth}{ETH Zurich, Switzerland}
\icmlaffiliation{mpi}{MPI for Intelligent Systems, Tubingen, Germany}

\icmlcorrespondingauthor{Sidak Pal Singh}{sidak.singh@inf.ethz.ch}



\icmlkeywords{Machine Learning, ICML}

\vskip 0.3in
]



\printAffiliationsAndNotice{}  

\begin{abstract}
While Convolutional Neural Networks (CNNs) have long been investigated and applied, as well as theorized, we aim to provide a slightly different perspective into their nature --- through the perspective of their Hessian maps. The reason is that the loss Hessian captures the pairwise interaction of parameters and  therefore forms a natural ground to probe how the architectural aspects of CNN get manifested in its structure and properties. We develop a framework relying on Toeplitz representation of CNNs, and then utilize it to reveal the Hessian structure and, in particular, its rank. We prove tight upper bounds (with linear activations), which closely follow the empirical trend of the Hessian rank and hold in practice in more general settings. Overall, our work generalizes and establishes the key insight that, even in CNNs, the Hessian rank grows as the square root of the number of parameters. 
\end{abstract}

\section{Introduction}

\begin{figure}[!t]
    \centering
    \includegraphics[width=0.35\textwidth]{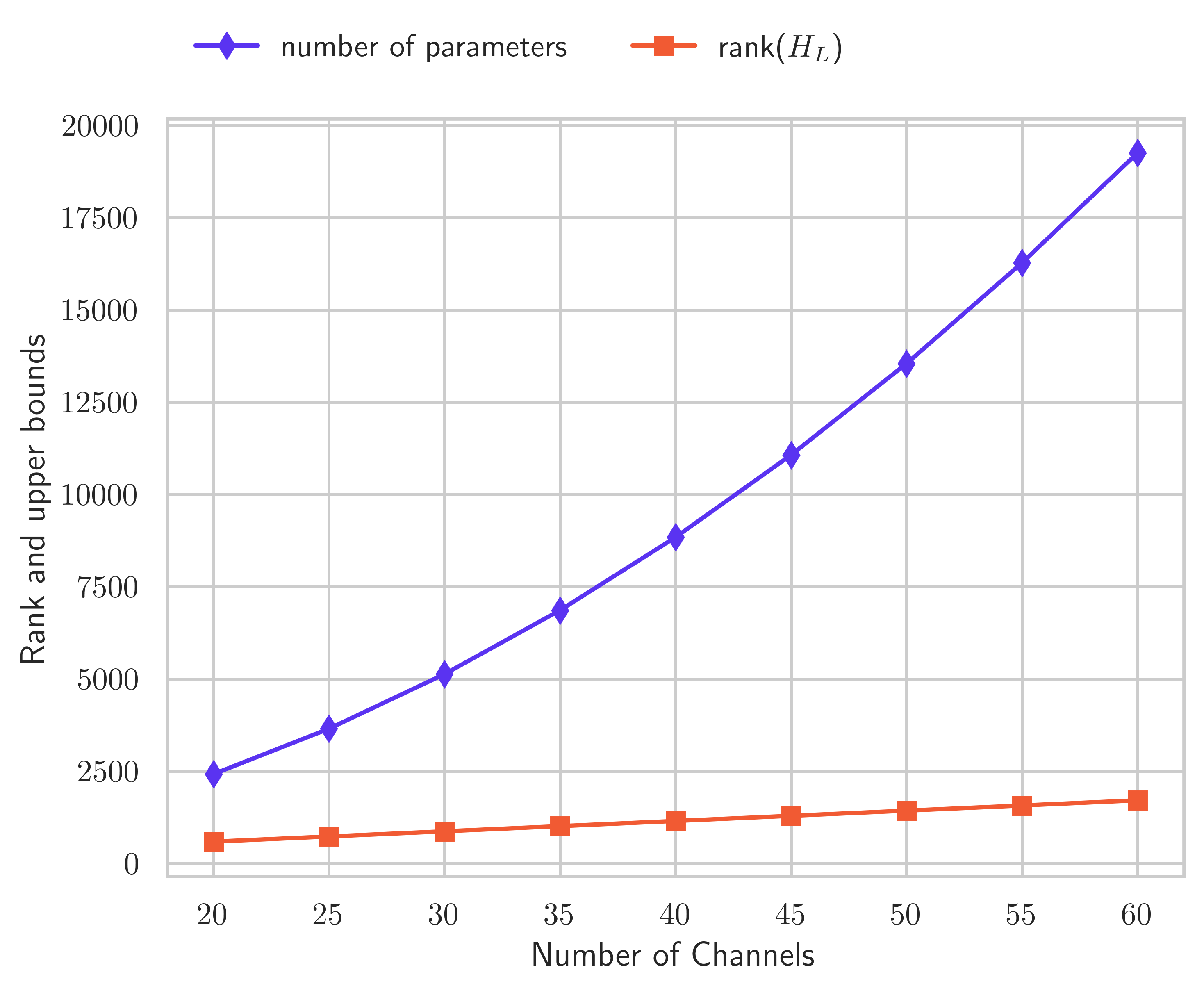}
    \caption{ A comparison of the \textcolor{params}{\textbf{number of parameters}} with the empirically observed \textcolor{rank}{\textbf{loss Hessian rank}} for increasing number of channels $m$ for a $2$-hidden layer CNN on \textsc{CIFAR10}. \textit{Can we estimate the precise scaling behaviour of the Hessian rank?} \looseness=-1}
    \label{fig:teaser}
\end{figure}

Nobody would deny that CNNs~\cite{fukushima1980self,lecun1995convolutional,krizhevsky2017imagenet}, with their baked-in equivariance to translations, have played a key role in the practical successes of deep learning and computer vision. Yet several aspects of their nature are still unclear. Cutting directly to the chase,  for instance, 
take a look at Figure~\ref{fig:teaser}. As the number of channels in the hidden layers increases, the number of parameters grows, as expected, quadratically. But, the rank of the loss Hessian at initialization, measured as precisely as it gets, grows at a much calmer, linear rate. How come?\looseness=-1

This question lies at the core of our work.  Generally speaking, we would like to investigate the inherent `nature'  of CNNs, i.e., how its architectural characteristics manifest themselves in terms of a property or a phenomenon at hand --- here that of Figure~\ref{fig:teaser}, which, among other things, would precisely indicate the effective dimension of the (local) loss landscape~\cite{mackay,gurari2018gradient}.\looseness=-1

Of course, when the likes of Transformer~\cite{vaswani2017attention,dosovitskiy2020image}, MLPMixer~\cite{tolstikhin2021mlp} and their kind, have ushered in a new wave of architecture design, especially when equipped with heaps of data, it is tempting to think that studying CNNs might not be as worthwhile an enterprise. That only time and tide can tell --- but 
it seems that, at least for now, the concepts and principles behind CNNs (such as patches, larger strides, weight sharing, downsampling) continue to be carried along while designing newer architectures in the 2020s~\cite{liu2021swin,liu2022convnet}.  
And, more broadly, if the perspectives and techniques that help us understand the nature of a given architecture are general enough, they may also be of relevance when considering another architecture of interest.\looseness=-1

We are inspired by one such account of an intriguing perspective: the extensive redundancy in the parameterization of fully-connected networks, as measured through the Hessian degeneracy~\cite{sagun2016eigenvalues}, and recently outlined rigorously in the work of~\cite{singh2021analytic}. We aim to chart this out in detail for CNNs. \looseness=-1

\section{Related Work}
\textbf{The Nature of CNNs.}
To start with, there's the intuitive perspective of enabling a hierarchy of features~\cite{lecun1995convolutional}. A more mathematical take is that of~\cite{bruna2013invariant} where filters are constructed as wavelets and which as a whole provide Lipschitz continuity to deformations. The approximation theory view~\cite{poggio2015theory,maoTheory} reinforces the intuitive benefits of hierarchy and compositionality with explicit constructions. 
On the optimization side, the implicit bias perspective~\cite{gunasekar2018implicit} stresses on how gradient descent on CNNs leads to a particular solution. Other notable takes are from the viewpoint of Gaussian processes~\cite{garriga2018deep}, loss landscapes~\cite{nguyen2018optimization,gu2020characterize}, arithmetic circuits~\cite{cohen2016inductive} --- to list a few.  In contrast, we focus on how, \textit{from the initialization itself}, the CNN architecture induces structural properties of the loss Hessian and by extension, of the loss landscape.\looseness=-1

\textbf{Hessian maps and Deep Learning.}
The Hessian characterizes parameter interactions via second derivative of the loss. Consequently, it has been a central object of study and has been extensively utilized for applications and theory alike, both in the past as well as the present. For instance, generalization~\cite{mackay1992practical,keskar2016large,yang2019fastrate,singh2022phenomenology}, optimization~\cite{lbfgs,setiono1995use,DBLP:journals/corr/MartensG15,https://doi.org/10.48550/arxiv.2103.00065}, network compression~\cite{lecun1990optimal,hassibiStork,singh2020woodfisher}, continual learning~\cite{kirkpatrick2017overcoming}, hyperparameter search~\cite{LeCun1992AutomaticLR,schaul2013pesky}, and more.\looseness=-1

In recent times, it has been revitalized in significant part due to the `flatness hypothesis'~\cite{keskar2016large,hochreiter1997flat} and in turn, flatness has become a popular method to probe the extent of generalization as it seems to consistently rank ahead of traditional norm-based complexity measures~\cite{jiang2019fantastic} in multiple scenarios. Given the increasing size of the networks, and the inherent limitation of the Hessian being quadratic in cost, measuring flatness has almost become synonymous with measuring the top eigenvalue of the Hessian~\cite{https://doi.org/10.48550/arxiv.2103.00065} or even just the zeroth-order measurement of the loss stability in the parameter space. To a lesser extent, some works still utilize efficient approximations to the Hessian trace~\cite{yao2020pyhessian} or log determinant~\cite{jia2020information}. But largely the reliance on the Hessian for neural networks has become a black-box affair.

\textbf{Understanding of the Neural Network Hessian maps.} Lately, significant advances have been made in this direction --- a few of the most prominent being: the characterization of its spectra as bulk and outliers~\cite{sagun2016eigenvalues,pmlr-v70-pennington17a,ghorbani2019investigation}, the empirically observed significant rank degeneracy~\cite{sagun2017empirical}, and the class/cross-class structure~\cite{papyan2020traces}. Despite these advancements, its structure for neural networks primarily gets seen only up to the surface level of the chain rule~\footnote{This is known otherwise as the Gauss-Newton decomposition~\cite{schraudolph2002fast} and discussed in Eqn.~\ref{eq:hess}.} for a composition of functions, with a few exceptions such as~\cite{wu2020dissecting,singh2021analytic}. 

We take our main inspiration from the latter of these works, namely~\cite{singh2021analytic}, where the authors precisely characterize the structure for deep fully-connected networks (FCNs) resulting in concise bounds and formulae on the Hessian rank of arbitrary sized networks. Our aim, thus, is to thoroughly exhibit the structure of the Hessian for deep convolutional networks and explore the distinctive facets that arise in CNNs --- as compared to FCNs. 

\textbf{Our contributions:}
(1) We develop a framework to analyze CNNs that rests on a Toeplitz representation of convolution\footnote{Convolution as used in practice in deep learning, and not, say circular convolution ---despite its relative theoretical ease.} and applies for general deep, multi-channel, arbitrary-sized CNNs, while being amenable to matrix analysis. We then utilize this framework to unravel the Hessian structure for CNNs and provide upper bounds on the Hessian rank. Our bounds are exact for the case of 1-hidden layer CNNs, and in the general case, are of the order of square root of the trivial bounds.  

(2) Next, we verify our bounds empirically in a host of settings, where we find that our upper bounds remain rather close to the true empirically observed Hessian rank. Moreover, they even hold faithfully outside the confines of the theoretical setting (choice of loss and activation functions) used to derive them.

(3) Further, we make a detailed comparison of the key ingredients in CNNs, i.e., local connectivity and weight sharing, in a simplified setting through the perspective of our Hessian results. We also discuss some elements of our proof technique in the hope that it helps provide a better grasp of the results.

We would also like to make a quick remark about the
\textit{difference with regards to~\cite{singh2021analytic}.} While we borrow heavily from their approach, the framework we develop here provides us with the flexibility to handle convolutions, pooling operations, and even fully-connected layers. In particular, our analysis captures their results as a special case when the filter size is equal to the spatial dimension. We also introduce a novel proof technique relative to~\cite{singh2021analytic}, without which one cannot attain exact bounds on the Hessian rank for the one-hidden layer case. 

Overall, we hope that by building on the prior work, we can further push this research direction of understanding the nature of various network architectures through an in-depth, white-box, analysis of the Hessian structure.




\vspace{-1em}
\section{Setup and Background}
\paragraph{Notation.} We denote vectors in lowercase bold ($\x$), matrices in uppercase bold ($\Xm$), and tensors in calligraphic letters ($\mathcal{X}$). We will denote the $i$-th row and $j$-th columns of some matrix $\Am$ by $\matrow{\Am}{i}$ and $\matcol{\Am}{j}$ respectively. We will often use the notation $\Am\up{i:j}$, for $i>j$ to indicate a sequence of matrices from $i$ down to $j$, i.e., $\Am\up{i:j}=\Am\up{i} \Am\up{i-1} \cdots \Am\up{j+1}\Am\up{j}$. For $i< j$, the same notation\footnote{When either of $i$ or $j$ are outside the bounds of a particular index set, this notation would devolve to an identity matrix.} would mean the sequence of matrices from $i$ up to $j$, but \textit{tranposed}.  To express the structure of the gradients and the Hessian, we will employ matrix derivatives~\cite{magnus2019matrix}, wherein we vectorize row-wise ($\vect_r$) the involved matrices and organize the derivative in Jacobian (numerator layout). Concretely, for matrices $\Xm \in\Reals{m\times n}$ and $\Ym \in\Reals{p\times q}$, we have $$\dfrac{\partial \Ym}{\partial \Xm} := \dfrac{\partial\vect_r \Ym}{\partial\,(\vect_r \Xm)^\top}\in\Reals{pq\times mn}\,.$$
\paragraph{Setting.} Suppose we are given an i.i.d. dataset $S=\{(\x_1, \y_1), \dots, (\x_n, \y_n)\}$, of size $|S| = n$,  drawn  from an unknown distribution $p_{\inpspace, \outspace}$, consisting of inputs $\x \in \inpspace\subseteq\mathbb{R}^{d}$ and targets $\y \in \outspace\subseteq\mathbb{R}^{K}$. Based on this dataset $S$, consider we use a neural network to learn the mapping from the inputs to the targets, $\Fn: \inpspace \mapsto \outspace$, parameterized by $\btheta\in\bTheta\subseteq\mathbb{R}^p$. To this end, we follow the framework of Empirical Risk Minimization~\cite{vapnik1991principles}, and optimize a suitable loss function $\Loss: \Theta  \mapsto \mathbb{R}$. In other words, we solve the following optimization problem, $$\opt = \argmin_{\btheta \in \bTheta} \; \Loss(\btheta) = \dfrac{1}{n} \sum\limits_{i=1}^n \ell\left(\btheta; (\x_i, \y_i)\right)\,,$$ say with a first-order method like (stochastic) gradient descent and the choices for $\ell$ could be mean-squared error (MSE), cross-entropy (CE), etc. \looseness=-1

\paragraph{Hessian.} We analyze the properties of the Hessian of the loss function, $\HL = \dfrac{\partial^2 \Loss(\btheta)}{\partial \btheta \, \partial \btheta{}^\top}$, with respect to the parameters $\btheta$. It is quite well known~\citep{Schraudolph2002FastCM,sagun2017empirical} that, via the chain rule, the Hessian can be decomposed as a sum of the following two matrices:
\begin{align}\label{eq:hess}
    \HL =  \HO + \HF &=  \frac{1}{n}\sum_{i=1}^n \nabla_{\btheta} \Fn(\x_i) \big[\nabla^2_{\Fn} \,\ell_i\big] \, \nabla_\btheta \Fn(\x_i)^\top \notag\\
    &+ \frac{1}{n}\sum_{i=1}^n \,
    \sum_{c=1}^K [\nabla_{\Fn}\ell_i]_c\, \nabla^2_\btheta \, \Fn^{c}(\x_i)\end{align}
where, $\nabla_\btheta \Fn(\x_i) \in\mathbb{R}^{p\times K}$ is the Jacobian of the function and $\nabla^2_{\Fn} \,\ell_i \in\mathbb{R}^{K\times K}$ is the Hessian of the loss with respect to the network function, at the $i$-th sample. To facilitate comparison,  we refer to these two matrices as the \textit{outer-product Hessian} and the \textit{functional Hessian} following~\citep{singh2021analytic}.\looseness=-1
\subsection{Background}
\paragraph{Precise bounds on the Hessian rank.} Despite the much interest in Hessian over the decades, the upper bounds remained quite trivial (like a factor of the number of samples) and the empirically observed degeneracy~\cite{sagun2016eigenvalues}, because of the need to measure rather small eigenvalues and judge a suitable threshold, remained intractable to establish.  Exact upper bounds and formulae have only been made available very recently for fully-connected networks due to~\cite{singh2021analytic}, as a result of subtle choice to have linear activations, together with a novel analysis technique adapted from matrix analysis~\cite{doi:10.1080/03081087408817070,CHUAI2004129}. While they find that the presence of non-linearities like ReLU impedes a theoretical analysis,~\cite{singh2021analytic} thoroughly demonstrate that their bounds hold empirically for ReLU-based FCNs as well. Their empirical analysis is equally rigorous, for they compute exact Hessians (without any approximation) in Float64 precision. \looseness=-1

Overall, the surprising finding of~\cite{singh2021analytic} is that the Hessian rank for FCNs scales\footnote{More precisely, this should be $\mathcal{O}(q\cdot m)$, however, $q$ denotes the bottleneck dimension in the network --- inclusive of input and output dimensions, and hence can be thought of as a constant.} linearly in the total number of neurons $m$, i.e.,  $\mathcal{O}(m)$; while the number of parameters scale quadratically, $\mathcal{O}(m^2)$.  While this concrete bound of $\mathcal{O}(m)$ is shown at initialization, their analysis applies to any point during training --- but with  resulting bound in terms of the rank of the individual weight matrices. Further, they highlight that during training, the rank can only decrease (a simple consequence of the functional Hessian $\HF$ being driven to zero since it is scaled by the gradient of the loss $\nabla_{\Fn}\ell$). In this sense, their upper bounds hold pointwise throughout the loss landscape.

\paragraph{Why Hessian Rank?} The finding about the growth of Hessian rank carries thought-provoking implications on the number of effective parameters within a particular architecture. Let us illustrate this by considering the Occam's factor~\cite{mackay,Gull1989} which, said roughly, describes the extent to which the prior hypothesis space shrinks on observing the data. More formally, this is the ratio of posterior to prior volumes in the parameter space. This is then used to derive the following measure for the effective degrees of freedom, assuming a quadratic approximation to the posterior:
$\sum\limits_{i=1}^p \dfrac{\lambda_i}{\lambda_i + \epsilon}\,,$
where, $\lambda_i$ denotes the $i$-th eigenvalue of the Hessian, $p$ is the total number of parameters, and $\epsilon$ is the weight set upon the prior. In other words, the above measure compares the extent ($\lambda_i$) to which a particular direction (along the $i$-th eigenvector) in the parameter space is determined by the data relative to that determined by the prior $\epsilon$. For small $\epsilon$,  which amounts to little or no explicit regularization towards the prior (as is often the case with deep networks in practice), this measure of the degrees of freedom approaches the Hessian rank. In a recent work~\cite{parametercountingMaddox} empirically noted that this measure also explains double descent~\cite{belkin2019reconciling} in neural networks. Thus, it makes it all the more pertinent\footnote{As a sidenote, we carry out a (limited) scale study where we sweep over the filter sizes and number of channels in a CNN, and find that the Hessian rank, \textit{at initialization,} has a higher correlation coefficient (see Figure~\ref{fig:gen}) with generalization error as compared to the raw count of parameters. However, an extensive study is beyond the scope of our paper. } to explore how rank of the Hessian scales with various architectural parameters in a CNN.




\section{Toeplitz Framework for CNN Hessians}\label{sec:rankformulae-1hl}
\textbf{Preliminaries.}
In this section, we will lay out the formalism that will lie at the core of our analysis. For brevity, we will develop this in the case of 1D CNNs which are commonly employed for biomedical applications or audio data~\cite{kiranyaz20211d}. One can otherwise think of applying our framework to flattened filters and input patches, and such an assumption is also prevalent in the theory literature~\cite{montufarCnn}. Besides, throughout our framework we will assume there are no bias parameters, although one can simply consider homogeneous coordinates in the input. 



\paragraph{Warmup.} Let's say we want to represent the convolution $\Wm*\x$ of an input $\x\in\Reals{d}$ with $m$ filters of size $k\leq d$ that have been organized in the matrix $\Wm\in\Reals{m\times k}$. For now, consider that we have stride $1$ and zero padding. We will later touch upon these aspects and see the results for strides $> 1$ in Section~\ref{sec:lcn+ws}. Further, for some vector $\z \in \Reals{d}$, we will use the notation $\z_{j:j+k-1}\in\Reals{k}$ to denote the (shorter) vector formed by considering the indices $j$ to $j+k-1$ (both inclusive) of the original vector. The output of the above convolution can be expressed as the following matrix of shape $m\times (d-k+1)$,
\vspace{-0.1em}
\begin{align}\label{eq:conv-op}
\small
    \hspace{-1em}\Wm \ast \x = \begin{pmatrix}
     \langle\matrow{\Wm}{1}, \,\x_{1:k}\rangle & \cdots 
     &\langle\matrow{\Wm}{1}, \,\x_{d-k+1:d}\rangle 
     \\[2mm]
     \vdots & & \vdots \\[2mm]
     \langle\matrow{\Wm}{m}, \,\x_{1:k}\rangle & 
     \cdots &
     \langle\matrow{\Wm}{m}, \,\x_{d-k+1:d}\rangle
    \end{pmatrix}\,.
\end{align}
Now, define Toeplitz\footnote{These are matrices $\Am$ with $\Am_{ij} = \a_{i-j}$ formed via some underlying vector $\a$} matrices for each filter, $\{\toepSupRow{\Wm}{i}\}_{i=1}^m$, with $\toepSupRow{\Wm}{i} := \toep(\matrow{\Wm}{i}, d) \in \Reals{(d-k+1) \times d}$ such that, 
$$\small\toepSupRow{\Wm}{i}= \begin{pmatrix}
 w_{i1} & \cdots & w_{ik} & 0 & \cdots & 0 \\[2mm]
 0 & w_{i1} & \cdots & w_{ik} & 0 & \vdots \\[2mm]
 \vdots &0 & \ddots & \ddots & \ddots& 0\\[2mm]
 0  & \cdots & 0 &  w_{i1}  & \cdots & w_{ik} \\[2mm]
\end{pmatrix}\,.$$

Note, the above representation of the Toeplitz matrix also depends on the base dimension (here, $d$) where the given vector must be `toeplitzed', i.e., circulated in the above fashion. But we will omit specifying this unless necessary. 
Let us also denote the matrix formed by stacking the $\toepSupRow{\Wm}{i}$ matrices in a row-wise fashion as $$\small\toepMat{\Wm}:=\begin{pmatrix}
 \toepSupRow{\Wm}{1}\\[1mm]
 \vdots \\[1mm]
 \toepSupRow{\Wm}{m} \\[1mm]
\end{pmatrix}\,\in\Reals{m(d-k+1)\times d}\,.$$
We can now see that the above matrix, $\toepMat{\Wm}$, when multiplied by the input $\x$ gives us the output of the convolution operation in Eqn.~\eqref{eq:conv-op} when vectorized row-wise , i.e., $$\vect_r(\Wm*\x) = \toepMat{\Wm}\x\,.$$

\subsection{Toeplitz representation of deep CNNs} 
Now, let us assume we have $L$ hidden layers, each of which is a convolutional kernel. Hence, the parameters of the $l$-th layer are denoted by the tensor $\conv{W}{l}\in\Reals{m_{l}\times m_{l-1} \times k_{l}}$, where $m_{l}$ represent the number of output channels, $m_{l-1}$ the number of input channels, and $k_l$ the kernel size at this layer. As we assume a one-dimensional input, without loss of generality, we can set the number of input channels $m_0=1$. In other words, they are already assumed to be flattened when passing the input of dimension $d_0:=d$ into the network. The spatial dimension after being convolved with the $l$-th layer is denoted by $d_l = d_{l-1} - k_l + 1$ (which is basically the number of hops we can make with the given kernel over its respective input), since we have stride $1$ and zero padding. 
Assume, say the ReLU nonlinearity $\sigma(x):=\max(x,0)$. The network function can then be formally represented as (although it will be actually defined through Eqn.~\eqref{eq:conv-general-actual} later):
$$
\Fn(\x) = \conv{W}{L+1}\ast \sigma(\conv{W}{L}\ast \sigma(\,\cdots \ast \sigma(\conv{W}{1} \ast \x)))\,.
$$
As before, we would like to express the above function in terms of a sequence of appropriate Toeplitz matrix products. Unlike the warmup scenario, the convolutional kernels in this general case will be tensors. The key idea is to do a column-wise stacking of individual Toeplitz matrices across the input channels while maintaining the row-wise stacking, as before, across the output channels.

First, we need to introduce a notation about indexing the fibres of a tensor $\conv{W}{l}\in\Reals{m_l\times m_{l-1}\times k_l}$. Say we need the fibre going in to the plane, across the third mode. Then, the $(i, j)$-th fibre is denoted by $\convfibre{W}{l}{i, j}\in\Reals{k_l}$, whose associated Toeplitz matrix will be $\toepMat{\convfibre{W}{l}{i, j}}   := \toep(\convfibre{W}{l}{i, j}, d_{l-1})\in\Reals{d_l\times d_{l-1}}$. Finally, the Toeplitz matrix associated with the entire $l$-th convolutional layer $\conv{W}{l}$, for which we use the shorthand $\toepMat{(l)}\in\Reals{m_l d_l \times m_{l-1} d_{l-1}}$,  can be expressed as:
$${\toepMat{(l)}:=\begin{pmatrix}
 \toepMat{\convfibre{W}{l}{1,1}} & \cdots & \toepMat{\convfibre{W}{l}{1, m_{l-1} }}\\[1mm]
 \vdots & & \vdots \\[1mm]
 \toepMat{\convfibre{W}{l}{m_l, 1 }} & \cdots & \toepMat{\convfibre{W}{l}{m_l, m_{l-1} }}  \\[1mm]
\end{pmatrix}\,.}$$

In other words, the Toeplitzed representation $\toepMat{(l)}$ consists of $m_l \cdot m_{l-1}$ many Toeplitz blocks formed by the vector $\convfibre{W}{l}{i, j}\in\Reals{k_{l}}$ of size $d_{l}\times d_{l-1}$. The output (spatial) dimension  $d_{L+1}$ is typically 1 (corresponding to $k_{L+1}=d_{L}$), and the number of output channels for the last layer $m_{l+1}=K$  where $K$ is the number of targets. 
Now, the network function can be written in the general case as:
\begin{equation}\label{eq:conv-general-actual}
\Fn(\x) = \toepMat{(L+1)}\Lambda\up{L}\toepMat{(L)}\cdots \Lambda\up{1}\toepMat{(1)}\,\x\,,
\end{equation}
where, $\Lambda\up{i}$ is an input-dependent diagonal matrix that contains a $1$ or $0$, based on whether the neuron was activated or not. As the rank analysis of~\cite{singh2021analytic} requires linear activations, we will follow the same course. However, we can expect that this should still let us contrast the distinctive facets of CNN relative to a FCN. Anyways, later we will elaborate on the case of nonlinearities. Besides, we will refer to the above network function, more concisely as $\tmat{L+1:1}\x$.\looseness=-1

\textbf{Remark.} As evident, the fact that a convolution of a set of vectors can be expressed as a matrix-vector product with a suitable Toeplitz matrix is rather straightforward. Also, a Toeplitz representation for CNN is not new --- works from approximation theory incorporate a similar formalism~\cite{zhao2017theoretical,fang2020theory}. However, unlike past work~\cite{sedghi2018singular,montufarCnn}, we develop our framework in a way that doesn't overlook how convolutions are prominently used, i.e.,  without circular convolution, with multiple channels, and possibly unequal-sized layers.\looseness=-1



\subsection{Matrix derivatives of Toeplitz representations} For the gradient and Hessian calculations, we make use of matrix derivatives and the corresponding chain rule. Thus we frequently compute the gradient of the Toeplitz representation $\toepMat{(l)}$ with respect to the suitably matricized convolutional tensor, $\mat\conv{W}{l}$. In order to be consistent with the form of $\toepMat{(l)}$, we define our matricization of $\conv{W}{l}$ as:
\begin{equation}
\small
    \mat{\conv{W}{l}} := \begin{pmatrix}
        {\convfibre{W}{l}{1, 1}} & \cdots 
 &{\convfibre{W}{l}{m_l, 1} } \\
        \vdots &  & \vdots \\
        {\convfibre{W}{l}{1, m_{l-1}}} & \cdots &  {\convfibre{W}{l}{m_l, m_{l-1}}} 
    \end{pmatrix}^\top\,.
\end{equation}
where the matricization $\mat{\conv{W}{l}}\in \Reals{m_l \times m_{l-1} k_l}$ and we will use the notation $\wMsup{l}:=\mat{\conv{W}{l}}$ as a shorthand. Essentially, we have arranged each of the mode-$3$ fibres as rows in the output channels times input channels format. 

The following lemma equips us with the way to carry this out (the proof can be found in Section~\ref{sec:toeplitz-proof} of the Appendix). 
\begin{lemma}\label{lemma:toep-grad}
	The matrix derivative of  $\toepMat{(l)}$ with respect to  $\wMsup{l}$,  is given as follows:
	\[\widetilde{\Qm}^{(l)}:=\dfrac{\partial \toepMat{(l)} }{\partial \wMsup{l}}:=\dfrac{\partial \vect_r \toepMat{(l)}}{\partial \left(\vect_r \wMsup{l}\right)^\top } = \Im_{m_l}  \, \kro\, \qmat{l}\,.\]
\end{lemma}

The particular structure of $\qmat{l}$ is a bit complex, involving various permutation matrices. So, for simplicity, we abstract it out here in the main text.

\subsection{CNN Hessian Structure}
Like~\cite{singh2021analytic}, for our theoretical analysis, we will consider the case of MSE loss. But the results still hold empirically, say, for CE loss. Let's now have a glance at the $kl$-th block, $k\leq l$ for both the outer-product and the functional Hessian (the derivations are in Section~\ref{sec:struct}). This corresponds to looking at the submatrix of the Hessian corresponding to the $k$-th and $l$-th convolutional parameter tensors.\footnote{In the case of $\HF$, the present form is for $k\leq l$. For, $k\geq l$, it's a bit different and detailed in the Appendix.} Let's  
start with the outer-product Hessian $\HO$.

\begin{proposition}\label{prop:ho}
The $kl$-th block of $\HO$ is, 
\begin{align}\label{eq:ho-kl-main}
    \hspace{-4em}\HOsup{(kl)} &={\tildeqmatt{k}}
    \bigg(\tmat{k+1:L+1} \tmat{L+1:l+1} \,\,\kro \notag\\
    &\hspace{5em} \tmat{k-1:1}\covx \tmat{1:l-1}\bigg)\tildeqmat{l}\,.
\end{align}
\end{proposition}

\textbf{A word about $\HO$.} We should also emphasize that the outer-product Hessian shares exactly the same non-zero spectrum as the Neural Tangent Kernel~\cite{jacot2018neural}, or roughly up to scaling, in the case of Fisher Information~\cite{amari1998natural}, empirical Fisher~\cite{kunstner2019limitations}).

Moving on to the functional Hessian $\HF$, denote $\Om = \E[\pmb \delta_{\x, \y} \,\x^\top]\in\Reals{K\times d_0}$ as  the (uncentered) covariance of the residual $\left(\y_i-\Fn(\x_i)\right)$ with the input. Then we have,
\begin{proposition}\label{prop:hf}
    The $kl$-th block of $\HF$ is:
\begin{align}\label{eq:hf-lk-main}
     \hspace{-1em}\HFsup{(kl)} &= \left(\Im_{m_k} \kro \qmatt{k}  \right)\bigg(\tmat{k+1:l-1} \, \, \kro \notag\\
   & \hspace{3em}\tmat{k-1:1}\Om^\top\tmat{L+1:l+1} \bigg) \left(\qmat{l} \kro \Im_{m_l}\right)\,,
\end{align}
\end{proposition}
\textbf{A word about $\HF$.} As the outer-product Hessian is positive semi-definite, the functional Hessian is the source of all the negative eigenvalues of the Hessian and is important for optimization as there may be numerous saddles in the landscape~\cite{dauphin2014identifying}. It also has a very peculiar block-hollow structure (i.e., zero diagonal blocks), which leads to the number of negative eigenvalues being approximately half of its rank (c.f.,~\cite{singh2021analytic}).

\section{Key results on the CNN Hessian Rank}
Finally, we can now present our key results. A quick note about assumptions: for simplicity, assume that the (uncentered) input-covariance  $\covx = \operatorname{cov}(\x)$ has full rank $\operatorname{rank}\left(\covx\right):=r=d$. We will analyze the ranks of the outer product and functional Hessian, later combining them to yield a bound on the rank of the loss Hessian.
This is without loss of generality for one can always pre-process the input to ensure that this is the case, and results of~\cite{singh2021analytic} hold for $r\neq d$ with appropriate modifications.  

\textbf{Outer-Product Hessian $\HO$.}
From the structure of the $kl$-th block in Eqn.~\eqref{eq:ho-kl-main}, it's easy to see to arrive at the following proposition: 

\begin{proposition}\label{eq:outer-decomp}
For a deep linear convolutional network, $\,\HO = \Qm_o^\top\Am_o  \Bm_o {\Am_o^\top}\Qm_o \,$, where $\,\Bm_o = \Im_K \kro \covx\, \in \mathbb{R}^{Kd\times Kd}$,
$\,\,\,\Am_o^\top=\begin{pmatrix}
\tmat{L+1:2} \kro \Im_d \\[1mm]
\vdots \\[1mm]
\tmat{L+1:l+1} \kro \tmat{1:l-1}\\[1mm]
\cdots \\[1mm]
\Im_K \kro \tmat{1:L}
\end{pmatrix}\,\, \in \mathbb{R}^{Kd\times \widehat{p}}\,,
$ and $\Qm_o =\diag\left(\qtilmat{1}, \cdots, \qtilmat{L+1}\right)\in\Reals{\widehat{p}\times p}$ where $\diag(\cdot)$ denotes a block-diagonal matrix. \looseness=-1
\end{proposition}
Besides, $p=\sum_{i=1}^{L+1} m_l m_{l-1}{k_l}$ and $\widehat{p} = \sum_{i=1}^{L+1} m_l m_{l-1} d_{l} d_{l-1}$. The former denotes the number of parameters in the CNN while the latter is the number of parameters in the `Toeplitzed' fully-connected network.

Our first key result, the proof of which is located in Section~\ref{sec:ho-cnn-proof} of the Appendix, can then be described as follows : \looseness=-1

\begin{mdframed}[leftmargin=1mm,
    skipabove=2mm, 
    skipbelow=-1mm, 
    backgroundcolor=gray!10,
    linewidth=0pt,
    leftmargin=-1mm,
    rightmargin=-1mm,
    innerleftmargin=2mm,
    innerrightmargin=2mm,
    innertopmargin=2mm,
innerbottommargin=1mm]
\begin{theorem}\label{theorem:ub-outer} The rank of the outer-product Hessian  is upper bounded as
\begin{align*}
\rank(\HO) &\leq \min\big(p, 
d_0 \rank(\tmat{2:L+1}) + K \rank(\tmat{L:1}) \, - \\
&\hspace{5em }\rank(\tmat{2:L+1})\rank(\tmat{L:1})\big) \\
&= \min\left(p, q \,( d_0 + K  - q)\right)\,.
\end{align*}
Here, $q:=\min(d_0, m_1 d_1, \cdots, m_L d_L, K)$.
\end{theorem}
\end{mdframed}

Assuming no bottleneck layer, we will have that $q=\min(d, K)$, and resulting in $\rank(\HO)\leq Kd_0$.

\paragraph{Functional Hessian $\HF$.}
Our approach here will be similar to that in the Theorem above. We will try to factor out all the $\qmat{l}$ matrices and then analyze the rank of the resulting decomposition. But, this requires more care as the form of the $kl$-th block is different depending on $k\leq l$ or not.

\begin{mdframed}[leftmargin=1mm,
    skipabove=2mm, 
    skipbelow=-1mm, 
    backgroundcolor=gray!10,
    linewidth=0pt,
    leftmargin=-1mm,
    rightmargin=-1mm,
    innerleftmargin=2mm,
    innerrightmargin=2mm,
    innertopmargin=2mm,
innerbottommargin=1mm]
\begin{theorem} \label{theorem:func-hess-cols}
For a deep linear convolutional network, the rank of $l$-th column-block, $\HFhatsup{\bullet l}$, of the matrix $\HFhat$, can be upper bounded as
 $$\rank(\HFhatsup{\bullet l}) \leq \min(\qtil\, m_{l -1} d_{l-1} + \qtil\,m_{l} d_{l}  - \qtil^{\,2}\,, m_l m_{l-1} k_l)\,,$$ 
 for $l \in [2, \cdots , L]\,.$ When $l=1$, we have 
 $$
\rank(\HFhatsup{\bullet 1}) \leq \min(\qtil\, m_{1} d_1 + \qtil \, s - \qtil^{\,2}\,, m_1 m_0 k_1)\,.$$ And, when $l=L+1$, we have $$
\rank(\HFhatsup{\bullet L+1}) \leq \min(\qtil\, m_{L} d_L + \qtil \, s - \qtil^{\,2}\,, m_{L+1}m_L k_{L+1})\,.$$
Here, $\qtil := \min(d_0, m_1 d_1, \cdots, m_{L} d_L, K, s) = \min(q, s)$ and $s:=\rank(\Om)=\rank(\E[\pmb \delta_{\x,\y}\,\x^\top])$.
\end{theorem}
\end{mdframed}

The proof is located in Section~\ref{sec:hf-cnn-proof} of the Appendix. The upper bound on the rank of $\HF$ follows by summing the above result over all the columns, $\rank(\HF) \leq\sum\limits_{l=1}^{L+1} \rank(\HFhatsup{\bullet l})$ by the sub-additivity of rank. A remarkable empirical observation, like in the case of FCNs, is that the block-columns are mutually orthogonal --- hence, we don't loose anything by simply summing the ranks of the block columns.


\textbf{Loss Hessian $\HL$.}
One can then bound the rank of the loss Hessian simply as, $\rank(\HL)\leq \rank(\HO) + \rank(\HF)$. Also, we can infer that, in the likely case where $q=\min(K, d_0)$, rank of the loss Hessian grows linearly with number of channels, i.e., $\mathcal{O}(m\cdot L\cdot d_0)$, while the number of parameters grow quadratically in the number of channels, i.e., $\mathcal{O}(m^2\cdot L\cdot d_0)$. Thereby, we confirm that like FCNs, a similar linear trend also holds for CNNs (and hence the Figure~\ref{fig:teaser}).
Besides, for very large networks, $m$ will be the dominating factor and we can infer that rank will show a square root behaviour relative to the number of parameters. Hence, we generalize the key finding of~\cite{singh2021analytic} in the fully-connected case to the case of convolutional neural networks. 



\begin{figure*}[!t]
    \centering
    \begin{subfigure}[b]{0.31\textwidth}
    \includegraphics[width=\textwidth]{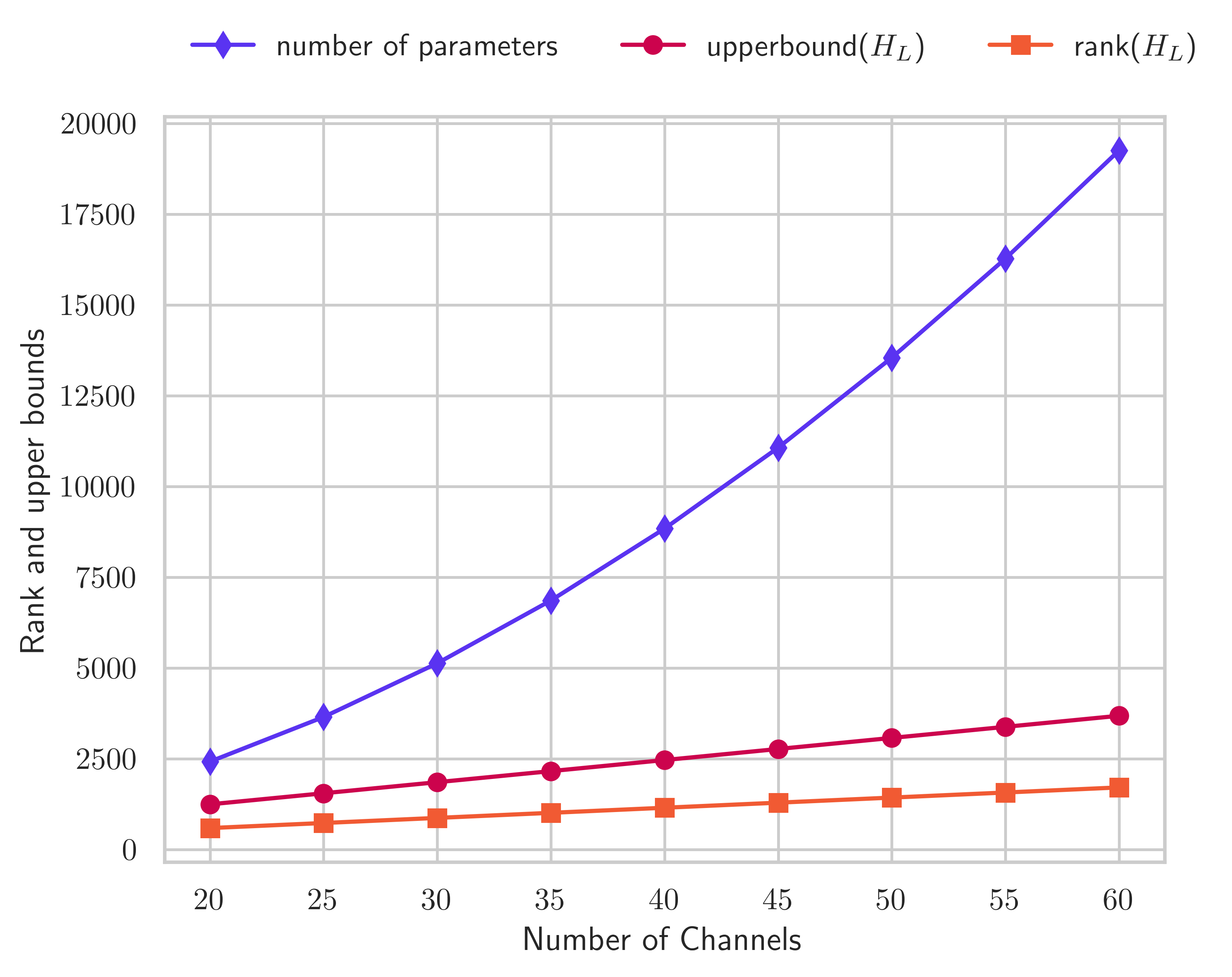}
    \caption{Linear CNN: Rank vs \# Channels}
    \label{fig:mylabel}
    \end{subfigure}
    \begin{subfigure}[b]{0.33\textwidth}
    \includegraphics[width=\textwidth]{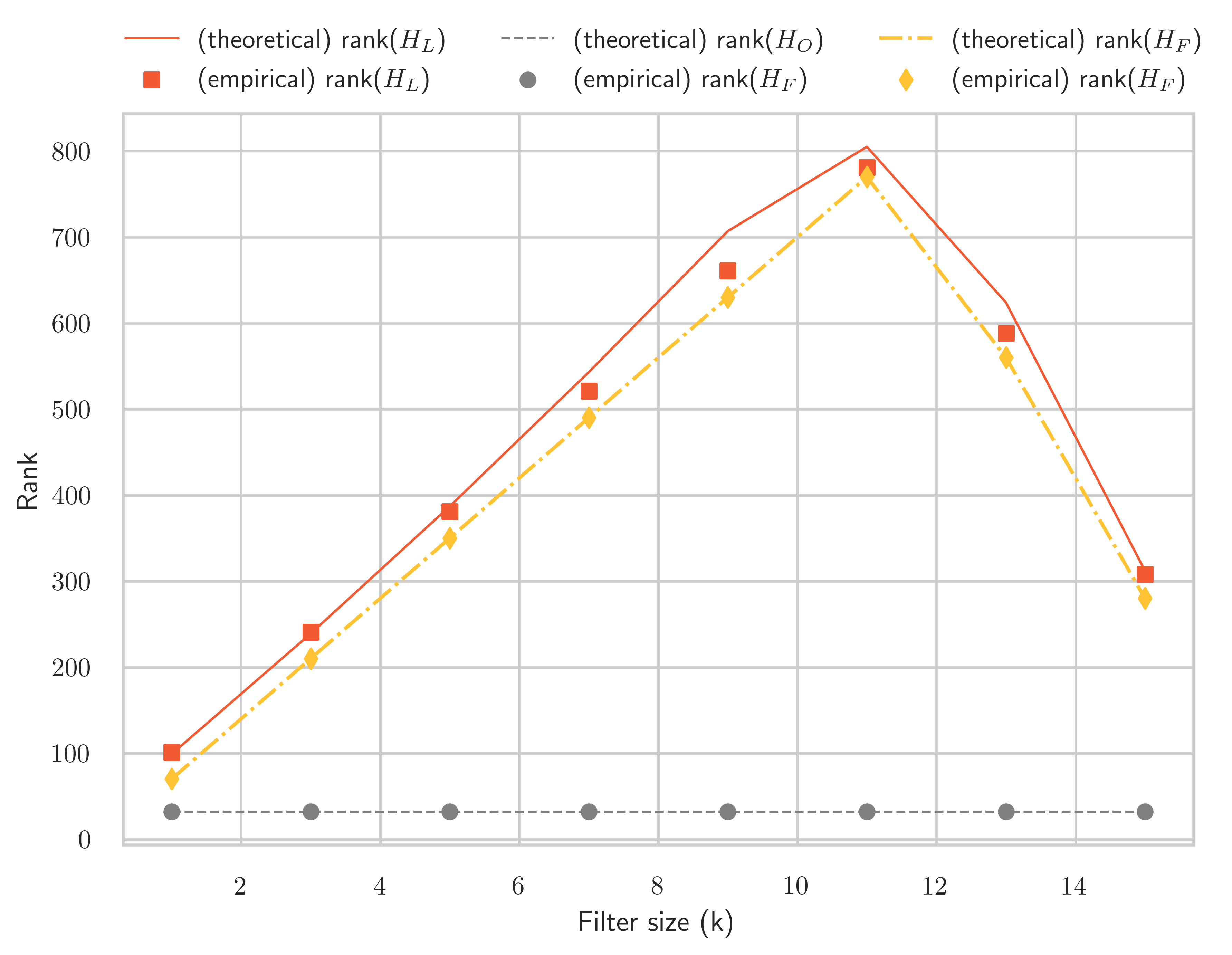}
    \caption{Linear CNN: Rank vs Filter size}
    \label{fig:linfil}
    \end{subfigure}
    \begin{subfigure}[b]{0.33\textwidth}
    \includegraphics[width=\textwidth]{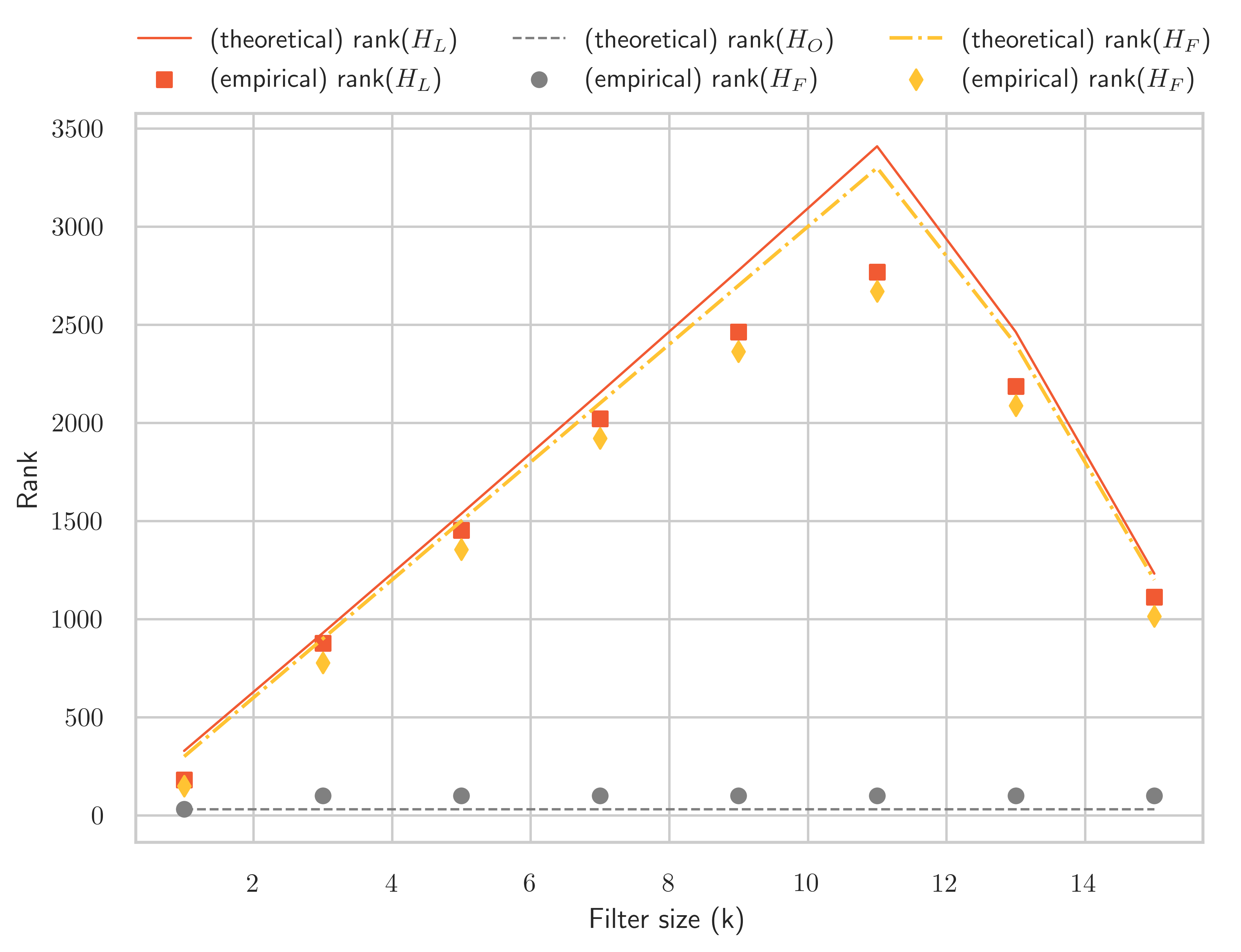}
    \caption{ReLU CNN: Rank vs Filter size}
    \label{fig:relfil}
    \end{subfigure}
    \caption{Trend of the \textcolor{upperbound}{\textbf{upper bound}} on the (loss) Hessian rank, compared to the \textcolor{rank}{\textbf{true rank}}, and the \textcolor{params}{\textbf{number of parameters}}. In other subfigures, we see the rank of the \textcolor{rankhf}{functional Hessian} and \textcolor{gray}{outer-product Hessian}.}
    \vspace{-1em}
\end{figure*}

    
\subsection{Exact results for one-hidden layer case}
While the above bounds are much tighter than any existing bounds, we now try to understand how tight our bounds are by looking at the case of a 1-hidden layer:\looseness=-1 
\begin{mdframed}[leftmargin=1mm,
    skipabove=2mm, 
    skipbelow=-1mm, 
    backgroundcolor=gray!10,
    linewidth=0pt,
    leftmargin=-1mm,
    rightmargin=-1mm,
    innerleftmargin=2mm,
    innerrightmargin=2mm,
    innertopmargin=2mm,
innerbottommargin=1mm]
\begin{theorem}\label{thm:one-hid-cnn-ho}
The rank of the outer product Hessian can be bounded as:
$\operatorname{rank}\left(\HO\right) \leq \min\big(Kd_0, q (k_1 + K(d_0-k_1+1) - q)\big)\,$, where $q=\min(k_1, K(d_0 - k_1 +1), m_1)$.
\end{theorem}
\end{mdframed}
\begin{mdframed}[leftmargin=1mm,
    skipabove=2mm, 
    skipbelow=-1mm, 
    backgroundcolor=gray!10,
    linewidth=0pt,
    leftmargin=-1mm,
    rightmargin=-1mm,
    innerleftmargin=2mm,
    innerrightmargin=2mm,
    innertopmargin=2mm,
innerbottommargin=1mm]
\begin{theorem}\label{thm:one-hid-cnn-hf}
The rank of the functional Hessian is bounded as:
$\operatorname{rank}\left(\HF\right) \leq 2\min\big(k_1, K(d_0-k_1+1)\big)\, m_1\,.$
\end{theorem}
\end{mdframed}

The strategy behind the proofs (section~\ref{sec:cnntightproofs}) is to write the CNN as a superposition of the functions that act on different input patches. Now, we must also utilize the form of Toeplitz derivatives and involved auxiliary permutation matrices. In terms of the results, 
interestingly, we find that now the filter size $k_1$ has entered inside the $\min$ terms in each of the bounds; thereby further reducing the rank. However, as shown ahead, for larger networks with many channels $m$, we find that our earlier upper bound still fares decently. 
\section{Empirical Verification}

\paragraph{Verification of upper bounds.} 
We empirically validate our upper bounds in a variety of settings, in particular, with both linear and ReLU activations, MSE and CE loss, as well as on datasets such as CIFAR10, FashionMNIST, MNIST, and a synthetic dataset. However, given the page constraints, we only show a selection of the plots, while the rest can be found in the Appendix~\ref{sec:figures}. Following~\cite{singh2021analytic}, to rigorously illustrate the match with our bounds, we compute the exact Hessians, without approximations, and in Float64 precision. These precautions are taken to avoid any imprecision in calculating the rank, since the boundary of non-zero eigenvalues with zero can be otherwise a bit blurry.

\paragraph{Rank vs number of channels.} We begin by illustrating the trend of our general upper bounds in Figure~\ref{fig:mylabel} for the case of 2-hidden layer CNN on CIFAR10 and is thus the counterpart to the Figure~\ref{fig:teaser} presented before. The upper bound is relatively close to the true rank and similarly shows a linear trend with the number of channels.

\paragraph{Rank vs Filter Size.} Next, in Figure~\ref{fig:linfil}, we demonstrate the match of our exact bound with the rank as observed empirically. We hold the number of channels as fixed and vary the filter size. First of all, here the markers and the lines indeed coincide for the functional Hessian and outer-product Hessian. On the other hand, the loss Hessian which we bound as the sum of the ranks of $\HO$ and $\HF$ forms a canopy over all the empirically observed values of the rank across the filter sizes. The upper bound itself is also quite close; it becomes even closer for large values of $m$, see Section~\ref{supp:fil-lin-mse}). 

\vspace{-1em}
\paragraph{The case of non-linearities.} Further, in a similar comparison across filter sizes, we showcase that our bounds which were derived for linear activations still remain as valid upper bounds and form a canopy as seen in Figure~\ref{fig:relfil}.

\section{Locality and Weight-sharing}\label{sec:lcn+ws}

We would like to do a little deep dive and explore a simplified setting, in order to gather some intuition. 
We study the two crucial components behind CNN, namely, local connectivity (i.e, the localized patches of the input are convolved with a possibly distinct filter) and weight sharing (where the same filter is applied to each of these patches). 

\textbf{Setting.}
Concretely, let us begin by considering the case of a one-hidden layer neural network (with linear activations for simplicity). We now try to get rid of the layer indices for the sake of clarity. So the input dimension is $d$, output dimension is $K$, filter size is $k$, number of hidden layer filters is $m$. Further, we consider non-overlapping filters, like in MLPMixer~\cite{tolstikhin2021mlp}. In other words, the stride is set equal to the filter size $k$ and so the number of patches under consideration will be $t=d/k$, and to avoid unnecessary complexity, we assume $d$ is an integer multiple of $k$.\looseness=-1

\paragraph{Locally Connected Networks (LCNs).} 
As the filters that get applied on the patches are distinct, let us denote them by $\wmat{ii} \in \mathbb{R}^{m\times k}$ and the output layer matrix as $\Vm:=\begin{pmatrix}
    \vmat{1} & \cdots  & \vmat{t}
\end{pmatrix}$, with $\vmat{i} \in \mathbb{R}^{K\times m}$, where the superscript denotes the index of the local patch upon which they get applied. Mathematically we can represent the resulting neural network function as (where $\x\up{i}\in\mathbb{R}^k$ denotes the $i$-th chunk of the input):
\begin{align*}
\small
\begin{pmatrix}
    \vmat{1} & \cdots & \vmat{t} 
\end{pmatrix} \, \begin{pmatrix}
    \wmat{11} & \cdots & \bm{0} \\
    \vdots & \ddots & \vdots \\[1mm]
     \bm{0} & \cdots & \wmat{tt} \\
\end{pmatrix} \, \begin{pmatrix}
    \x\up{1} \\
    \vdots \\[1mm]
    \x\up{t}
\end{pmatrix}
\end{align*}
    \vspace{-1em}

We indexed the local weight matrices of the first layer as $\wmat{ii}$ to reflect that it is $(i,i)$-th block on the diagonal. After carrying out the block matrix multiplication, the above formulation can also be expressed as:
\begin{align}
	\vspace{-3mm}
	\label{eq:lcn2}\Fn^{\text{LCN}}(\x) = \sum\limits_{i=1}^t \vmat{i}\,\wmat{ii}\,\x\up{i}\,.\end{align}
We can then easily bring to our minds the case of fully-connected networks which will also contain the off-diagonal blocks and would be represented as:\vspace{-1mm}
$$\Fn^{\text{FCN-large}}(\x) = \sum\limits_{i,j=1}^t \vmat{i}\,\wmat{ij}\,\x\up{j}\,.$$ Clearly, fully-connected networks form a generalization of locally-connected networks. 
Another point of comparison for LCNs in Eq.~\eqref{eq:lcn2} with FCNs is that the former may be viewed as a superposition of $s$ distinct \textit{smaller} FCNs acting on disjoin patches of the input.\footnote{This superposition point of view would hold even if we had a non-linearity, and so do our empirical results.} $$\Fn^{\text{LCN}}(\x) = \sum\limits_{i=1}^t \, \Fn^{\text{FCN-small}}\big(\x\up{i}\,;\, \lbrace\vmat{i}, \wmat{ii}\rbrace\big)\,.$$

In this scenario of LCNs, we get the following bounds on the rank of the outer-product and functional Hessian:

\begin{mdframed}[leftmargin=1mm,
    skipabove=2mm, 
    skipbelow=-1mm, 
    backgroundcolor=gray!10,
    linewidth=0pt,
    leftmargin=-1mm,
    rightmargin=-1mm,
    innerleftmargin=2mm,
    innerrightmargin=2mm,
    innertopmargin=2mm,
innerbottommargin=1mm]
\begin{theorem}\label{theorem:lcn-ho-hf} For the locally connected network as described in Eqn.~\eqref{eq:lcn2}, the rank of the outer-product and the functional Hessian can be upper bounded as follows:
$\rank(\HO^{\text{LCN}}) \leq t\cdot q(k+K-q)\,,$ and $\rank(\HF^{\text{LCN}}) \leq t\cdot 2m\min(k, K)$, where $q=\min(k, K, m)$ and $t=d/k$.
\end{theorem}
\end{mdframed}
The proof is located in Section~\ref{sec:lcn-proofs}. The neat thing about the above rank expressions is that they are identical to that obtained for the smaller fully-connected network, except where we change the input dimension  from $d\rightarrow k$ and scale the bounds by a factor of $t=d/k$ (the number of smaller fully-connected that are being superpositioned). In short, even though these weight matrices must `act' together in the loss, their contributions to the Hessian came out individually (i.e, $\HF$ and $\HO$ can be factorized as block diagonals).

\paragraph{Incorporating Weight Sharing (WS).} Now all the weight matrices in the first layer are shared, i.e., $\wmat{ii}=\Wm$.
\begin{align}\label{eq:lcn+ws}
\small
 \begin{pmatrix}
    \vmat{1}& \cdots & \vmat{t} 
\end{pmatrix} \, \begin{pmatrix}
    \Wm  & \cdots & \bm{0} \\
\vdots & \ddots & \vdots \\[1mm]
 \bm{0} & \cdots & \Wm \\
\end{pmatrix} \, \begin{pmatrix}
    \x\up{1} \\
    \vdots \\[1mm]
    \x\up{t}
\end{pmatrix}
\end{align}

\begin{mdframed}[leftmargin=1mm,
    skipabove=2mm, 
    skipbelow=-1mm, 
    backgroundcolor=gray!10,
    linewidth=0pt,
    leftmargin=-1mm,
    rightmargin=-1mm,
    innerleftmargin=2mm,
    innerrightmargin=2mm,
    innertopmargin=2mm,
innerbottommargin=1mm]
\begin{theorem}\label{theorem:cnn1hidden-ho-hf} For the locally connected network with weight sharing defined in Eqn.~\eqref{eq:lcn+ws}, the rank of the outer-product and the functional Hessian can be bounded as:
$\rank(\HO^{\text{LCN+WS}}) \leq q(k+Kt-q)\,,$ and $\rank(\HF^{\text{LCN+WS}}) \leq  2m\min(k, Kt)$, where $q=\min(k, Kt, m)$ and $t=d/k$.
\end{theorem}
\end{mdframed}
The proof can be found in Section~\ref{sec:lcn+ws-proofs}, but the intuition is that the weight matrix $\Wm$ is shared across the $\vmat{i}$, resulting in the intersection of their column space inside the Hessian. Hence, the rank shrinks for both $\HO$, $\HF$. Comparing the $\rank(\HF)$ bounds, we see that here $t$ slides inside the minimum, but only in the second term (i.e., $K$).

\begin{figure}[!h]
    \centering
    \includegraphics[width=0.4\textwidth]{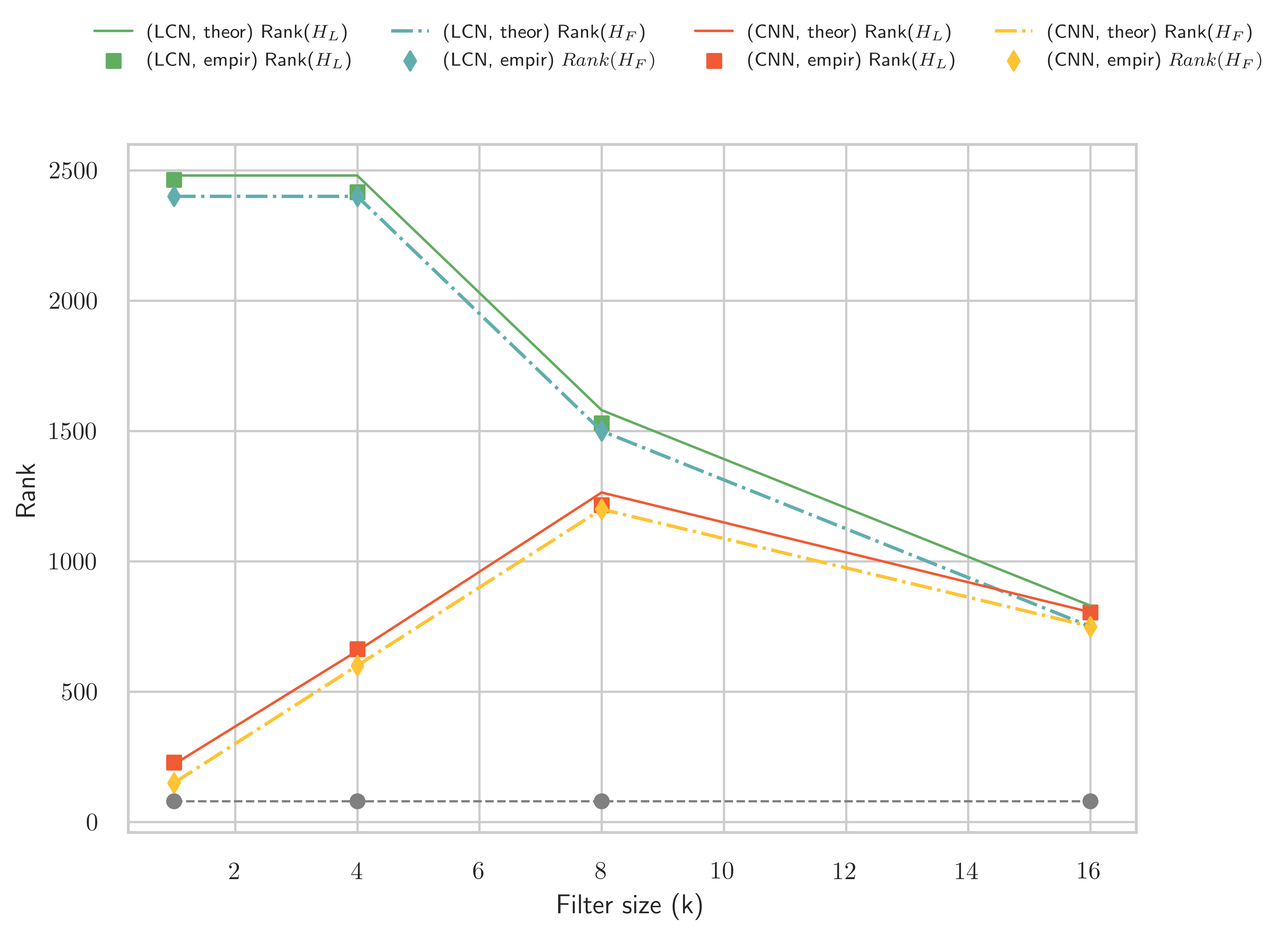}
    \caption{\vspace{-3mm}\textit{Hessian rank: \textcolor{ranklcn}{LCN} vs \textcolor{rankhf}{CNN}, (Linear, CIFAR10).}}
    \label{fig:lcn-cnn-comp}
\end{figure}

Lastly, in Figure~\ref{fig:lcn-cnn-comp}, we present an illustration of the trend of the rank with increasing filter size for both LCN and the LCN + WS variant. Besides, the interesting scaling behaviour, this figure also validates our theoretical bounds.

\vspace{-1mm}

\section{Conclusion}
All in all,  we have illustrated how the key ingredients of CNNs, such as local connectivity and weight sharing, as well as architectural aspects like filter size, strides, and number of channels get manifested through the Hessian structure and rank. Moreover, we can utilize our Toeplitz representation framework to deliver tight upper bounds in the general case of deep convolutional networks and generalize the recent finding of~\cite{singh2021analytic} about square root growth of rank relative to parameter count for CNNs as well. 
Looking ahead, our work raises some very interesting questions: (a) Is the growth of rank as a square root in terms of the number of parameters a universal characteristic of all deep architectures? Including Transformers? Or are there some exceptions to it? (b) Given the uncovered structure of the Hessian in CNNs, there are also questions about understanding which parts of the architecture affect the spectrum more --- normalization or pooling layers? 
(c) On the application side, it would be interesting to see if we can use our results to understand the approximation quality of existing pre-conditioners such as K-FAC or come up with better ones, given the rich properties of Toeplitz matrices.

\section*{Acknowledgements}
We would like to thank Max Daniels and Gregor Bachmann for their suggestions as well as rest of DALab for useful comments and support. Sidak Pal Singh
would also like to acknowledge the financial support from Max Planck ETH Center for Learning
Systems and the travel support from ELISE (GA no 951847).

\bibliography{example_paper}
\bibliographystyle{icml2023}

\newpage
\appendix
\onecolumn





\section{Tools}
\subsection{Toeplitz Derivative}\label{sec:toeplitz-proof}
\begin{replemma}{lemma:toep-grad}
	The matrix derivative of  $\toepMat{(l)}$ with respect to  $\wMsup{l}$,  is given as follows:
	\[\dfrac{\partial \toepMat{(l)} }{\partial \wMsup{l}}:=\dfrac{\partial \vect_r \toepMat{(l)}}{\partial \left(\vect_r \wMsup{l}\right)^\top } = \Im_{m_l}  \, \kro\, \qmat{l}\,,\]
	which lives in $\Reals{m_l \,m_{l-1} \, d_\ell\, d_{l-1} \times m_l \,m_{l-1}\, k_{l}}$, and the matrix $\qmat{l}\in\Reals{m_{l-1} d_l d_{l-1}\times m_{l-1} k_{l}}$ is defined as:
    \[\qmat{l} := \begin{pmatrix}
        \Im_{m_{l-1}} \kro \left(\pi_R^{0} \, \Im_{d_{l-1}\times k_{l}}\right) \\[1mm]
        \vdots \\[1mm]
        \Im_{m_{l-1}} \kro \left(\pi_R^{d_{l-1} - k_l} \,  \Im_{d_{l-1}\times k_{l}}\right) \\[1mm]
    \end{pmatrix}\]
  where $\pi_R$ is the permutation matrix performs clockwise rotation of the rows of the matrix it is left-multiplied with, and its superscript the matrix power (so, $\pi_R^0=\Im_{d_{l-1}})$. Also, in the above expression $\Im_{d_{l-1}\times k_l}$ is the `tall' identity matrix, i.e., $[\Im_{k_l} , \bm{0}]^\top$ (as $k_l\leq d_{l-1}$).
\end{replemma}
\begin{proof}
It is quite clear that the non-zero parts in the derivative will arise only when compute the derivative of Toeplitz of one row with respect to the elements of the same row. In other words, only when considering: $$\dfrac{\partial \toepMat{\convfibre{W}{l}{i,j}}}{\partial \convfibre{W}{l}{r,s}}\quad\quad \text{for}\; i=r, \;\; j=s \,.  $$ And rest of the blocks will be zeros. This explains the occurrence of two Kronecker products, with respect to $\Im_{m_l}$ and $\Im_{m_{l-1}}$. 

More concretely, recall that:
$${\toepMat{(l)}:=\begin{pmatrix}
 \toepMat{\convfibre{W}{l}{1,1}} & \cdots & \toepMat{\convfibre{W}{l}{1, m_{l-1} }}\\[1mm]
 \vdots & & \vdots \\[1mm]
 \toepMat{\convfibre{W}{l}{m_l, 1 }} & \cdots & \toepMat{\convfibre{W}{l}{m_l, m_{l-1} }}  \\[1mm]
\end{pmatrix}\,,}\quad \text{and}$$

\begin{equation}
    \wmat{l}:=\mat{\conv{W}{l}} := \begin{pmatrix}
        {\convfibre{W}{l}{1, 1}} & \cdots 
 &{\convfibre{W}{l}{m_l, 1} } \\
        \vdots &  & \vdots \\
        {\convfibre{W}{l}{1, m_{l-1}}} & \cdots &  {\convfibre{W}{l}{m_l, m_{l-1}}} \\
    \end{pmatrix}^\top = \begin{pmatrix}
        w_{1, 1, 1} & \cdots & w_{1, 1, k_l} & \cdots & w_{1, m_{l-1}, 1} & \cdots  & w_{1, m_{l-1}, k_l}\\
        \vdots &  & \vdots & & \vdots  & & \vdots \\
        w_{m_l, 1, 1} & \cdots & w_{m_l, 1, k_l} & \cdots & w_{m_l, m_{l-1}, 1} & \cdots  & w_{m_l, m_{l-1}, k_l}\\
    \end{pmatrix}\,.
\end{equation}
Let's use the shorthand $\tmat{l}_{(i, \bullet)} := \begin{pmatrix}
    \toepMat{\convfibre{W}{l}{i,1}} & \cdots & \toepMat{\convfibre{W}{l}{i,m_{l-1}}}\\
\end{pmatrix}$. Now, the general structure of the required Toeplitz derivative has the following form:

\NiceMatrixOptions{code-for-first-row = \color{gray},
                   code-for-first-col = \color{gray},}
\begin{align}\label{eq:startA}
\dfrac{\partial \toepMat{(l)} }{\partial \wMsup{l}} &= 
\begin{pNiceArray}[first-row,first-col,nullify-dots]{ccc}
        & \vect_r \matrow{\Wm}{1}\up{l} & \cdots & \vect_r \matrow{\Wm}{m_l}\up{l}\\[2mm]
\vect_r \tmat{l}_{(1, \bullet)}& \qmat{l}   & \cdots &\bm{0} \\[1mm]
\Vdots & \vdots &\ddots & \vdots  \\[1mm]
\vect_r \tmat{l}_{(m_l, \bullet)} & \bm{0} & \cdots & \qmat{l}\\[2mm]
\end{pNiceArray} = \Im_{m_l} \kro \qmat{l}.
\end{align}
The diagonal blocks in the above matrix are identical, for if the same weight is located in the Toeplitz representation with respect to which we differentiate, we would get a $1$ else a $0$. Besides, it should also be noted that $\qmat{l}$ is just a binary matrix (i.e., contains either $0$ or $1$), with atmost a single $1$ per row. 

Next, each of the $d_l = d_{l-1}- k_l +1$ rows of $\tmat{l}_{(i, \bullet)}$ is a clockwise rotation of its first row. Thus the derivative of each of its row will depend upon the amount of clockwise rotation shift, and to get the derivative of the entire row block $\tmat{l}_{(i, \bullet)}$ we will just need to stack rowwise each of these obtained derivatives with respect to $\matrow{\Wm}{i}\up{l}$. Since, all the diagonal blocks are identical, without loss of generality, let's assume $i=1$ and consider the derivative of the \textit{first row} of $\tmat{l}_{(1, \bullet)}$ with respect to $\matrow{\Wm}{1}\up{l}$. 

The derivative will be zero whenever we try to a differentiate the Toeplitz representation of one input channel with respect to parameters of a different input channel. Hence, the form of this derivative will be $\Im_{m_{l-1}} \kro \Am$ for some matrix $\Am\in\Reals{d_{l-1}\times k_l}$. For the derivative of the first row, $\Am=\pi_R^{0}\Im_{d_{l-1}\times k_l}$, while for the derivative of the $j$-th row it will be $\Am=\pi_R^{j-1}\Im_{d_{l-1}\times k_l}$, for $j\in[1,\cdots, d_{l}]$. Here, by $\pi_R^j$ we mean taking the $j$-th matrix power of the following matrix, $\pi_R$ (which performs clockwise rotation of the rows of matrix it left-multiplies):
$$\pi_R=\begin{pmatrix}
     0 &  \cdots &  & 0 & 1 \\
     1 & 0 & \cdots  & & 0 \\
      0 & 1 & 0 & \cdots & 0\\
      \vdots &  & \ddots & &  \vdots \\
      0 & \cdots & 0  & 1 & 0  \\[1mm]
\end{pmatrix}\,,$$
 and $\Im_{m\times n}\in\Reals{m\times n}$, for $m>n$, denotes the tall identity matrix $[\Im_{n\times n} , \bm{0}_{(m-n)\times n}]^\top$, which is padded by $m-n$ rows of all zeros. This tall identity matrix gets post-multiplied to the clockwise rotation matrix, since we only require the first $k_l$ columns of it.

Therefore, the form of $\qmat{l}$ is given as follows:
 \[\qmat{l} := \begin{pmatrix}
        \Im_{m_{l-1}} \kro \left(\pi_R^{0} \, \Im_{d_{l-1}\times k_{l}}\right) \\[1mm]
        \vdots \\[1mm]
        \Im_{m_{l-1}} \kro \left(\pi_R^{d_{l-1} - k_l} \,  \Im_{d_{l-1}\times k_{l}}\right) \\[1mm]
    \end{pmatrix}\]

 \end{proof}

\paragraph{Remark.} Note that the above matrix is equivalent, upto row permutations, to the following more concisely formulated matrix:

$$\Im_{m_l}\kro \qmat{l} \equiv \Im_{m_{l} m_{l-1}} \, \kro\, \pi_R^{0\,\downarrow\, d_{l-1}-k_{l}}\, \Im_{d_{l-1} \times k_{l}}\,,$$
where, $ \pi_R^{0\,\downarrow\, d_{l-1}-k_l}:=[\pi_C^0,\, \pi_C^1,\, \cdots,  \pi_C^{d_{l-1}-k_l}]^\top$ and similar to $\pi_R$,  $\pi_C := \pi_R^\top$ performs clockwise rotation of the columns of the matrix it right-multiplies.
\subsection{Technical Background}
\subsubsection{Helper Lemmas}
\begin{lemma}
\label{lemma:supp-kronecker-block}

Let $\Xm\in\mathbb{R}^{m\times n} $ and $\Ym\in\mathbb{R}^{p\times q}$. Then the row-partitioned matrix $\left[\begin{array}{c}\Im_q \otimes \Xm \\ \Ym \otimes \Im_n\end{array}\right]$ has the rank, 
\[
\rank\left[\begin{array}{c}\Im_q \otimes \Xm \\ \Ym \otimes \Im_n\end{array}\right]=q \rank(\Xm)+n \rank(\Ym)-\rank(\Xm) \rank(\Ym)
\]
\end{lemma}
We defer the reader to~\citet{CHUAI2004129,singh2021analytic} for the proof of this lemma.

\subsubsection{Auxiliary Matrices}\label{sec:aux-mat}
To get tight bounds for convolutional networks, one has to `play' around with permutation matrices. So, to this end, let us introduce the auxiliary matrices from~\cite{singh1972some,tracy1969multivariate}, denoted\footnote{We denote the matrix by $\Pm$ instead of $\Im$ as in the original paper, to avoid confusion with widespread occurrences of the identity matrix in our analysis.} by $\Pm_{(m)}\in\Reals{mn\times mn}$ and defined as a rearrangement of the identity $\Im_{mn}$ by taking every $m$-th row from the first, then every $m$-th row from the second etc. Likewise, one can define an rearrangement of the identity by taking every $m$-th column from the first, and so on, and this will be denoted by $\Pm\up{m}$. The following are some of the essential properties of this auxiliary matrix:
\begin{itemize}
    \item $\Pm_{(m)} = \Pm_{(n)}^\top$
    \item $\Pm_{(m)} \Pm_{(n)} = \Pm_{(n)} \Pm_{(m)} =  \Im_{mn}$
    \item $\Pm_{(m)} = \Pm\up{n}$, $\Pm_{(n)} = \Pm\up{m}$
\end{itemize}
In other contexts, this is also known as the commutation matrix~\cite{magnus2019matrix}. In particular $K_{(m,n)}\in\Reals{mn\times mn}$ is a matrix partitioned into $m\times n$ blocks such that $ij$-th block has $ji$-th entry $1$ and rest all $0$. One can check that $\Pm_{(m)} = \Km_{(n, m)}$. For ease of understanding, we will use auxiliary matrices $\Pm_{(m)}$, but if needed we rely upon the extensive results obtained for it under the name of commutation matrices. 
\subsubsection{Frequently used properties of Kronecker Products}
Often in the analysis, we use several properties of Kronecker products, and we cannot afford to reiterate them at every step of the proof. Hence, if a reader feels a bit uneasy about a particular step, we recommend they consult some of the properties mentioned here. The proofs are readily available online, or check~\cite{magnus2019matrix} as a good reference for matrix analysis.

For conformable matrices $\Am,\Bm,\Cm,\Dm,\Xm$, we have that:
\begin{itemize}
    \item $(\Am\kro\Bm) (\Cm\kro\Dm) = \Am\Bm\kro \Cm\Dm$
    \item $\rank(\Am\kro\Bm) = \rank(\Am)\cdot \rank(\Bm)$
    \item $(\Am\kro\Bm)\kro\Cm = \Am \kro (\Bm\kro\Cm)$
    \item $(\Am\kro\Bm)^\top =\Am^\top \kro \Bm^\top$
    \item $\vect_r(\Am\Xm\Bm) = (\Am\kro\Bm^\top)\vect_r(\Xm)$
\end{itemize}

For column vectors $\a,\b$, we have that: $\a\kro\b =\vect_r(\a\b^\top)$.

\section{CNN Hessian Structure}\label{sec:struct}
\subsection{Outer-Product Hessian}
In particular, since $\nabla^2_{\Fn} \,\ell_i=\Im_K\,,\;\forall i \in [n]$, the choice of MSE implies that the outer product Hessian simplifies to:
$$ \HO =  \frac{1}{n}\sum_{i=1}^n \nabla_{\btheta} \Fn(\x_i) \, \nabla_\btheta \Fn(\x_i)^\top$$

\begin{repproposition}{prop:ho}
The $kl$-th block of the outer-product Hessian is given by, 
\begin{align}\label{eq:ho-kl}
    \hspace{-4em}\HOsup{(kl)} &={\tildeqmatt{k}}
    \bigg(\tmat{k+1:L+1} \tmat{L+1:l+1} \,\,\kro \notag\\
    &\hspace{5em} \tmat{k-1:1}\covx \tmat{1:l-1}\bigg)\tildeqmat{l}\,,
\end{align}
where, $\widetilde{\Qm}^{(l)}=\Im_{m_l}\kro\qmat{l}$.
\end{repproposition}
\begin{proof}
\begin{align}
\dfrac{\partial \Fn(\x)}{\partial \Wm^{(l)}} &= \left(\tmat{L+1: l+1} \kro \x^\top\tmat{1:l-1}\right)\dfrac{\partial \tmat{l}}{\partial \Wm^{(l)}} \\
&=\left(\tmat{L+1: l+1} \kro \x^\top\tmat{1:l-1}\right)(\Im_{m_l} \kro \qmat{l})
\end{align}

Note, we cannot use mixed product property of Kronecker product as the shapes are not conformable. Further the above computation is in the Jacobian format (numerator layout), and to compute the outer-product Hessian we need to transpose in order to obtain it in the gradient format.

\begin{align}
\label{eq:grad-fn-wl}
\nabla_{\Wm^{(l)}}\, {\Fn(\x)} &=(\Im_{m_l} \kro {\qmat{l}}^\top)\left(\tmat{l+1:L+1} \kro \tmat{l-1:1}\x\right)
\end{align}
Next, we compute the outer-product of the above and average over the input, resulting in the $l$-th block in the $\HO$ diagonal.

\begin{align*}
    \hspace{-1em}\HOsup{(ll)} &:=  \frac{1}{n}\sum_{i=1}^n \nabla_{\Wm^{(l)}}\, {\Fn(\x_i)} \,\cdot  \, {\nabla_{\Wm^{(l)}}\, {\Fn(\x_i)}}^\top \\
    &\hspace{-2em}= (\Im_{m_l} \kro {\qmat{l}}^\top) \bigg(\tmat{l+1:L+1} \tmat{L+1:l+1} \, \, \kro \\
    &\hspace{5em} \tmat{l-1:1}\covx \tmat{1:l-1}\bigg) (\Im_{m_l} \kro \qmat{l})
\end{align*}
Let's use the shorthand $\widetilde{\Qm}^{(l)}=\Im_{m_l}\kro\qmat{l}$ to make the expressions more compact.
Similarly, we can this computation to obtain the $kl$-th block of the outer-product Hessian,
\begin{align}\label{eq:ho-kl}
    \hspace{-4em}\HOsup{(kl)} &={\tildeqmatt{k}}
    \bigg(\tmat{k+1:L+1} \tmat{L+1:l+1} \,\,\kro \notag\\
    &\hspace{5em} \tmat{k-1:1}\covx \tmat{1:l-1}\bigg)\tildeqmat{l}
\end{align}
\end{proof}

\subsection{Functional Hessian}
In our functional Hessian calculations, we consider the second derivative wrt the transpose of that matrix. This is not a problem for us as rank is invariant to row or column permutations.
\begin{repproposition}{prop:hf}
    For $k\leq l$, the $kl$-th block of $\HF$ is given by:
\begin{align}\label{eq:hf-lk}
  \HFsup{(kl)} &= \left(\Im_{m_k} \kro \qmatt{k}  \right)\bigg(\tmat{k+1:l-1} \, \, \kro \notag\\
& \hspace{3em}\tmat{k-1:1}\Om^\top\tmat{L+1:l+1} \bigg) \left(\qmat{l} \kro \Im_{m_l}\right)\,,
\end{align}
    While, for $k\geq l$, it is:
    \begin{align}\label{eq:hf-kl-smaller}
  \HFsup{(kl)} &= \left(\Im_{m_k} \kro \qmatt{k}  \right)\bigg(\tmat{k+1:L+1}\Om \tmat{1:l-1} \, \, \kro \notag\\
& \hspace{4em}\tmat{k-1:l+1} \bigg) \left(\qmat{l} \kro \Im_{m_l}\right)\,.
\end{align}
\end{repproposition}
\begin{proof}
This requires more care. First, let us recall its form:
\[\HF =\frac{1}{n}\sum_{i=1}^n \,
    \sum_{c=1}^K [\nabla_{\Fn}\ell_i]_c\, \nabla^2_\btheta \, \Fn^{c}(\x_i)\]
The gradient of the loss with respect to the function output, in the case of MSE loss, is just the residual. And so we will denote $\nabla_{\Fn}\ell_i =  \hat\y_i-\y_i=: \pmb \delta_{\x_i,\,\y_i}$ hereafter.

To begin, we need to compute the Hessian of the network function at the $c$-th index, i.e., $\nabla^2_\btheta \, \Fn^{c}(\x)$. We already have the gradient (for all $K$ outputs) with respect to $\wmat{l}$, the matricized convolutional tensor at layer $l$, in Eqn.~\eqref{eq:grad-fn-wl}. The gradient with respect to only the output at the $c$-th index is given by:

\begin{align}
    \nabla_{\Wm^{(l)}}\, {\Fn^c(\x)} &= \tildeqmatt{l} \left(\tmat{l+1:L+1} \kro \tmat{l-1:1}\x\right) \e_c\notag\\
    &= \tildeqmatt{l} \left(\tmat{l+1:L+1}\e_c \kro \tmat{l-1:1}\x\right) \,.
\end{align}
Here, $\e_c\in\Reals{K}$ denotes the $c$-th canonical basis vector. In the second line, we used the mixed-product property of Kronecker products. Now, we need to compute another derivative of the above expression with respect to the other layer matrices. We can clearly see that the block-diagonal terms in the Functional Hessian will be zero, since there is no more $\tmat{l}$ or $\wmat{l}$ left in the above expression.

Consider, we take the derivative with respect to layer $k$. We will have two cases depending upon whether $k<l$ or $k>l$. Before doing that, let's first rewrite the gradient expression and perform the sum  over the inputs and targets, as shown below:

\begin{align}
    \hspace{-1em}\nabla_{\Wm^{(l)}}\, {\Fn^c(\x)} &= \tildeqmatt{l} \vect_r\left(\tmat{l+1:L+1}\e_c \x^\top \tmat{1:l-1}\right) \notag\\
    &\hspace{-6em} = \tildeqmatt{l} \left(\tmat{l+1:L+1}\e_c \x^\top \kro \Im_{m_{l-1} d_{l-1}} \right)\vect_r\left(\tmat{1:l-1}\right) \label{eq:func-grad-smaller}\\
    &\hspace{-6em}= \tildeqmatt{l} \left(\Im_{m_l d_l} \kro\tmat{l-1:1} \x \e_c^\top\right)\vect_r\left(\tmat{l+1:L+1}\right) \,. \label{eq:func-grad-bigger}
\end{align}

Here, we first used the fact the Kronecker product of two vectors $\a$ and $\b$ can be written as vectorization of their outer-product, namely, $\a\kro\b =\vect_r\left(\a \b^\top\right)$. In the second line, we use the identity $$\vect_r(\Am\Xm\Bm) = \left(\Am\kro \Bm^\top\right) \vect_r(\Xm)\,,$$ for $\Am \in \mathbb{R}^{m\times n}\,,\, \Xm \in \mathbb{R}^{n\times p}\,,\, \Bm \in \mathbb{R}^{p\times q}$.
\subsubsection{When $k<l$}
In this case, we will use the form of the gradient in Eqn.~\eqref{eq:func-grad-smaller} to differentiate with respect to $\wmat{k}$. As a matter of fact, we will actually perform the derivatives with respect to ${\wmat{k}}^\top$ to avoid keeping track of the special kind of permutation matrices known as commutation matrices. This will not restrain the analysis since the rank of a matrix is invariant to both row and column permutations.  

Let's focus on taking the derivative with respect to $\vect_r\tmat{1:l-1}$ since that's the only term which will depend on $\wmat{k}$.
\begin{align}
    & \hspace{-2em}\dfrac{\partial }{\partial \wmatt{k}} \vect_r \tmat{1:l-1}\\
    & \hspace{-2em}= \dfrac{\partial }{\partial \wmatt{k}} \, \vect_r \tmat{1:k-1} \tmatt{k} \tmat{k+1:l-1}\\
    &\hspace{-2em}= \left(\tmat{1:k-1}\kro \tmat{l-1:k+1}\right) \dfrac{\partial }{\partial \wmatt{k}} \vect_r \tmatt{k}
\end{align}

We can exploit Lemma~\ref{lemma:toep-grad} and equivalently write $\tmat{k}\in\Reals{m_k d_k \times m_{k-1} d_{k-1}}$ in terms of $\wmat{k} \in \Reals{m_k \times m_{k-1} k_{k}}$ and $\qmat{k} \in\Reals{m_{k-1} d_k d_{k-1} \times m_{k-1} k_{k}}$, as shown below:

\begin{align}
    \vect_r\tmat{k} &= (\Im_{m_k} \kro \qmat{k}) \vect_r \wmat{k}\\
    \text{or,}\quad \vect_r\tmat{k} &= \vect_r\wmat{k}\qmatt{k}
\end{align}

Thus we can write $\dfrac{\partial }{\partial \wmatt{k}} \vect_r \tmatt{k}$ as:

\begin{align}
    \dfrac{\partial }{\partial \wmatt{k}} \vect_r \tmatt{k} &= \dfrac{\partial }{\partial \wmatt{k}} \vect_r \qmat{k}\wmatt{k} \\
    &\hspace{-3em}=\left(\qmat{k}\kro \Im_{m_k}\right) \dfrac{\partial \vect_r \wmatt{k}}{\partial \wmatt{k}}\\
    &\hspace{-3em}=\qmat{k}\kro \Im_{m_k}
\end{align}

Finally, we can complete the overall expression of derivative of the gradient with respect to layer $k$:

\begin{align}
    &\hspace{-2em} \dfrac{\partial\, \nabla_{\Wm^{(l)}}\, {\Fn^c(\x)}}{\partial \wmatt{k}} \notag\\
    &\hspace{-2em}= \left(\Im_{m_l} \kro \qmatt{l}\right)\bigg(\tmat{l+1:L+1}\e_c \x^\top \tmat{1:k-1} \, \, \kro \notag\\
    & \hspace{7em}\tmat{l-1:k+1}\bigg) \left(\qmat{k} \kro \Im_{m_k}\right)\,.
\end{align}

Summing the above expression over the inputs and targets, along with scaling by the residuals, yields the $(lk)$-th block of the functional Hessian:
\begin{align}\label{eq:hf-lk}
   \hspace{-1em}\HFsup{(lk)} &= \left(\Im_{m_l} \kro \qmatt{l}\right)\bigg(\tmat{l+1:L+1}\Om\tmat{1:k-1} \, \, \kro \notag\\
    & \hspace{6em}\tmat{l-1:k+1}\bigg) \left(\qmat{k} \kro \Im_{m_k}\right)\,,
\end{align}
where, $\Om:=\E[\pmb \delta_{\x, \y} \, \x^\top]$ is the (uncentered) residual-input covariance matrix.

Similarly, the $(kl)$-th block can also be obtained by repeating the same procedure\footnote{Attention: it is not the same as the transpose of the expression in Eqn.~\eqref{eq:hf-lk}, since recall on one axis of the Hessian we are taking derivatives in the order of $\vect_r(\wmat{l})$ and on the other axis in the order $\vect_r(\wmatt{l})$, causing the asymmetry.}, resulting in:
\begin{align}\label{eq:hf-kl-smaller}
   \HFsup{(kl)} &= \left(\Im_{m_k} \kro \qmatt{k}  \right)\bigg(\tmat{k+1:l-1} \, \, \kro \notag\\
    & \hspace{3em}\tmat{k-1:1}\Om^\top\tmat{L+1:l+1} \bigg) \left(\qmat{l} \kro \Im_{m_l}\right)\,,
\end{align}


\subsubsection{When $k>l$}
We can repeat a similar calculation to obtain the following expression:
\begin{align}\label{eq:hf-kl-bigger}
   \HFsup{(kl)} &= \left(\Im_{m_k} \kro \qmatt{k}  \right)\bigg(\tmat{k+1:L+1}\Om \tmat{1:l-1} \, \, \kro \notag\\
    & \hspace{4em}\tmat{k-1:l+1} \bigg) \left(\qmat{l} \kro \Im_{m_l}\right)\,.
\end{align}

By comparing the expressions in Eqns.~\eqref{eq:hf-kl-smaller} and~\eqref{eq:hf-kl-bigger}, we get the hint to factorize out $\qmat{l}\kro\Im_{m_l}$ from the columns, and likewise $\Im_{m_l} \kro \qmatt{l}$ from the rows,  $\forall \, l \in [L+1]$. 
\paragraph{Remark.} The arrangement of the Toeplitz derivatives on the left and right hand side in the Eqn.~\eqref{eq:hf-kl-smaller} above is very similar to that in the case of Eqn.~\eqref{eq:ho-kl} for the outer-product Hessian --- except, now on the right hand side the order of Kronecker product between $\Im_{m_l}$ and $\qmat{l}$ gets changed to reflect the fact that we are taking the derivatives with respect to $\wmatt{l}$ here.

\end{proof}

\section{Omitted Proofs}
\subsection{Rank of Outer-Product Hessian}\label{sec:ho-cnn-proof}
\vspace{1em}
\begin{mdframed}[leftmargin=1mm,
    skipabove=2mm, 
    skipbelow=-1mm, 
    backgroundcolor=gray!10,
    linewidth=0pt,
    leftmargin=-1mm,
    rightmargin=-1mm,
    innerleftmargin=2mm,
    innerrightmargin=2mm,
    innertopmargin=2mm,
innerbottommargin=1mm]
\begin{reptheorem}{theorem:ub-outer} The rank of the outer-product Hessian  is upper bounded as
\begin{align*}
\rank(\HO) &\leq \min\big(p, 
d_0 \rank(\tmat{2:L+1}) + K \rank(\tmat{L:1}) \, - \\
&\hspace{5em }\rank(\tmat{2:L+1})\rank(\tmat{L:1})\big) \\
&= \min\left(p, q \,( d_0 + K  - q)\right)\,.
\end{align*}
Here, $q:=\min(d_0, m_1 d_1, \cdots, m_L d_L, K)$.
\end{reptheorem}
\end{mdframed}
\begin{proof}
By the decomposition in Proposition~\ref{eq:outer-decomp}, we have that:
    \begin{align}
    \rank(\HO)&\leq \min\left(\rank(\Am_o), \rank(\Bm_o), \rank(\Qm_o)\right)\,.
    \end{align}
    Now, $\rank(\Bm_o) = Kd_0$, $\rank(\Qm_o) \leq \min(\widehat{p}, p)$, and via Theorem 3 of \cite{singh2021analytic}, we have that $\rank(\Am_o)=q(d_0 + K - q)$ which is naturally bounded above by $\widehat{p}$.
    Hence, we obtain: 
    \begin{align*}
    \rank(\HO)&\leq \min\left(p, q(d_0 + K-q)\right)\,.
    \end{align*}
    Here, $q:=\min(d_0, m_1 d_1, \cdots, m_L d_L, K)$.
\end{proof}

\subsection{Rank of Functional Hessian}\label{sec:hf-cnn-proof}
\vspace{1em}
\begin{mdframed}[leftmargin=1mm,
    skipabove=2mm, 
    skipbelow=-1mm, 
    backgroundcolor=gray!10,
    linewidth=0pt,
    leftmargin=-1mm,
    rightmargin=-1mm,
    innerleftmargin=2mm,
    innerrightmargin=2mm,
    innertopmargin=2mm,
innerbottommargin=1mm]
\begin{reptheorem}{theorem:func-hess-cols}
For a deep linear convolutional network, the rank of $l$-th column-block, $\HFhatsup{\bullet l}$, of the matrix $\HFhat$, can be upper bounded as
 $$\rank(\HFhatsup{\bullet l}) \leq \min(\qtil\, m_{l -1} d_{l-1} + \qtil\,m_{l} d_{l}  - \qtil^{\,2}\,, m_l m_{l-1} k_l)\,,$$ 
 for $l \in [2, \cdots , L]\,.$ When $l=1$, we have 
 $$
\rank(\HFhatsup{\bullet 1}) \leq \min(\qtil\, m_{1} d_1 + \qtil \, s - \qtil^{\,2}\,, m_1 m_0 k_1)\,.$$ And, when $l=L+1$, we have $$
\rank(\HFhatsup{\bullet L+1}) \leq \min(\qtil\, m_{L} d_L + \qtil \, s - \qtil^{\,2}\,, m_{L+1}m_L k_{L+1})\,.$$
Here, $\qtil := \min(d_0, m_1 d_1, \cdots, m_{L} d_L, K, s) = \min(q, s)$ and $s:=\rank(\Om)=\rank(\E[\pmb \delta_{\x,\y}\,\x^\top])$.
\end{reptheorem}
\end{mdframed}
\begin{proof}

\begin{align*}
\small
\HFhatsup{\bullet l}&= \begin{pmatrix}
\left(\Im_{m_1}\kro \qmatt{1}\right)\left(\tmat{2:l-1} \kro \Om^\top  \tmat{L+1:l+1}\right)\left(\qmat{l} \kro \Im_{m_l}\right) \\[2mm]
\vdots \\[1mm]
\left(\Im_{m_j}\kro \qmatt{j}\right)\left(\tmat{j +1:l-1} \kro \tmat{j-1:1} \Om^\top \tmat{L+1:l+1}  \right)\left(\qmat{l} \kro \Im_{m_l}\right)\\[1mm]
\vdots \\[2mm]
\left(\Im_{m_{l-1}}\kro \qmatt{l-1}\right)\left(\Im_{m_{l-1}}\kro \tmat{l-2:1} \Om^\top  \tmat{L+1:l+1}\right)\left(\qmat{l} \kro \Im_{m_l}\right)  \\[2mm]
\bm{0}  \\[3mm]
\left(\Im_{m_{l+1}}\kro \qmatt{l+1}\right)\left(\tmat{l+2:L+1} \Om \tmat{1:l-1} \kro \Im_{m_{l}} \right)\left(\qmat{l} \kro \Im_{m_l}\right)\\[1mm]
 \vdots \\[1mm]
 \left(\Im_{m_k}\kro \qmatt{k}\right)\left(\tmat{k+1:L+1} \Om \tmat{1:l-1} \kro \tmat{k-1:l+1} \right)\left(\qmat{l} \kro \Im_{m_l}\right)\\[1mm]
\vdots \\[1mm]
\left(\Im_{m_{L+1}}\kro \qmatt{l+1}\right)\left(\Om \tmat{1:l-1} \kro \tmat{L-1:l +1} \right)\left(\qmat{l} \kro \Im_{m_l}\right)\\[1mm]
\end{pmatrix} \\
&= \Qm_F \,\underbrace{\begin{pmatrix}
\tmat{2:l-1} \kro \Om^\top  \tmat{L+1:l+1} \\[2mm]
\vdots \\[1mm]
\tmat{j +1:l-1} \kro \tmat{j-1:1} \Om^\top \tmat{L+1:l+1}  \\[1mm]
\vdots \\[2mm]
\Im_{m_{l-1}}\kro \tmat{l-2:1} \Om^\top  \tmat{L+1:l+1}  \\[2mm]
\bm{0}  \\[3mm]
\tmat{l+2:L+1} \Om \tmat{1:l-1} \kro \Im_{m_{l}} \\[1mm]
 \vdots \\[1mm]
 \tmat{k+1:L+1} \Om \tmat{1:l-1} \kro \tmat{k-1:l+1} \\[1mm]
\vdots \\[1mm]
\Om \tmat{1:l-1} \kro \tmat{L-1:l +1} \\[1mm]
\end{pmatrix}}_{\Am_F\up{l}} \left(\qmat{l} \kro \Im_{m_l}\right)
\end{align*}

In the above, the matrix $\Qm_F\in\Reals{p\times \widehat{p}}$ is the block diagonal matrix containing on its $l$-th block the matrix $\Im_{m_l} \kro \qmatt{l}$. The rank of $\HFhatsup{\bullet l}$ can then be upper-bounded as:
\begin{align*}
    \rank(\HFhatsup{\bullet l}) &\leq \min(\rank(\Qm_F), \rank(\Am_F\up{l}), \rank(\Qm\up{l}\kro\Im_{m_l}))\\
    &= \min(p, \widehat{p},\, \qtil\, m_{l -1} d_{l-1} + \qtil\,m_{l} d_{l}  - \qtil^{\,2}\,, m_l m_{l-1} k_l)\\
    &= \min(\qtil\, m_{l -1} d_{l-1} + \qtil\,m_{l} d_{l}  - \qtil^{\,2}\,, m_l m_{l-1} k_l) \,,
\end{align*}
where in the second line, we used the result of the $\rank(\Am_F\up{l})$ from~\cite{singh2021analytic}, while the $\rank(\Qm\up{l}\kro\Im_{m_l}))\leq \min(m_l m_{l-1} k_l, m_{l}, m_{l-1} d_l d_{l-1}) = m_l m_{l-1} k_l$ as $k_l\leq d_{l-1}$  for valid convolution. 

Likewise, we can carry out a similar operation for the first block-columnn $\HFhatsup{\bullet 1}$
\begin{align*}
\HFhatsup{\bullet 1}&= \begin{pmatrix}
\bm{0}  \\[2mm]
\left(\Im_{m_2}\kro \qmatt{2}\right)\left(\tmat{3:L} \Om  \kro \Im_{m_1} \right)\left(\qmat{1} \kro \Im_{m_1}\right)\\[1mm]
\vdots \\[1mm]
\left(\Im_{m_k}\kro \qmatt{k}\right)\left(\tmat{k+1:L+1} \Om \kro \tmat{k-1:2} \right)\left(\qmat{1} \kro \Im_{m_1}\right)\\[1mm]
\vdots \\[1mm]
\left(\Im_{m_{L+1}}\kro \qmatt{l+1}\right)\left(\Om \kro \tmat{L:2} \right)\left(\qmat{1} \kro \Im_{m_1}\right)\\[1mm]
\end{pmatrix} 
= \Qm_F\underbrace{\begin{pmatrix}
\bm{0}  \\[2mm]
\tmat{3:L} \Om  \kro \Im_{m_1}  \\[1mm]
\vdots \\[1mm]
\tmat{k+1:L} \Om \kro \tmat{k-1:2} \\[2mm]
\vdots \\[1mm]
\Om \kro \tmat{L:2} \\[1mm]
\end{pmatrix}}_{\Am_F\up{1}} \left(\qmat{1} \kro \Im_{m_1}\right)
\end{align*}
The rank can be then bounded as:
\begin{align*}
    \rank(\HFhatsup{\bullet 1}) &\leq \min(\rank(\Qm_F), \rank(\Am_F\up{1}), \rank(\Qm\up{1}\kro\Im_{m_1}))\\
    &= \min(p, \widehat{p},\,  \qtil\,m_{1} d_{1} + \qtil\,  s   - \qtil^{\,2}\,, m_1 m_{0} k_1)\\
    &= \min(\qtil\,m_{1} d_{1} + \qtil\,  s  - \qtil^{\,2}\,, m_1 m_{0} k_1) \,,
\end{align*}

Finally, we discuss the case of last column-block:
\begin{align*}
\HFhatsup{\bullet {L+1}}&= \begin{pmatrix}
\left(\Im_{m_1}\kro \qmatt{1}\right)\left(\tmat{2:L} \kro \Om^\top \right)\left(\qmat{L+1} \kro \Im_{m_{L+1}}\right)\\[1mm]
 \vdots \\[1mm]
\left(\Im_{m_k}\kro \qmatt{l}\right)\left(\tmat{k +1:L}\kro \tmat{k-1:1} \Om^\top   \right)\left(\qmat{L+1} \kro \Im_{m_{L+1}}\right)\\[1mm]
\vdots \\[1mm]
\left(\Im_{m_{L}}\kro \qmatt{l}\right)\left(\Im_{m_{L}} \kro \tmat{L-1:1} \Om^\top  \right)\left(\qmat{L+1} \kro \Im_{m_{L+1}}\right)\\[1mm]
 \bm{0}  \\[2mm]
\end{pmatrix}
=\Qm_F\underbrace{\begin{pmatrix}
\tmat{2:L} \kro \Om^\top \\[1mm]
 \vdots \\[1mm]
\tmat{k +1:L}\kro \tmat{k-1:1} \Om^\top   \\[2mm]
\vdots \\[1mm]
\Im_{m_{L}} \kro \tmat{L-1:1} \Om^\top  \\[1mm]
 \bm{0}  \\[2mm]
\end{pmatrix}}_{\Am_F\up{L+1}} \left(\qmat{L+1} \kro \Im_{m_{L+1}}\right)
\end{align*}
The rank can be then bounded as:
\begin{align*}
    \rank(\HFhatsup{\bullet L+1}) &\leq \min(\rank(\Qm_F), \rank(\Am_F\up{L+1}), \rank(\Qm\up{L+1}\kro\Im_{m_{L+1}}))\\
    &= \min(p, \widehat{p},\, \qtil\, m_L d_{L} + \qtil\,s  - \qtil^{\,2}\,, m_{L+1} m_{L} k_{L+1})\\
    &= \min(\qtil\, m_L d_{L} + \qtil\,s  - \qtil^{\,2}\,, m_{L+1} m_{L} k_{L+1}) \,,
\end{align*}
which completes the proof

\end{proof}
A corollary is 
\begin{corollary} \label{corollary:functional-rank-simpler}
Under the setup of Theorem~\ref{theorem:func-hess-cols}, when $\qtil = q = s$, the rank of $\HF$ can be further upper bounded as,
$\rank(\HF) \leq 2\, q\, M  - (L-1)\, q^2\,,\, \text{where} \quad M=\sum_{i=1}^{L} m_i \, d_i \,.$
\end{corollary}

\subsection{Rank results for one-hidden layer CNN case}\label{sec:cnntightproofs}
\vspace{1em}
\begin{mdframed}[leftmargin=1mm,
    skipabove=2mm, 
    skipbelow=-1mm, 
    backgroundcolor=gray!10,
    linewidth=0pt,
    leftmargin=-1mm,
    rightmargin=-1mm,
    innerleftmargin=2mm,
    innerrightmargin=2mm,
    innertopmargin=2mm,
innerbottommargin=1mm]
\begin{reptheorem}{thm:one-hid-cnn-ho}
  For a one-hidden layer 1D CNN, the rank of the outer product Hessian can be bounded as:
$\operatorname{rank}\left(\HO\right) \leq \min\big(Kd_0, q (k_1 + K(d_0-k_1+1) - q)\big)\,$, where $q=\min(k_1, K(d_0 - k_1 +1), m_1)$.
\end{reptheorem}
\end{mdframed}
\begin{proof}
Consider a one-hidden layer CNN as follows:
$$
\Fn(\x) = \tmat{2}\tmat{1} \x\,,\quad\text{with}\quad \btheta=\lbrace\conv{W}{2}, \conv{W}{1}\rbrace\,.
$$
Now, at the output layer $k_2 = d_1$ so as to project the hidden features to an output of size $K=m_2$  hence Thus, spatially we will have that $d_2=1$, and as a result $\tmat{2}=\wmat{2}$. Also, the number of input channels is simply assumed to be $m_0=1$, else we can flatten the input to obtain this. 

Remember that, in this scenario, $\tmat{1}$ amounts to:
\[\tmat{1} = \begin{pmatrix}
\toepMat{\convfibre{W}{1}{1, 1}} \\
\vdots \\
\toepMat{\convfibre{W}{1}{m_1, 1}} \\
\end{pmatrix}\,.\]
While, as $m_0=1$, $\wmat{1} = \begin{pmatrix}
    w_{1, 1, 1} & \cdots & w_{1, 1, k_1}\\
    w_{m_1, 1, 1} & \cdots & w_{m_1, 1, k_1}\\
\end{pmatrix}$ and we would like to ideally group the elements of $\wmat{1}$ inside $\tmat{1}$. We will do this by suitable row permutations using the auxiliary permutation matrices, discussed in the section~\ref{sec:aux-mat}.

Let's rewrite our network function as follows:
\begin{align*}
    \Fn(\x) &= \wmat{2} \tmat{1} \x \\
    & = \wmat{2} \, \Pm\up{d_1} \Pm_{(d_1)} \,\tmat{1} \x
\end{align*}
where $\Pm_{(d_1)}\tmat{1} = \begin{pmatrix}
    \widetilde{\Wm}\up{1} \pi_C^0 \\
    \vdots \\
    \widetilde{\Wm}\up{1} \pi_C^{d_1-1} \\
\end{pmatrix}$ and $\widetilde{\Wm}\up{1} = \begin{pmatrix}
    \wmat{1}_{m_1\times k_1} & \bm{0}_{m_1\times (d_0-k_1)}
\end{pmatrix} = \wmat{1} \Im_{k_1\times d_0}$.

Next, let's denote the matrix $\wmat{2}\Pm\up{d_1} \in\Reals{K\times m_1 d_1}$ by $\Vm = \begin{pmatrix}
    \Vm\up{1} & \cdots & \Vm\up{d_1}
\end{pmatrix}$. Using these we can represent the network function in the following superposition form:

\begin{equation}\label{eq:cnn-efs-reform}
\Fn(\x) = \sum\limits_{i=1}^{d_1} \Vm\up{i}\, \wmat{1}\,\Im_{k_1\times d_0} \,\pi_C^{i-1} \x\,.
\end{equation}

Having done this reformulation, let's get back to the business of computing the derivatives, which are listed below:

\begin{align}\label{eq:gradv-efs-reform}
    \nabla_{\vmat{i}} \Fn(\x) = \Im_K \kro \wmat{1} \Im_{k_1\times d_0}\, \pi_C^{i-1} \x
\end{align}

\begin{align}\label{eq:gradw-efs-reform}
    \nabla_{\wmat{1}} \Fn(\x) = \sum\limits_{i=1}^{d_1} \vmat{i}^\top \kro \Im_{k_1\times d_0} \, \pi_C^{i-1} \x
\end{align}

Now let's collect these individual parameter gradients to form the overall gradient, 
\begin{align*}
    \nabla_\btheta \Fn(\x) &= \begin{pmatrix}
        \Im_K \kro \wmat{1} \Im_{k_1\times d_0}\, \pi_C^{0} \x\\ 
        \vdots \\
        \Im_K \kro \wmat{1} \Im_{k_1\times d_0}\, \pi_C^{d_1-1} \x \\
        \sum\limits_{i-1}^{d_1} \vmat{i}^\top \kro \Im_{k_1\times d_0} \, \pi_C^{i-1} \x
    \end{pmatrix} \\
    &= \begin{pmatrix}
         \begin{pmatrix}
              \Im_K \kro \wmat{1} &\cdots & \bm{0}  \\
              \vdots & \ddots & \vdots \\
               \bm{0} & \cdots & \Im_K \kro \wmat{1} 
         \end{pmatrix} \quad \begin{pmatrix}
             \Im_K \kro  \Im_{k_1\times d_0}\, \pi_C^{0} \x \\
             \vdots \\
             \Im_K \kro  \Im_{k_1\times d_0}\, \pi_C^{d_1-1} \x \\
         \end{pmatrix}\\[1em]
         \begin{pmatrix}
             \vmat{1}^\top \kro \Im_k & \cdots & \vmat{d_1}^\top \kro \Im_k \\
         \end{pmatrix}\,\,\,\begin{pmatrix}
             \Im_K \kro  \Im_{k_1\times d_0}\, \pi_C^{0} \x \\
             \vdots \\
             \Im_K \kro  \Im_{k_1\times d_0}\, \pi_C^{d_1-1} \x \\
         \end{pmatrix}\\[2mm]
    \end{pmatrix} \\
    &= \underbrace{\begin{pmatrix}
        \Im_{Kd_1} \kro \wmat{1} \\[1mm]
        \widetilde{\Vm} \kro \Im_k
    \end{pmatrix}}_{\Am_o} \, \underbrace{\begin{pmatrix}
             \Im_K \kro  \Im_{k_1\times d_0}\, \pi_C^{0} \\
             \vdots \\
             \Im_K \kro  \Im_{k_1\times d_0}\, \pi_C^{d_1-1}  \\
         \end{pmatrix}}_{\Cm} \underbrace{\begin{pmatrix}
             \Im_K \kro \x
         \end{pmatrix}}_{\Bm}
\end{align*}
where, $\widetilde{\Vm} = \begin{pmatrix}
    \vmat{1}^\top & \cdots & \vmat{d_1}^\top 
\end{pmatrix}\in\Reals{m\times Kd_1}$

Finally, we can take the outer product and average them over the dataset to yield the outer-product Hessian:
\begin{align}
    \HO&= \Am_o \Cm 
    \left(\Im_K \kro \covx\right)
    \Cm^\top\Am_o^\top
\end{align}

Hence, we have that:
\begin{align*}
    \rank(\HO)&\leq\min(\rank( \Im_K\kro\covx), \rank(\Am_o), \rank(\Cm) )) \\
    &=\min(Kd_0, q(Kd_1 + k -q))
\end{align*}
where $q=\min(k, Kd_1, m)$ and using the fact that, for $\Cm\in\Reals{Kd_1k_1\times Kd}$, it has $\rank(\Cm)\leq\min(Kd_1k_1, Kd) = Kd$.
\end{proof}

\begin{mdframed}[leftmargin=1mm,
    skipabove=2mm, 
    skipbelow=-1mm, 
    backgroundcolor=gray!10,
    linewidth=0pt,
    leftmargin=-1mm,
    rightmargin=-1mm,
    innerleftmargin=2mm,
    innerrightmargin=2mm,
    innertopmargin=2mm,
innerbottommargin=1mm]
\begin{reptheorem}{thm:one-hid-cnn-hf}
  For a one-hidden layer 1D CNN, the rank of the functional Hessian obeys the following formula:
$\operatorname{rank}\left(\HF\right) \leq 2\min\big(k_1, K(d_0-k_1+1)\big)\, m_1\,.$
\end{reptheorem}
\end{mdframed}

\begin{proof}
    We will start atop the network function, eq.~\eqref{eq:cnn-efs-reform}, and parameter gradients, eq.~\eqref{eq:gradv-efs-reform}, \eqref{eq:gradw-efs-reform}, obtained after carrying out the manipulations with permutation matrices in the proof of Theorem~\ref{thm:one-hid-cnn-ho}. Let's then get the gradient with respect to the parameters at the $c$-th output index:

\begin{align}
\nabla_{\vmat{i}}\Fn^{c}(\x) &= \left(\Im_K \kro \wmat{1}\Im_{k_1\times d_0} \pi_C^{i-1} \x\right)\e_c \notag\\ 
&=  \e_c \kro \wmat{1}\Im_{k_1\times d_0} \pi_C^{i-1} \x \notag\\
&= \vect_r\left(\e_c\x^\top\pi_R^{i-1} \Im_{d_0\times k_1}\wmatt{1}\right)
\end{align}

And that with respect to $\wmat{1}$ is:
\begin{align}
\nabla_{\wmat{1}}\Fn^{c}(\x) &= \sum\limits_{i=1}^{d_1}\left(\vmat{i}^\top \kro \Im_{k_1\times d_0} \pi_C^{i-1}\x\right)\e_c \notag\\
&=\sum\limits_{i=1}^{d_1}\vmat{i}^\top \e_c \kro \Im_{k_1\times d_0} \pi_C^{i-1}\x\up{i}\notag\\
&= \sum\limits_{i=1}^{d_1}\vect_r\left(\vmat{i}^\top\e_c\x^\top\pi_R^{i-1}\Im_{d_0 \times k_1}\right)
\end{align}

Now, we perform the second differentiation with respect to transposed matrices, i.e., $\wmatt{1}$ or $\vmat{i}^\top$.
\begin{align*}
    \dfrac{\,\partial \nabla_{\vmat{i}}\Fn^{c}(\x)}{\partial \wmatt{1}} = \e_c\x^\top\pi_R^{i-1} \Im_{d_0\times k_1}\kro \Im_{m_1}\,.
\end{align*}
\begin{align*}
    \dfrac{\,\partial \nabla_{\wmat{1}}\Fn^{c}(\x)}{\partial \vmat{i}^\top} = \Im_{m_1} \kro \x\e_c^\top\pi_C^{i-1}\Im_{k_1\times d_0}\,.
\end{align*}
Let's now take the sum of the above matrices over the input as well as, with appropriate scaling by the residual, over the output indices to get the blocks of the functional Hessian. This yields,
\begin{align}
    \HFsup{(\vmat{i},\wmat{1})} = \Om \pi_R^{i-1}\Im_{d_0\times k_1} \kro \Im_{m_1} \,.
\end{align}
\begin{align}
    \HFsup{(\wmat{1},\vmat{i})} =\Im_{m_1} \kro \Om^\top\pi_C^{i-1}\Im_{k_1\kro d_0}\,.
\end{align}
where, recall $\Om = \E[\pmb \delta_{\x, \y} \,\x^\top]\in\Reals{K\times d_0}$ denotes the (uncentered) covariance of the residual with the input. 

Since the matrix is zero on the block diagonals (i.e., block hollow), we can analyze the rank of all the row blocks with respect to $\wmat{1}$ (same as the column block with respect to all the $\vmat{i}$) or the row blocks with respect to all the $\vmat{i}$ (same as the column block with respect to the $\wmat{1}$, i.e., either $\HFsup{(\wmat{1},\vmat{i})}$ or $\HFsup{(\vmat{i},\wmat{1})}$. Focussing on the latter, the column blocks with respect to $\wmat{1}$, i.e., 
\begin{align*}
\HFsup{(\bullet, \wmat{1})} &:= \begin{pmatrix}
    \HFsup{(\vmat{1},\wmat{1})} \\[2mm] \vdots\\[2mm] \HFsup{(\vmat{d_1}, \wmat{1})}
\end{pmatrix}
=\begin{pmatrix}
   \Om \pi_R^{0}\Im_{d_0\times k_1} \kro \Im_{m_1} \\[2mm]
   \vdots \\[2mm]
   \Om \pi_R^{d_1-1}\Im_{d_0\times k_1} \kro \Im_{m_1}
\end{pmatrix}
\end{align*}
The above is equivalent upto row permutations to, 
\begin{align}\widetilde{\Om} \kro\Im_{m_1} :=
\begin{pmatrix}
   \Om \pi_R^{0}\Im_{d_0\times k_1} \\[2mm]
   \vdots \\[2mm]
   \Om \pi_R^{d_1-1}\Im_{d_0\times k_1}
\end{pmatrix}\kro \Im_{m_1} 
\end{align}
where, $\widetilde{\Om}\in\Reals{Kd_1\times k_1}$
The rank of the functional Hessian thus amounts to:
\begin{align*}
    \rank(\HF) &= 2\rank\left(\widetilde{\Om}\kro\Im_{m_1}\right) \\
    &=  2 m_1 \rank(\widetilde{\Om}) \\
    &\leq 2 \min(Kd_1, k_1) m_1\,.
\end{align*} 

In the first line, the rank is scaled by $2$ to account for both the non-zero blocks present in the functional Hessian. The rest follows from the properties of Kronecker product and rank, thus completing the proof. 
    
\end{proof}

\subsection{Rank results for Locally connected network}\label{sec:lcn-proofs}
\begin{mdframed}[leftmargin=1mm,
    skipabove=2mm, 
    skipbelow=-1mm, 
    backgroundcolor=gray!10,
    linewidth=0pt,
    leftmargin=-1mm,
    rightmargin=-1mm,
    innerleftmargin=2mm,
    innerrightmargin=2mm,
    innertopmargin=2mm,
innerbottommargin=1mm]
\begin{reptheorem}{theorem:lcn-ho-hf} For the locally connected network as described in Eqn.~\eqref{eq:lcn2}, the rank of the outer-product and the functional Hessian can be upper bounded as follows:
$\rank(\HO^{\text{LCN}}) \leq t\cdot q(k+K-q)\,,$ and $\rank(\HF^{\text{LCN}}) \leq t\cdot 2m\min(k, K)$, where $q=\min(k, K, m)$ and $t=\dfrac{d}{k}$.
\end{reptheorem}
\end{mdframed}
\begin{proof}
We now proceed to showing the proofs of the rank of outer-product and functional Hessian for the LCN case. Let's recall the Eqn.~\eqref{eq:lcn2} for the LCN:
\begin{align*}\Fn^{\text{LCN}}(\x) = \sum\limits_{i=1}^t \vmat{i}\,\wmat{ii}\,\x\up{i}\,.\end{align*}
\paragraph{Outer-product Hessian.}
Let's compute the gradient of the function with respect to the matrix $\vmat{i}$:
\begin{align}\label{eq:lcn-grad-v}
\nabla_{\vmat{i}}\Fn^{\text{LCN}}(\x) = \Im_K \kro \wmat{ii} \x\up{i}   
\end{align}

And that with respect to $\wmat{ii}$ is:
\begin{align}\label{eq:lcn-grad-w}
\nabla_{\wmat{ii}}\Fn^{\text{LCN}}(\x) = \vmat{i}^\top \kro \x\up{i}   
\end{align}
So the gradient of the function output with respect to all the parameters can be arranged as follows:
\begin{align*}
\nabla_\btheta \Fn^{\text{LCN}}(\x) &=\begin{pmatrix}
\nabla_{\vmat{1}}\Fn^{\text{LCN}}(\x) \\[2mm]
\nabla_{\wmat{11}}\Fn^{\text{LCN}}(\x) \\[2mm]
\vdots \\[2mm]
\nabla_{\vmat{t}}\Fn^{\text{LCN}}(\x) \\[2mm]
\nabla_{\wmat{tt}}\Fn^{\text{LCN}}(\x) \\[1mm]
\end{pmatrix} \\
&=\begin{pmatrix}
  \Im_K \kro \wmat{11} \x\up{1} \\[2mm]
\vmat{1}^\top \kro \x\up{1}   \\[2mm]
\vdots \\[2mm]
\Im_K \kro \wmat{tt} \x\up{t}  \\[2mm]
\vmat{t}^\top \kro \x\up{t}   \\[1mm]
\end{pmatrix}
= 
\begin{pmatrix}
  \Am_o\up{1} \left(\Im_K \kro \x\up{1}\right)\\[1mm]
    \vdots \\[2mm]
\Am_o\up{t} \left(\Im_K \kro \x\up{t}\right)\end{pmatrix}\,,
\end{align*}
where, $\Am_o\up{i}:=\begin{pmatrix}
      \Im_K \kro \wmat{ii} \\[2mm]
      \vmat{i}^\top \kro \Im_d \\[2mm]
  \end{pmatrix}$. Next, we take the outer-product of the gradient above, and then average over the inputs, resulting in:
  \begin{align}\label{eq:lcn-ho-decomp}
      \HO &= \Am_o\underbrace{\begin{pmatrix}
          \Im_K \kro \mathbf{\Sigma}\up{11} & \cdots &  \Im_K \kro \mathbf{\Sigma}\up{1t} \\
          \vdots & & \vdots \\
          \Im_K \kro \mathbf{\Sigma}\up{t1}  & \cdots & \Im_K \kro \mathbf{\Sigma}\up{tt}\\
      \end{pmatrix}}_{\Bm_o} \Am_o^\top
  \end{align}

Here, $\Am_o=\begin{pmatrix}
          \Am\up{1}_o & \cdots & \bm{0} \\
          \vdots & \ddots & \vdots \\
          \bm{0} & \cdots & \Am\up{t}_o \\
      \end{pmatrix}$,  $\mathbf{\Sigma}\up{ij}:=\operatorname{cov}(\x\up{i}, \x\up{j})$.
Then we have that, 
\begin{align}
\rank(\Am_o) &= t \rank(\Am_o\up{i})\\
&\hspace{-4em}=K\rank(\wmat{ii}) + k \rank(\vmat{i}) - \rank(\wmat{ii})\rank(\vmat{i})\\
&= t\,\cdot q(K + k - q)
\end{align}
where, $q:=\min(K, k, m)$.

Now, the matrix $\Bm_o$ is equivalent, upto row and column permutations, to the matrix $\widetilde{\Bm}_o = \Im_K \kro \covx$, since
\begin{align*}
    \begin{pmatrix}
        \mathbf{\Sigma}\up{11} & \cdots  & \mathbf{\Sigma}\up{1t}\\[1mm]
        \vdots &\cdots & \vdots \\[2mm]
        \mathbf{\Sigma}\up{t1} & \cdots  & \mathbf{\Sigma}\up{tt}\\[1mm]
    \end{pmatrix}\in\Reals{d\times d} = \covx = \left(\E[x_i x_j]\right)_{ij}
\end{align*}

Therefore, rank of $\Bm_o$ can be upper bounded as $\rank(\Bm_o)\leq K \rank(\covx) \leq K d$. 
Finally, we can bound the rank of the outer-product Hessian, since the rank of a product of matrices is upper bounded by the minimum of the ranks of individual matrices. Hence,
\begin{equation}
    \rank(\HO)\leq \min\left(\rank(\Am_o), \rank(\Bm_o)\right) = t\cdot q (K+k-q)\,.
\end{equation}
\paragraph{Functional Hessian.}
We need to differentiate again the network output gradients in Eqns.~\eqref{eq:lcn-grad-v},~\eqref{eq:lcn-grad-w}. First, let's just obtain the gradient for the output at the $c$-th index:
\begin{align}
\nabla_{\vmat{i}}\Fn^{c,\text{LCN}}(\x) &= \left(\Im_K \kro \wmat{ii} \x\up{i})\right)\e_c \notag\\ 
&=  \e_c \kro \wmat{ii} \x\up{i} \notag\\
&= \vect_r\left(\e_c\x\up{i}{}^\top\wmatt{ii}\right)
\end{align}

And that with respect to $\wmat{ii}$ is:
\begin{align}
\nabla_{\wmat{ii}}\Fn^{c, \text{LCN}}(\x) &= \left(\vmat{i}^\top \kro \x\up{i}\right)\e_c \notag\\
&=\vmat{i}^\top \e_c \kro \x\up{i}\notag\\
&= \vect_r\left(\vmat{i}^\top\e_c\x\up{i}{}^\top\right)
\end{align}
Clearly, if we are to differentiate the above gradients with respect to matrix $\vmat{j}$ or $\wmat{jj}$, for $j\neq i$, we will just get a $\bm{0}$ matrix. And, even more starkly, differentiating a second time with respect to the same matrix will also lead to that outcome. So, let's focus on the more relevant non-zero cases. A last reminder before we head off to the calculations, we will again consider this second differentiation with respect to transposed matrices, i.e., $\wmatt{ii}$ or $\vmat{i}^\top$.
\begin{align*}
    \dfrac{\,\partial \nabla_{\vmat{i}}\Fn^{c,\text{LCN}}(\x)}{\partial \wmatt{ii}} = \e_c\x\up{i}{}^\top\kro \Im_m\,.
\end{align*}
\begin{align*}
    \dfrac{\,\partial \nabla_{\wmat{ii}}\Fn^{c,\text{LCN}}(\x)}{\partial \vmat{i}^\top} = \Im_m \kro \x\up{i}\e_c^\top\,.
\end{align*}
Let's now take the sum of the above matrices over the input as well as, with appropriate scaling by the residual, over the output indices to get the blocks of the functional Hessian. This yields,
\begin{align}
    \HFsup{(\vmat{i},\wmat{ii})} = \Om\up{i} \kro \Im_m \,.
\end{align}
\begin{align}
    \HFsup{(\wmat{ii},\vmat{i})} = \Im_m \kro \Om\up{i}{}^\top\,.
\end{align}
where, $\Om\up{i} = \E[\pmb \delta_{\x, \y} \,\x\up{i}{}^\top]\in\Reals{K\times k}$ denotes the (uncentered) covariance of the residual with the $i$-th input chunk. The rank of the functional Hessian thus amounts to:
\begin{align*}
    \rank(\HF) &= t\cdot \left(\rank(\HFsup{(\vmat{i},\wmat{ii})}) +\rank(\HFsup{(\wmat{i},\vmat{ii})})\right) \\
    &= t\cdot 2\rank(\Om\up{i}\kro \Im_m) \\
    &= t\cdot 2m\rank(\Om\up{i}) = t\cdot 2m \min(k, K)\,.
\end{align*} 

\end{proof}

\subsubsection{Miscellaneous}
\paragraph{Comparison with the rank of the bigger FCN.}
Now, the bigger FCN has number of hidden neurons as $M= m.t$, while sharing the same input and output dimensions. 

\begin{itemize}
    \item $\operatorname{rank}\left(\HFsup{\text{FCN-large}}\right) = 2\,\min(d, K)\, m\,t = 2\,\min(d, K)\, M  \,=\,\dfrac{\min(d, K)}{\min(k, K)} \operatorname{rank}\left(\HFsup{\text{LCN}}\right) \,$.
    \item $\operatorname{rank}(\HOsup{\text{FCN-large}}) = q(d+K-q) \,$ where $q=\min(M, d, K)$. 
    \item $\operatorname{rank}\left(\HLsup{\text{FCN-large}}\right) = q (d+K-q) +  2\,\min\big(d, K\big)\, M\, - (2s-q)q\,,$ where $q=\min(M, d, K)$ and $s=\min(d, K)$.
\end{itemize} 

\begin{faact}
    For the LCN in Eqn.~\eqref{eq:lcn2}, we have that $\operatorname{rank}\left(\HLsup{{\text{LCN}}}\right) = \operatorname{rank}\left(\HF\right) + \operatorname{rank}(\HO) - t\cdot (2s-q)q$. 
\end{faact}

\subsection{Rank results for LCN with Weight Sharing}\label{sec:lcn+ws-proofs}
\begin{mdframed}[leftmargin=1mm,
    skipabove=2mm, 
    skipbelow=-1mm, 
    backgroundcolor=gray!10,
    linewidth=0pt,
    leftmargin=-1mm,
    rightmargin=-1mm,
    innerleftmargin=2mm,
    innerrightmargin=2mm,
    innertopmargin=2mm,
innerbottommargin=1mm]
\begin{reptheorem}{theorem:cnn1hidden-ho-hf} For the locally connected network with weight sharing defined in Eqn.~\eqref{eq:lcn+ws}, the rank of the outer-product and the functional Hessian can be bounded as:
$\rank(\HO^{\text{LCN+WS}}) \leq q(k+Kt-q)\,,$ and $\rank(\HF^{\text{LCN+WS}}) \leq  2m\min(k, Kt)$, where $q=\min(k, Kt, m)$ and $t=\dfrac{d}{k}$.
\end{reptheorem}
\end{mdframed}
\begin{proof}
The proof strategy is similar to the case of 1-hidden layer CNN as presented in the proofs of Theorems~\ref{thm:one-hid-cnn-ho},\ref{thm:one-hid-cnn-hf}. But over here since we are effectively dealing with a CNN that has stride equal to filter size, i.e., $k_1$, we will now present simplified proofs by not involving the permutation matrices $\pi_R$ or $\pi_C$.

\paragraph{Outer-product Hessian.}
Let's compute the gradient of the function with respect to the matrix $\vmat{i}$:
\begin{align}\label{eq:lcn+ws-grad-v}
\nabla_{\vmat{i}}\Fn^{\text{LCN + WS}}(\x) = \Im_K \kro \Wm \x\up{i}   
\end{align}

And that with respect to $\Wm$ is:
\begin{align}\label{eq:lcn+ws-grad-w}
\nabla_{\wmat{ii}}\Fn^{\text{LCN+WS}}(\x) = \sum\limits_{i=1}^t \vmat{i}^\top \kro \x\up{i}   
\end{align}
So the gradient of the function output with respect to all the parameters can be arranged as follows:
\begin{align*}
\nabla_\btheta \Fn^{\text{LCN+WS}}(\x) &=\begin{pmatrix}
\nabla_{\vmat{1}}\Fn^{\text{LCN+WS}}(\x) \\[2mm]
\vdots \\[2mm]
\nabla_{\vmat{t}}\Fn^{\text{LCN+WS}}(\x) \\[2mm]
\nabla_{\Wm}\Fn^{\text{LCN+WS}}(\x) \\[1mm]
\end{pmatrix} \\
&=\begin{pmatrix}
  \Im_K \kro \Wm \x\up{1} \\[2mm]
\vdots \\[2mm]
\Im_K \kro \Wm \x\up{t}  \\[2mm]
\sum\limits_{i=1}^t \vmat{i}^\top \kro \x\up{i}    \\[1mm]
\end{pmatrix}
= \begin{pmatrix}
\begin{pmatrix}
  \Im_K \kro \Wm & \cdots & \bm{0}\\[2mm]
  \vdots & \ddots & \vdots \\[2mm]
  \bm{0} & \cdots & \Im_K \kro \Wm \\[2mm]
\end{pmatrix}
\begin{pmatrix}
  \Im_K \kro \x\up{1} \\[2mm]  
\vdots \\[2mm]
\Im_K \kro \x\up{t}  \\[2mm]
\end{pmatrix} \\[5mm]
\begin{pmatrix}
  \vmat{1}^\top \kro \Im_k & \cdots & \vmat{t}^\top \kro \Im_k  \\[2mm]
\end{pmatrix}
\begin{pmatrix}
  \Im_K \kro \x\up{1} \\[2mm]  
\vdots \\[2mm]
\Im_K \kro \x\up{t}  \\[2mm]
\end{pmatrix}    \\[1mm]
\end{pmatrix}\\
&=  \begin{pmatrix}
  \Im_K \kro \Wm & \cdots & \bm{0}\\[2mm]
  \vdots & \ddots & \vdots \\[2mm]
  \bm{0} & \cdots & \Im_K \kro \Wm \\[2mm]
  \vmat{1}^\top \kro \Im_k & \cdots & \vmat{t}^\top \kro \Im_k  \\[2mm]
\end{pmatrix}\begin{pmatrix}
  \Im_K \kro \x\up{1} \\[2mm]  
\vdots \\[2mm]
\Im_K \kro \x\up{t}  \\[2mm]
\end{pmatrix}    \\[1mm]
&=\underbrace{\begin{pmatrix}
    \Im_{Kt} \kro \Wm \\[2mm]
    \Vm^\top \kro \Im_k \\[2mm]
\end{pmatrix}}_{\Am_o}\underbrace{\begin{pmatrix}
  \Im_K \kro \x\up{1} \\[1mm]  
\vdots \\[1mm]
\Im_K \kro \x\up{t}  \\[1mm]
\end{pmatrix}}_{\Bm}    
\end{align*}
In the last line, we used the associativity of Kronecker product and the fact that $\Cm\kro\Dm = \begin{pmatrix}\matcol{\Cm}{1} \kro \Dm \cdots \matcol{\Cm}{n} \kro \Dm\end{pmatrix}$. 

The outer-product Hessian can the be calculated by taking the outer-product of the above gradient and averaging over the samples in the dataset. This leads to $$\HO=\Am_o\left(\Im_K\kro\Bm_o\right)\Am_o^\top\,,$$
where, $\Bm_o$ comes out to be the same matrix as in Eqn.~\eqref{eq:lcn-ho-decomp} before. Now, let's analyze the rank of the resulting matrices:

\begin{align*}
\rank(\HO)&=\min(\rank(\Am_o), K\rank(\Bm_o))\leq \min(\rank(\Am_o), Kd)\\
&= \min(q\cdot (k+ Kt - q), Kd) = q\cdot (k+ Kt - q)\,,
\end{align*}
with $q=\min(k, Kt, m)$.

\paragraph{Functional Hessian.}
We need to differentiate again the network output gradients in Eqns.~\eqref{eq:lcn+ws-grad-v},~\eqref{eq:lcn+ws-grad-w}. First, let's just obtain the gradient for the output at the $c$-th index:
\begin{align}
\nabla_{\vmat{i}}\Fn^{c,\text{LCN+WS}}(\x) &= \left(\Im_K \kro \Wm\x\up{i})\right)\e_c \notag\\ 
&=  \e_c \kro \Wm\x\up{i} \notag\\
&= \vect_r\left(\e_c\x\up{i}{}^\top\Wm^\top\right)
\end{align}

And that with respect to $\Wm$ is:
\begin{align}
\nabla_{\Wm}\Fn^{c, \text{LCN+WS}}(\x) &= \sum\limits_{i=1}^t\left(\vmat{i}^\top \kro \x\up{i}\right)\e_c \notag\\
&=\sum\limits_{i=1}^t\vmat{i}^\top \e_c \kro \x\up{i}\notag\\
&= \sum\limits_{i=1}^t\vect_r\left(\vmat{i}^\top\e_c\x\up{i}{}^\top\right)
\end{align}

\begin{align*}
    \dfrac{\,\partial \nabla_{\vmat{i}}\Fn^{c,\text{LCN+WS}}(\x)}{\partial \Wm^\top} = \e_c\x\up{i}{}^\top\kro \Im_m\,.
\end{align*}
\begin{align*}
    \dfrac{\,\partial \nabla_{\Wm}\Fn^{c,\text{LCN+FS}}(\x)}{\partial \vmat{i}^\top} = \Im_m \kro \x\up{i}\e_c^\top\,.
\end{align*}
Let's now take the sum of the above matrices over the input as well as, with appropriate scaling by the residual, over the output indices to get the blocks of the functional Hessian. This yields,
\begin{align}
    \HFsup{(\vmat{i},\Wm)} = \Om\up{i} \kro \Im_m \,.
\end{align}
\begin{align}
    \HFsup{(\Wm,\vmat{i})} = \Im_m \kro \Om\up{i}{}^\top\,.
\end{align}
where, $\Om\up{i} = \E[\pmb \delta_{\x, \y} \,\x\up{i}{}^\top]$ denotes the (uncentered) covariance of the residual with the $i$-th input chunk.

Hence the rank of the functional Hessian, since it is block hollow, comes out to be:
\begin{align*}
    \rank(\HF)&= 2\rank \begin{pmatrix}
        \Om\up{1}\kro\Im_m \\[1mm]
        \vdots \\[1mm]
        \Om\up{t}\kro\Im_m\\[1mm]
    \end{pmatrix} = 2\rank(\widetilde{\Om}\kro\Im_m) = 2m \rank(\widetilde{\Om}) \leq 2m\min(k, Kt)\,.
\end{align*} 
Here in the above line, we can instead consider the rank of $\widetilde{\Om}=\begin{pmatrix}
    \Om\up{1}\\
    \vdots\\
    \Om\up{t}
\end{pmatrix}\in\Reals{Kt\times k}$ kroneckered with the identity matrix ($\Im_{m}$) as rank is unaffected by row or column permutations.
\end{proof}
\subsection{Miscellaneous}
\begin{faact}
         The rank of the overall loss Hessian has the following formulae \begin{align*} \operatorname{rank}\left(\HLsup{{\text{LCN+WS}}}\right)& = \operatorname{rank}\left(\HOsup{\text{LCN+WS}}\right) + 
 \\
 & \operatorname{rank}\left(\HFsup{\text{LCN+WS}}\right) - (2s-q)q\,,\end{align*} where $q=\min(k, K, Kt)$ and $s=\min(k, Kt)$.
    \end{faact} 
    
\section{Additional Results}\label{sec:figures}
\subsection{Effect of number of samples $n$}
\begin{figure*}[!t]
    \centering
    
    \begin{subfigure}[b]{0.45\textwidth}
    \includegraphics[width=\textwidth]{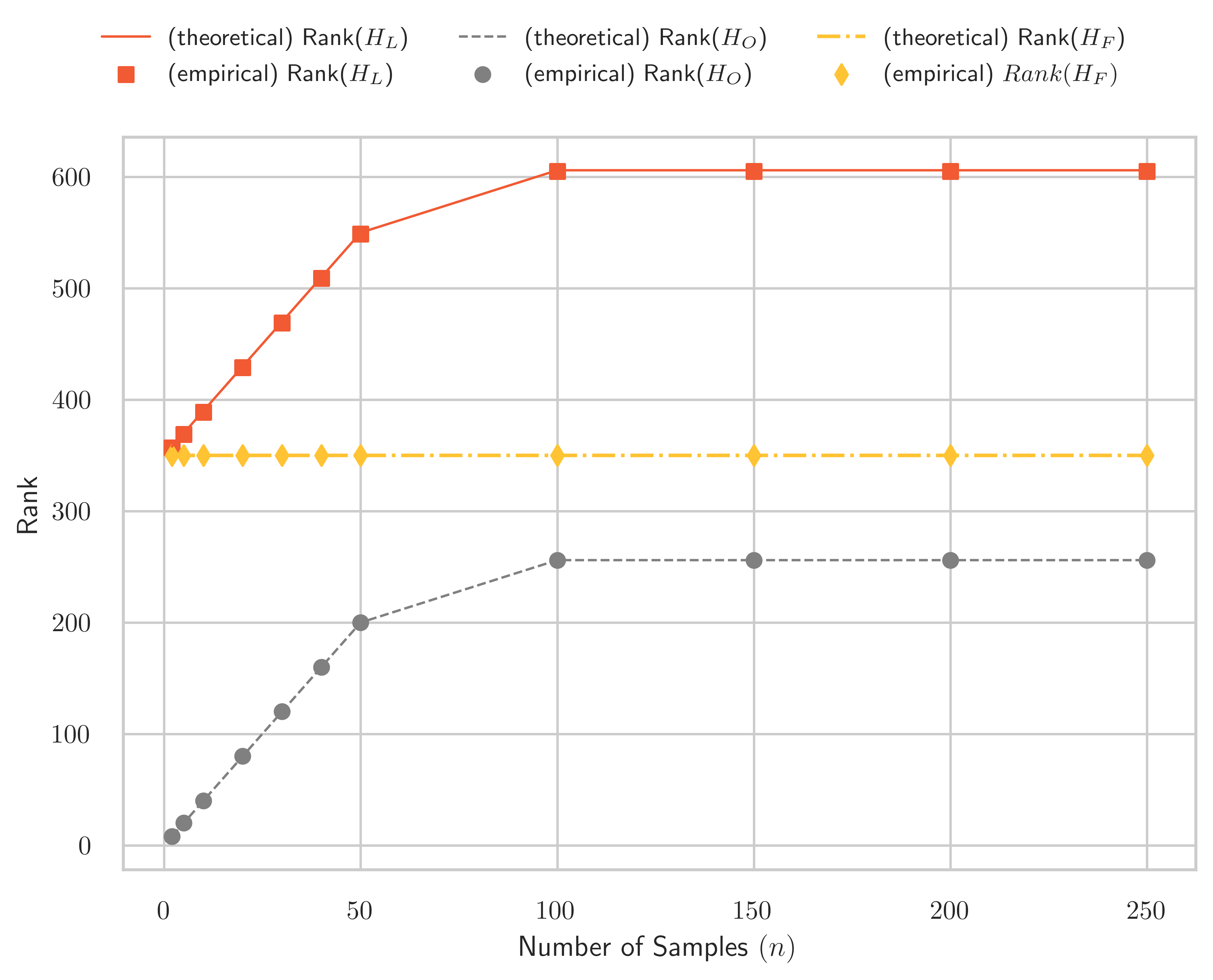}
    \caption{Effect of number of samples}
    \end{subfigure}
    \caption{The rank of outer product and loss Hessian increase for a while with increasing number of samples, as until then the covariance matrix has rank $n$ instead of $d$. Once $n=d$, the ranks remain constant. The rank of the functional Hessian is unaffected throughout.}
\end{figure*}
\subsection{Rank vs Number of Channels}


\subsubsection{2-hidden layer CNNs}

\begin{figure*}[!h]
	\centering
	\begin{subfigure}[b]{0.3\textwidth}
		
		\includegraphics[width=\textwidth]{figures/two_hidden/cifar10_linear_mse_two_hidden_layer/hess-L.png}
		\caption{$\HL$, Linear, CIFAR10}
	\end{subfigure}
	\begin{subfigure}[b]{0.3\textwidth}
		
		\includegraphics[width=\textwidth]{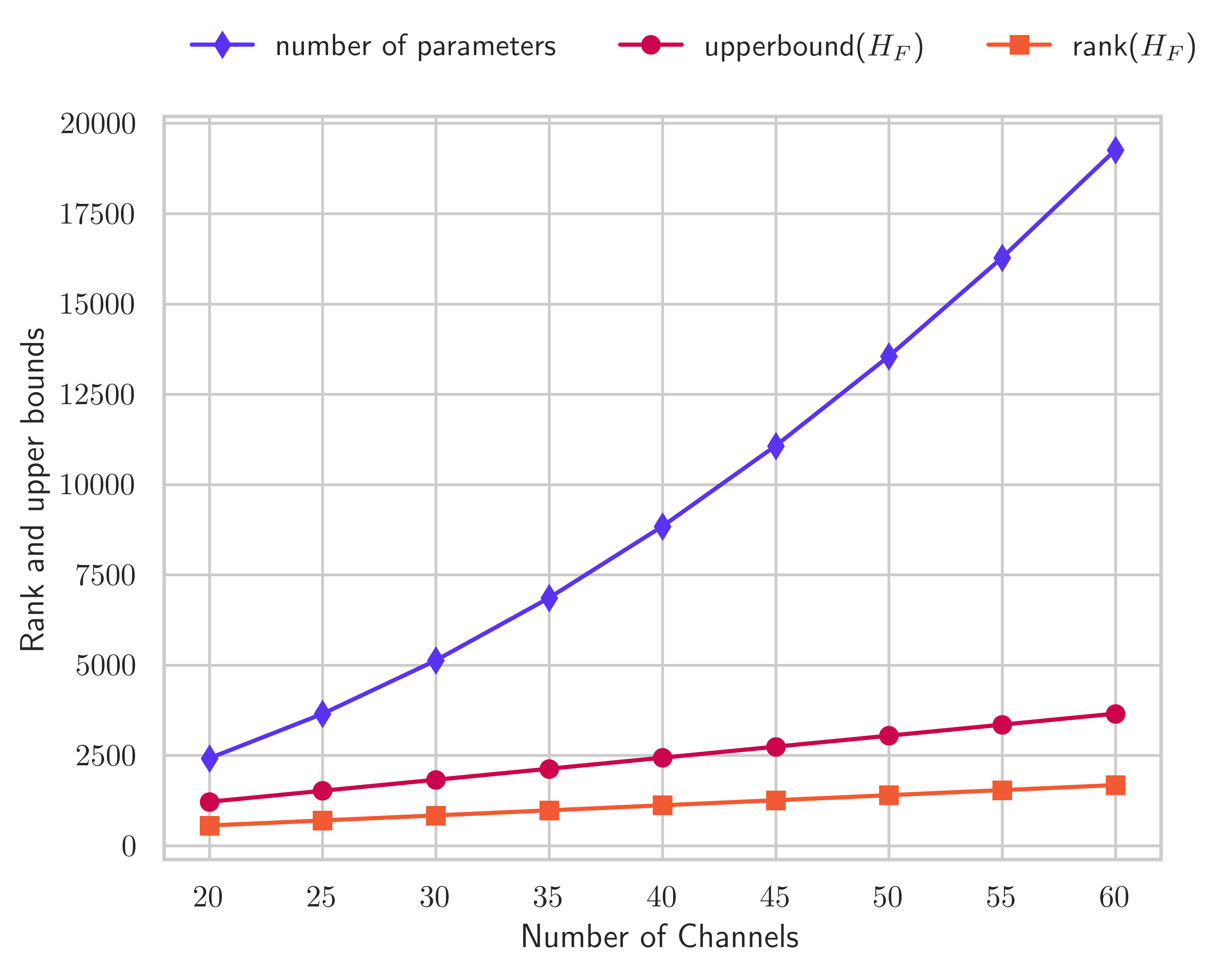}
		\caption{$\HF$, Linear, CIFAR10}
	\end{subfigure}
	\begin{subfigure}[b]{0.3\textwidth}
		
		\includegraphics[width=\textwidth]{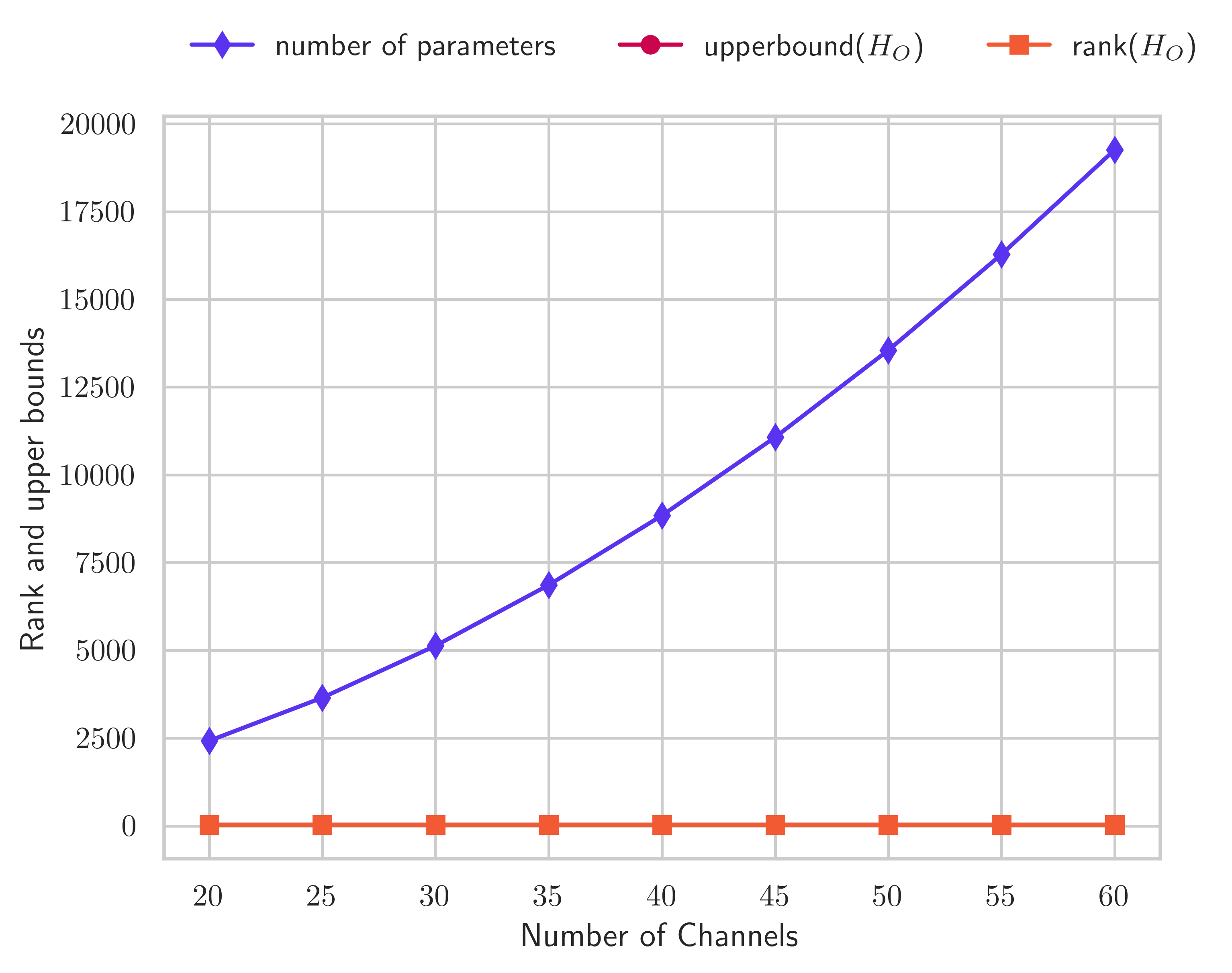}
		\caption{$\HO$, Linear, CIFAR10}
	\end{subfigure}

	\begin{subfigure}[b]{0.3\textwidth}
		
		\includegraphics[width=\textwidth]{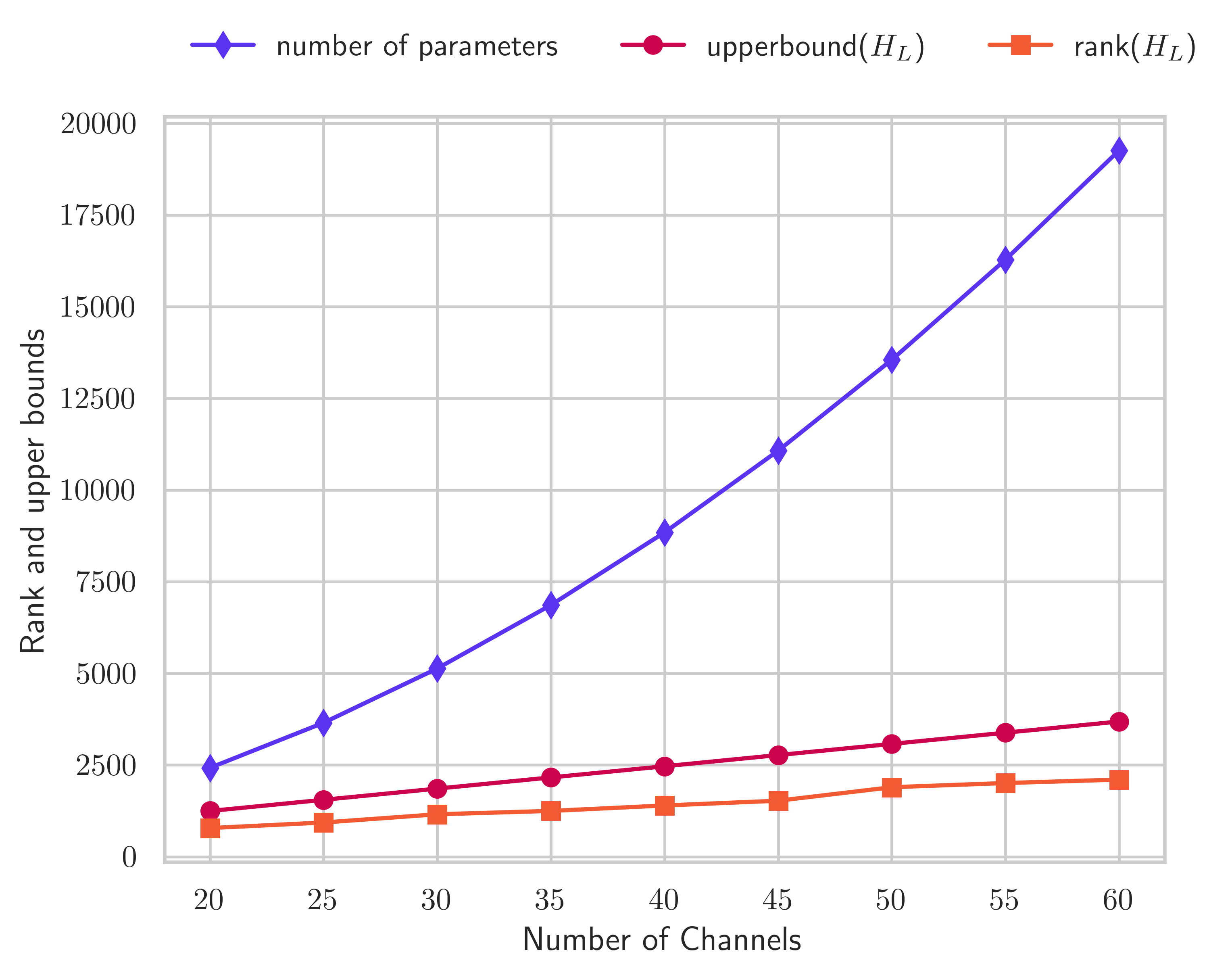}
		\caption{$\HL$, ReLU, CIFAR10}
	\end{subfigure}
	\begin{subfigure}[b]{0.3\textwidth}
		
		\includegraphics[width=\textwidth]{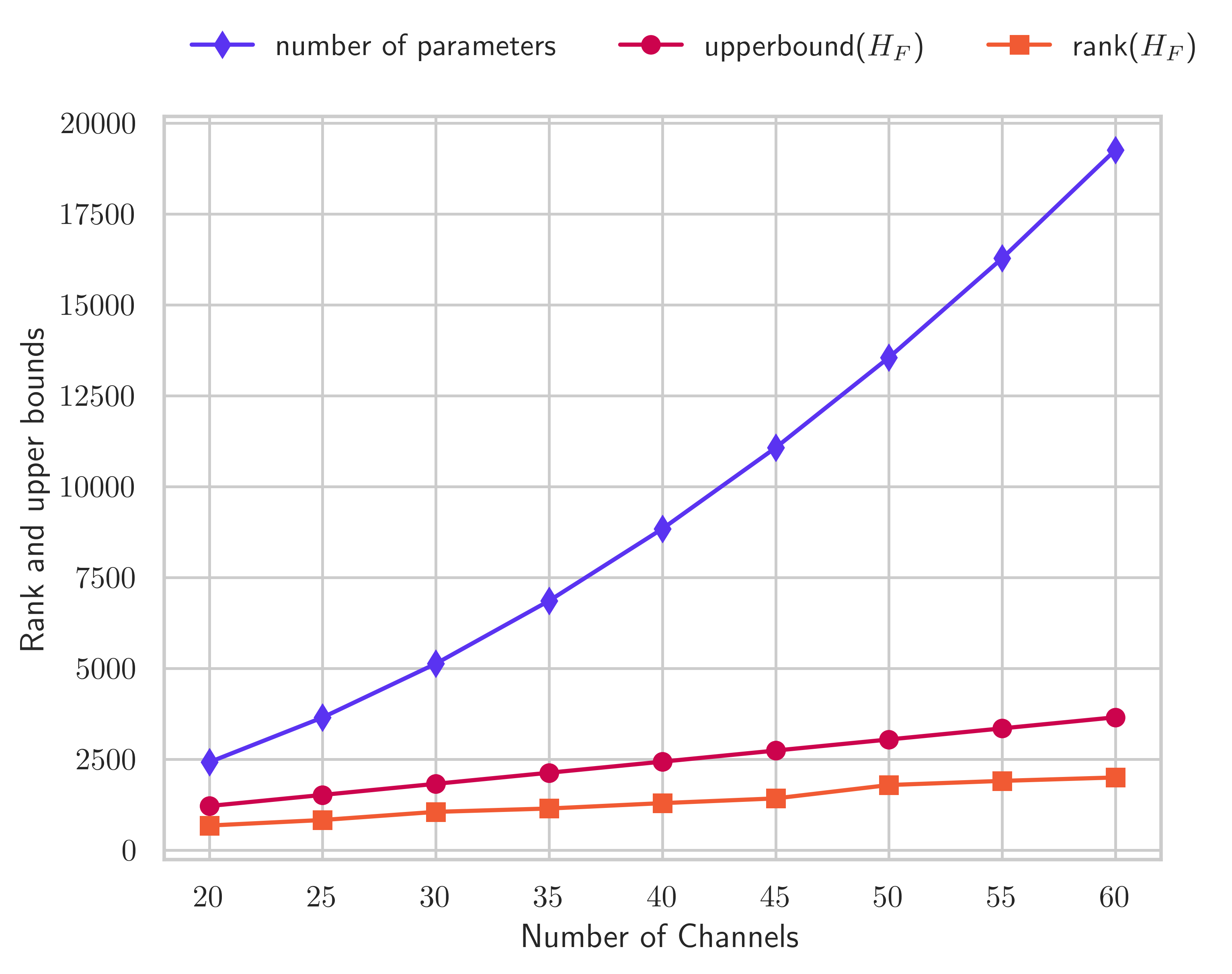}
		\caption{$\HF$, ReLU, CIFAR10}
	\end{subfigure}
	\begin{subfigure}[b]{0.3\textwidth}
		
		\includegraphics[width=\textwidth]{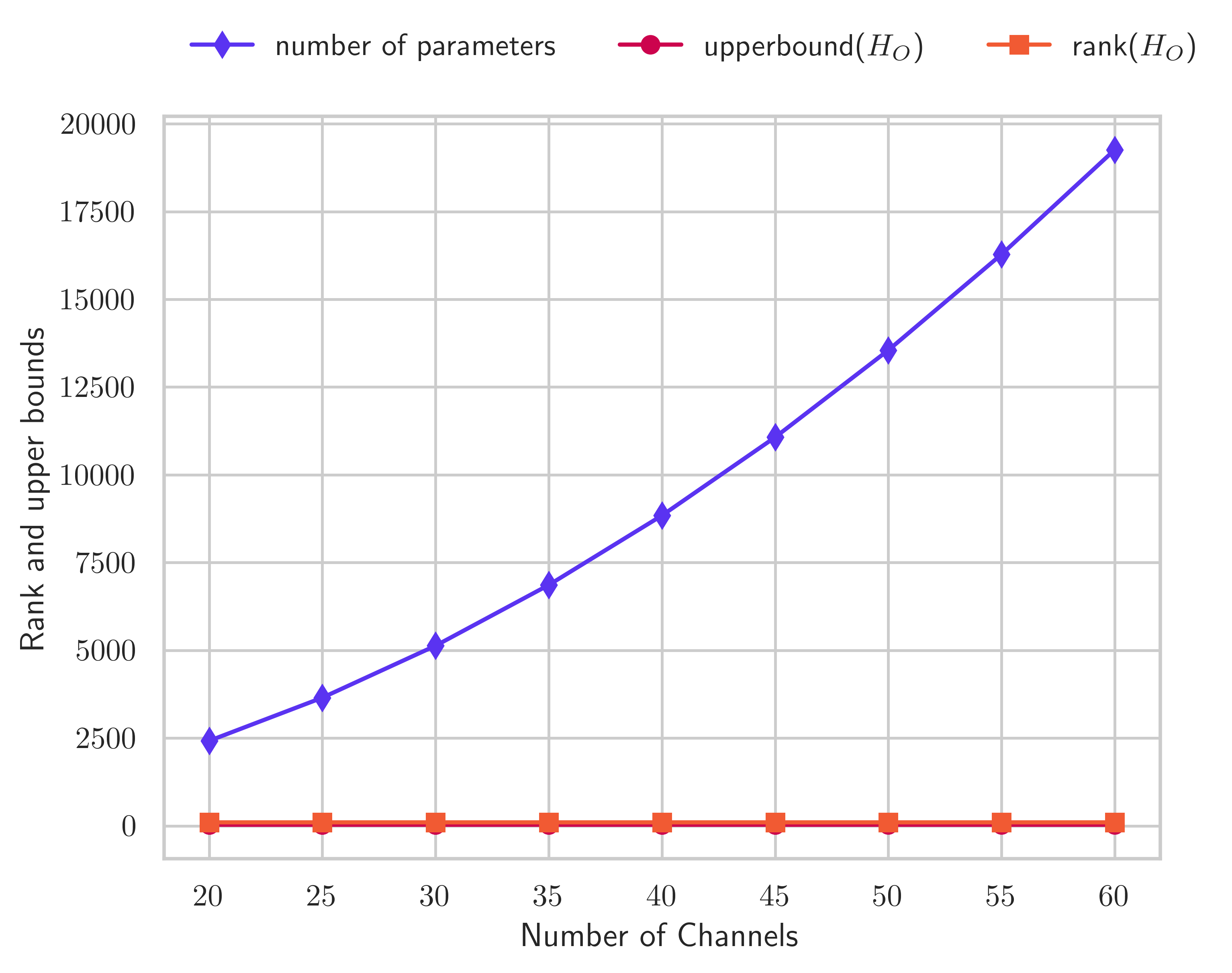}
		\caption{$\HO$, ReLU, CIFAR10}
	\end{subfigure}

	\begin{subfigure}[b]{0.3\textwidth}
		
		\includegraphics[width=\textwidth]{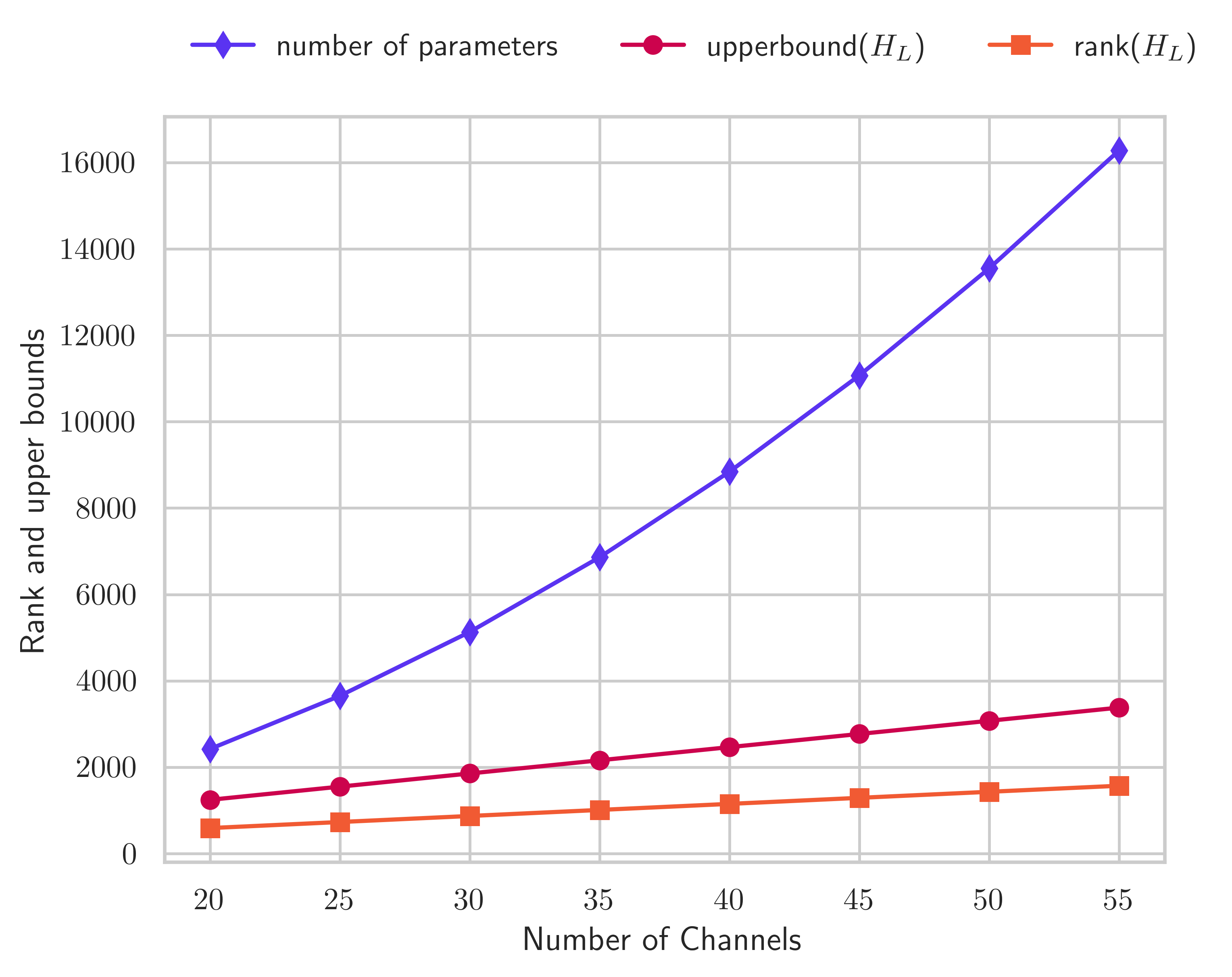}
		\caption{$\HL$, Linear, Gaussian data}
	\end{subfigure}
	\begin{subfigure}[b]{0.3\textwidth}
		
		\includegraphics[width=\textwidth]{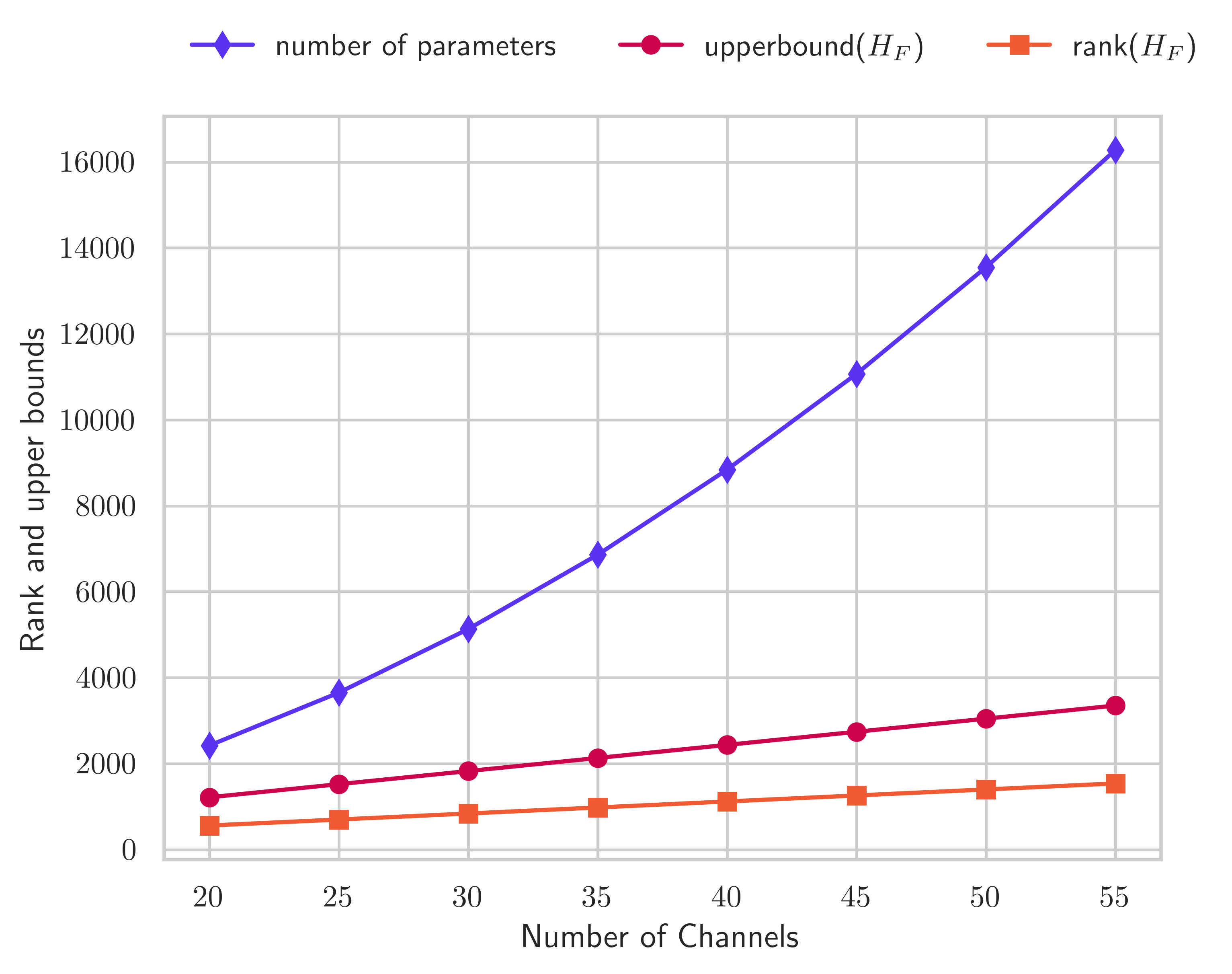}
		\caption{$\HF$, Linear, Gaussian data}
	\end{subfigure}
	\begin{subfigure}[b]{0.3\textwidth}
		
		\includegraphics[width=\textwidth]{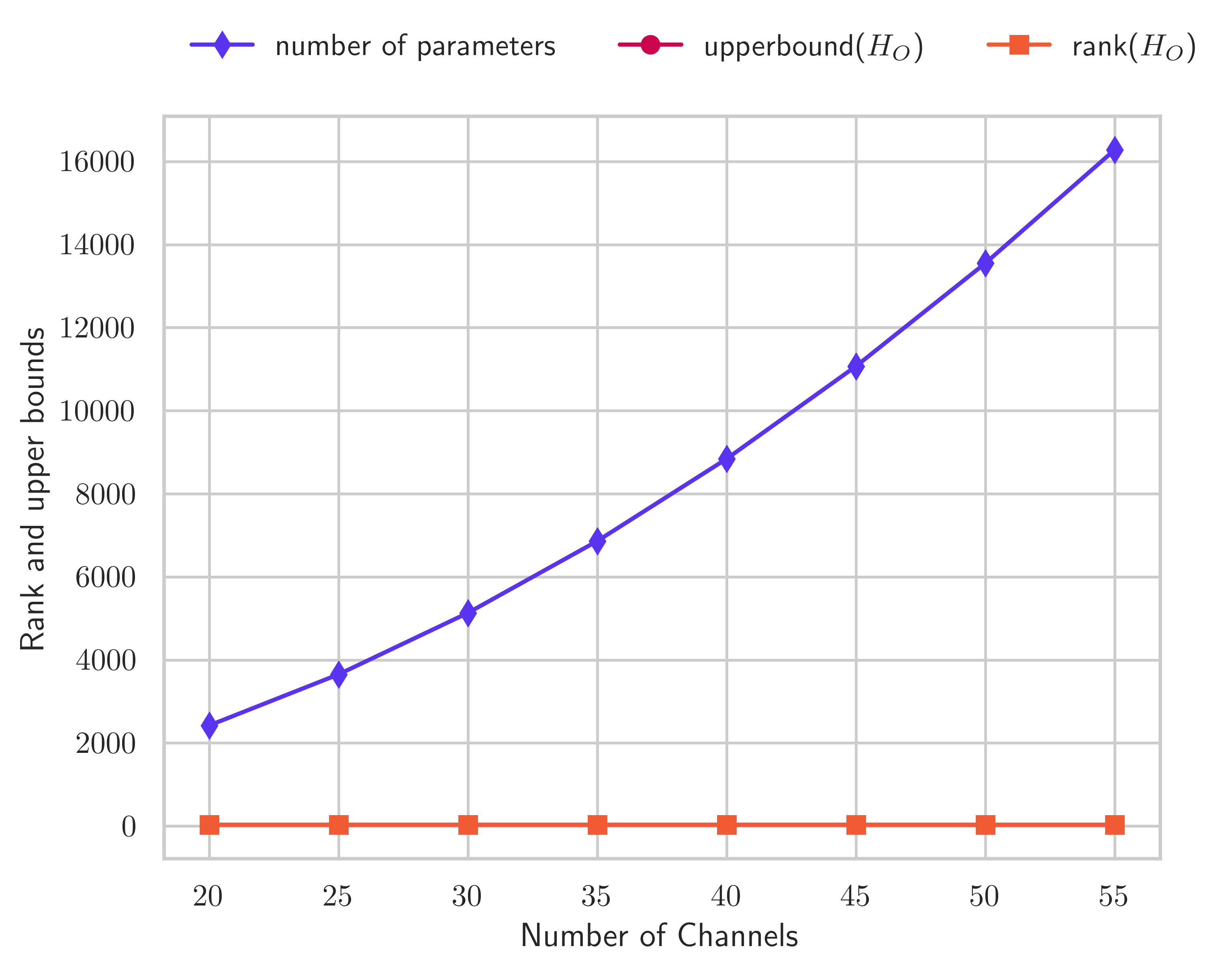}
		\caption{$\HO$, Linear, Gaussian data}
	\end{subfigure}

	\begin{subfigure}[b]{0.3\textwidth}
		
		\includegraphics[width=\textwidth]{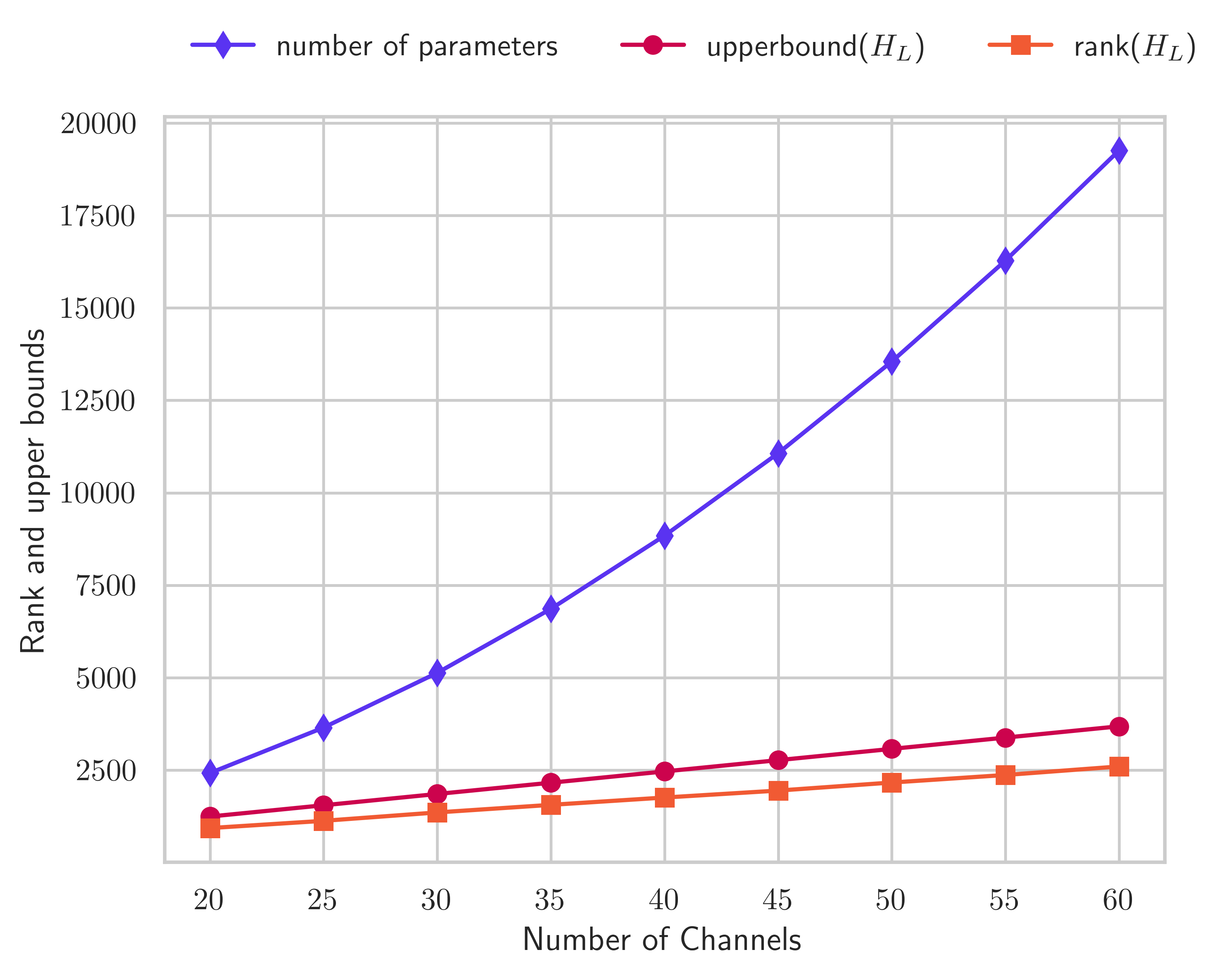}
		\caption{$\HL$, ReLU, Gaussian data}
	\end{subfigure}
	\begin{subfigure}[b]{0.3\textwidth}
		\includegraphics[width=\textwidth]{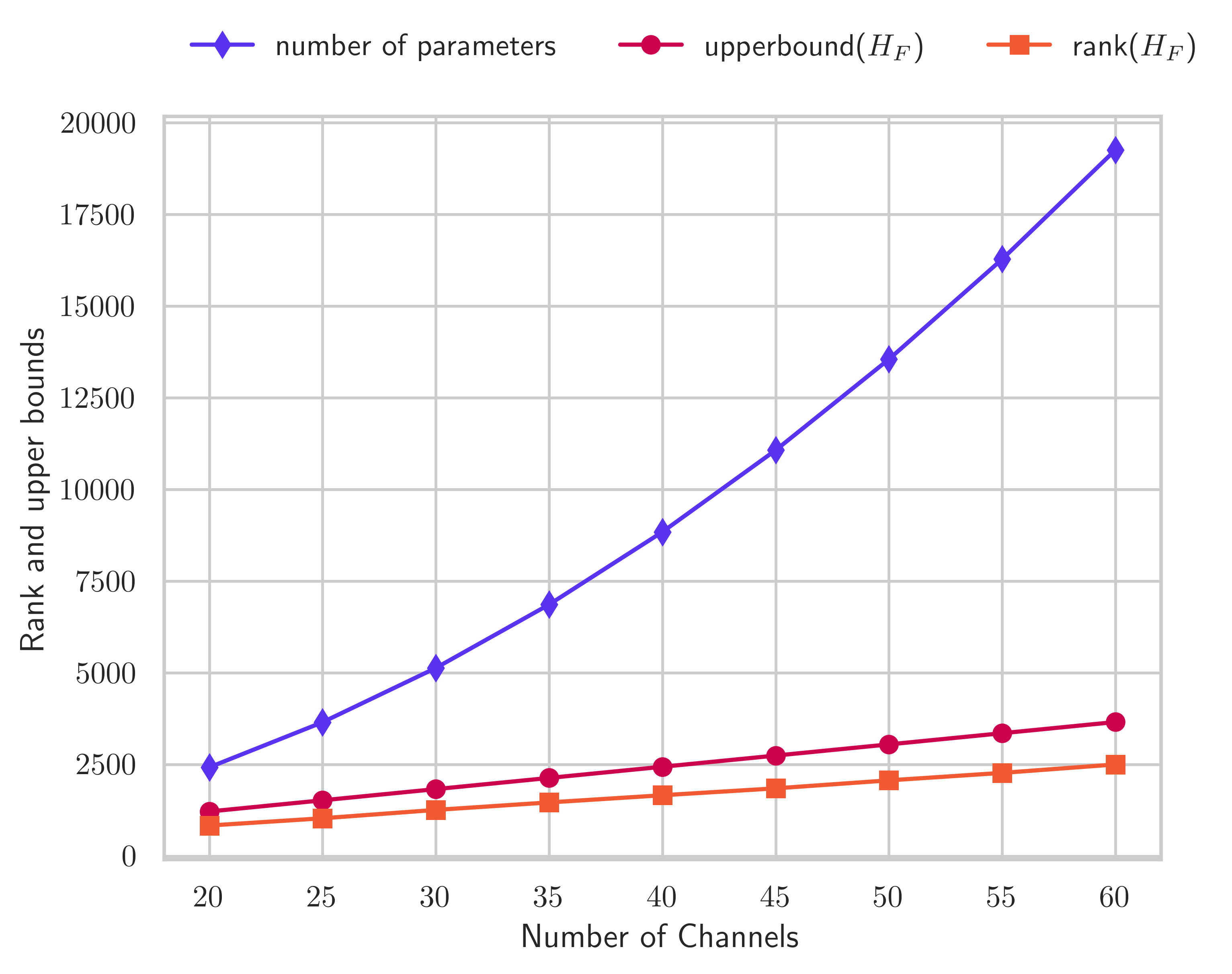}
		\caption{$\HF$, ReLU, Gaussian data}
	\end{subfigure}
     \begin{subfigure}[b]{0.3\textwidth}
		\includegraphics[width=\textwidth]{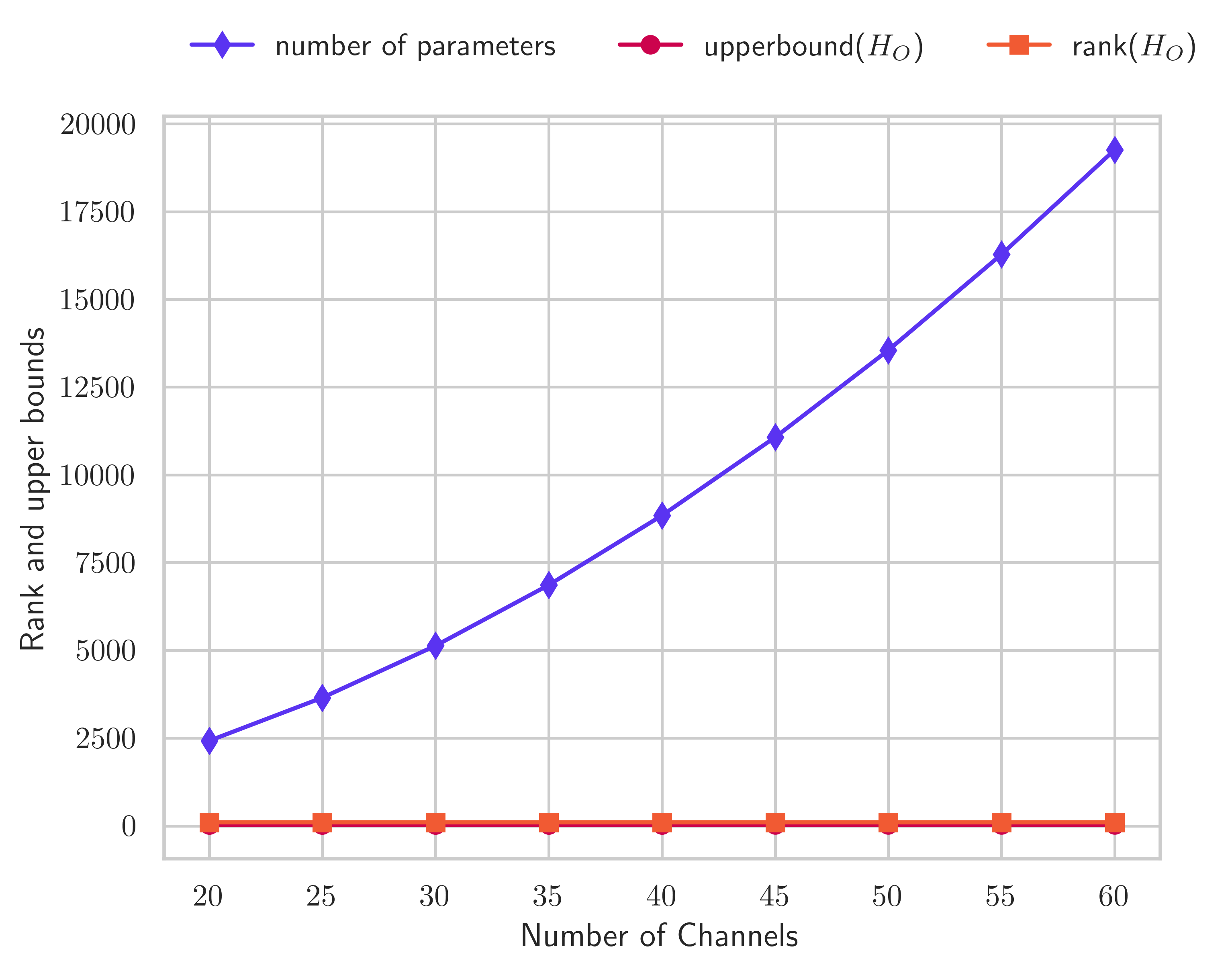}
		\caption{$\HO$, ReLU, Gaussian data}
	\end{subfigure}
\caption{Rank vs Number of Channels for 2-hidden layer $\lbrace$linear, ReLU$\rbrace$- CNNs on $\lbrace$CIFAR10, random Gaussians$\rbrace$ with Mean Squared Error loss. The loss, functional, and outer-product Hessian are each shown as separate figures. The bounds for outer-product Hessian exactly coincide with the true values.}
\end{figure*}

\clearpage
\subsection{Rank vs Filter Size}
\subsubsection{Linear activations, MSE loss }\label{supp:fil-lin-mse}
\begin{figure*}[!h]
	\centering
	\begin{subfigure}[b]{0.3\textwidth}
		
		\includegraphics[width=\textwidth]{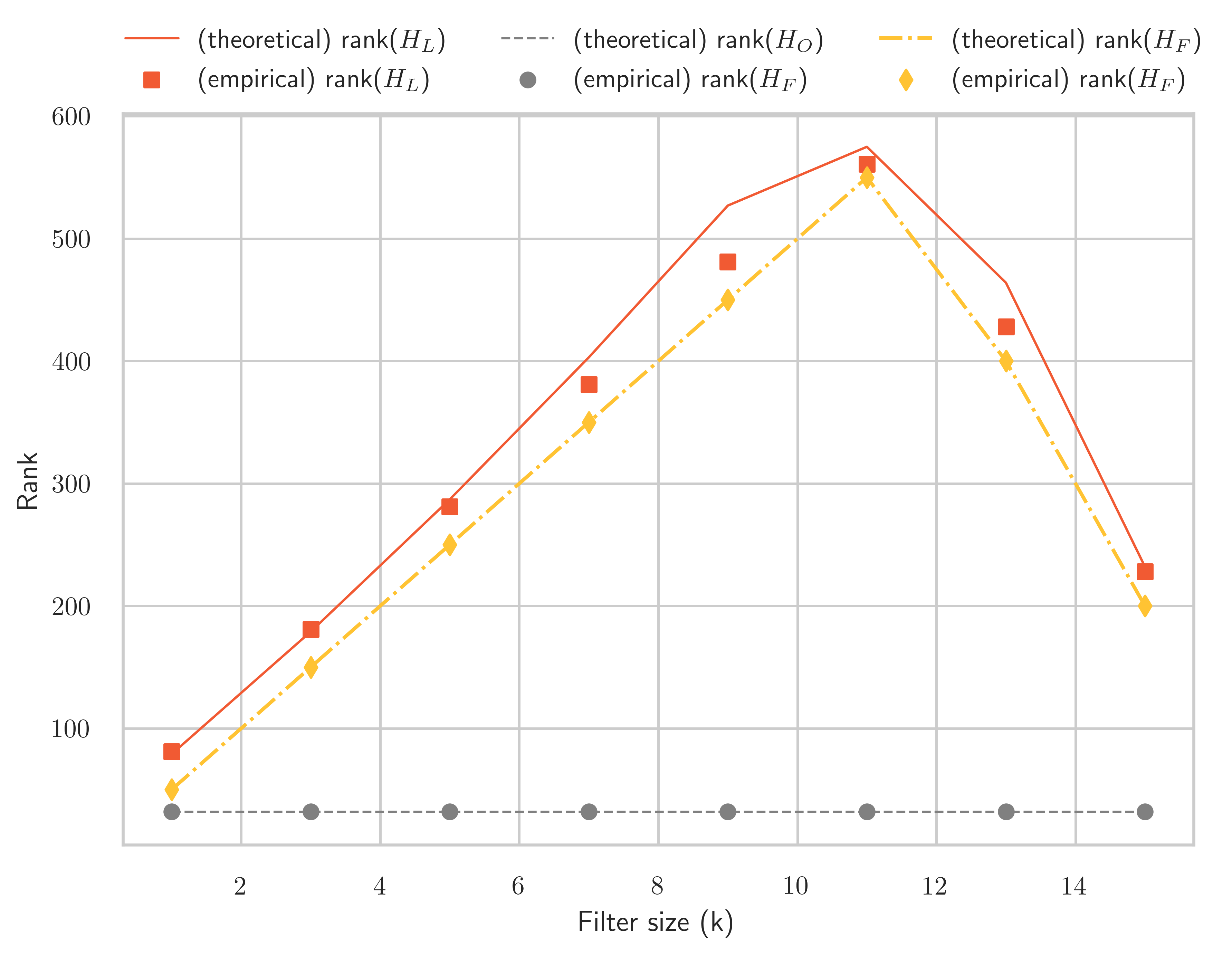}
		\caption{$m=25$}
	\end{subfigure}
	\begin{subfigure}[b]{0.3\textwidth}
		
		\includegraphics[width=\textwidth]{figures/filter_size/jaxcifar10_mse_linear_m-35_d-16/rank_vs_filtersize.png}
		\caption{$m=35$}
	\end{subfigure}
\begin{subfigure}[b]{0.3\textwidth}
	
	\includegraphics[width=\textwidth]{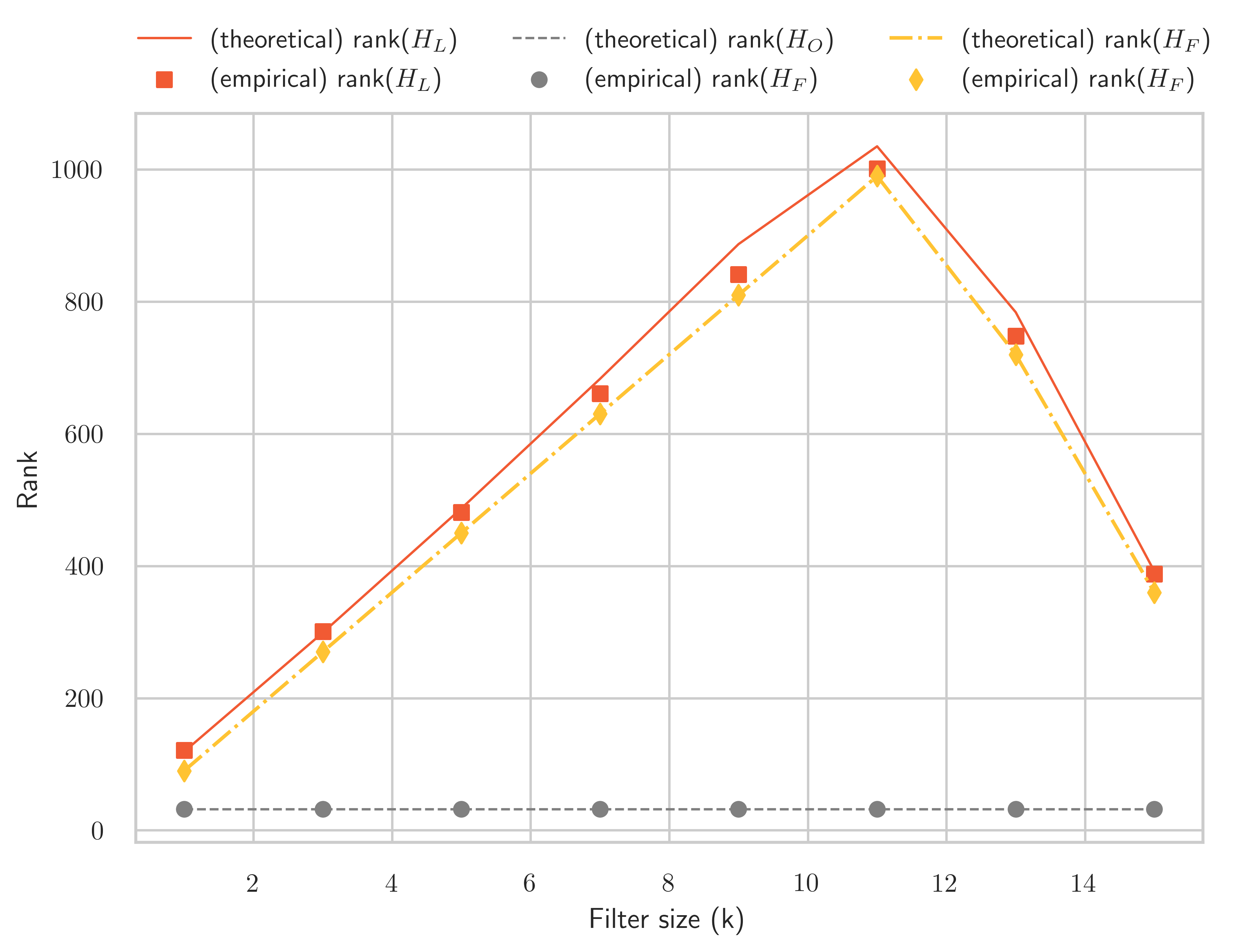}
	\caption{$m=45$}
\end{subfigure}
	\begin{subfigure}[b]{0.3\textwidth}
		
		\includegraphics[width=\textwidth]{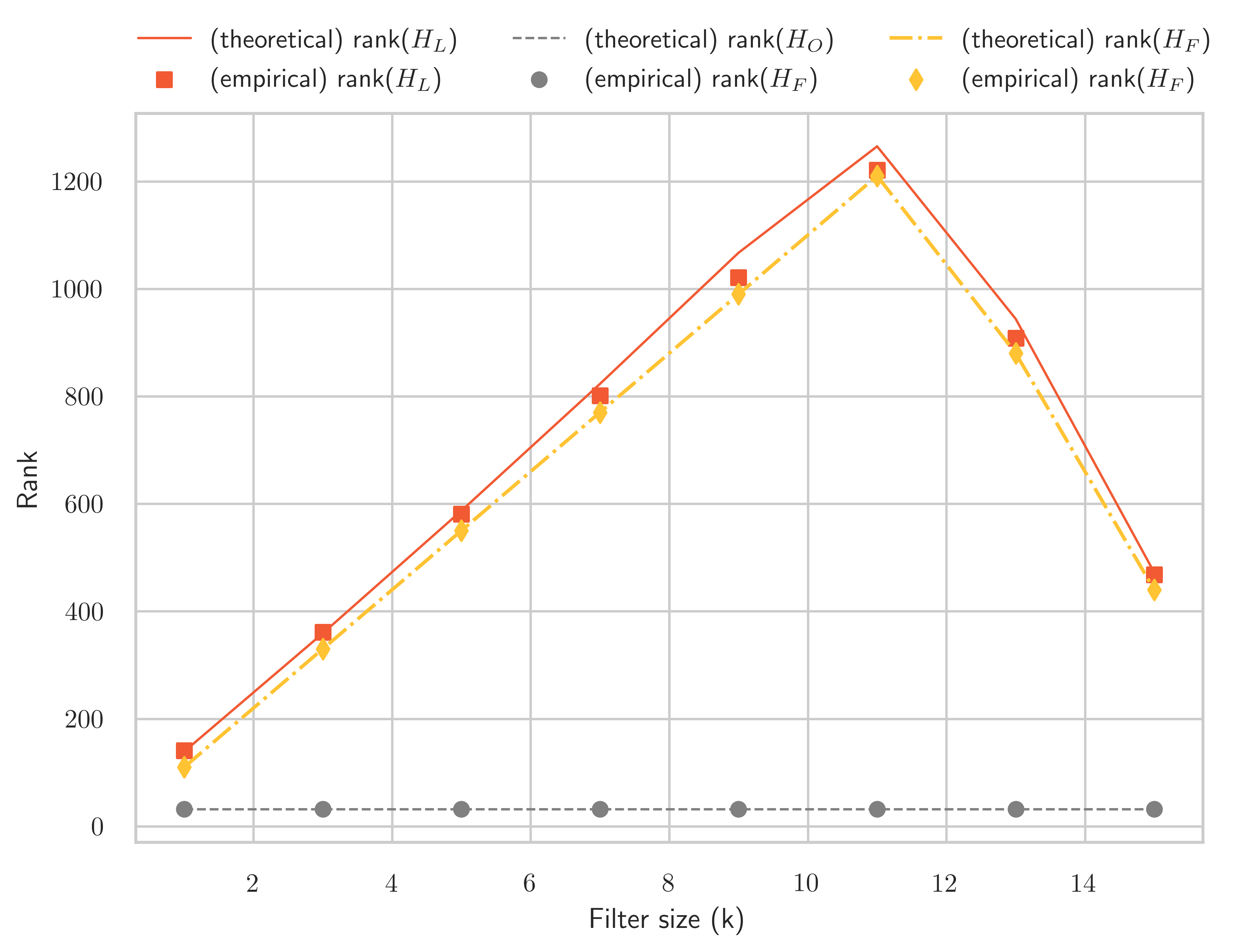}
		\caption{$m=55$}
	\end{subfigure}
\begin{subfigure}[b]{0.3\textwidth}
	
	\includegraphics[width=\textwidth]{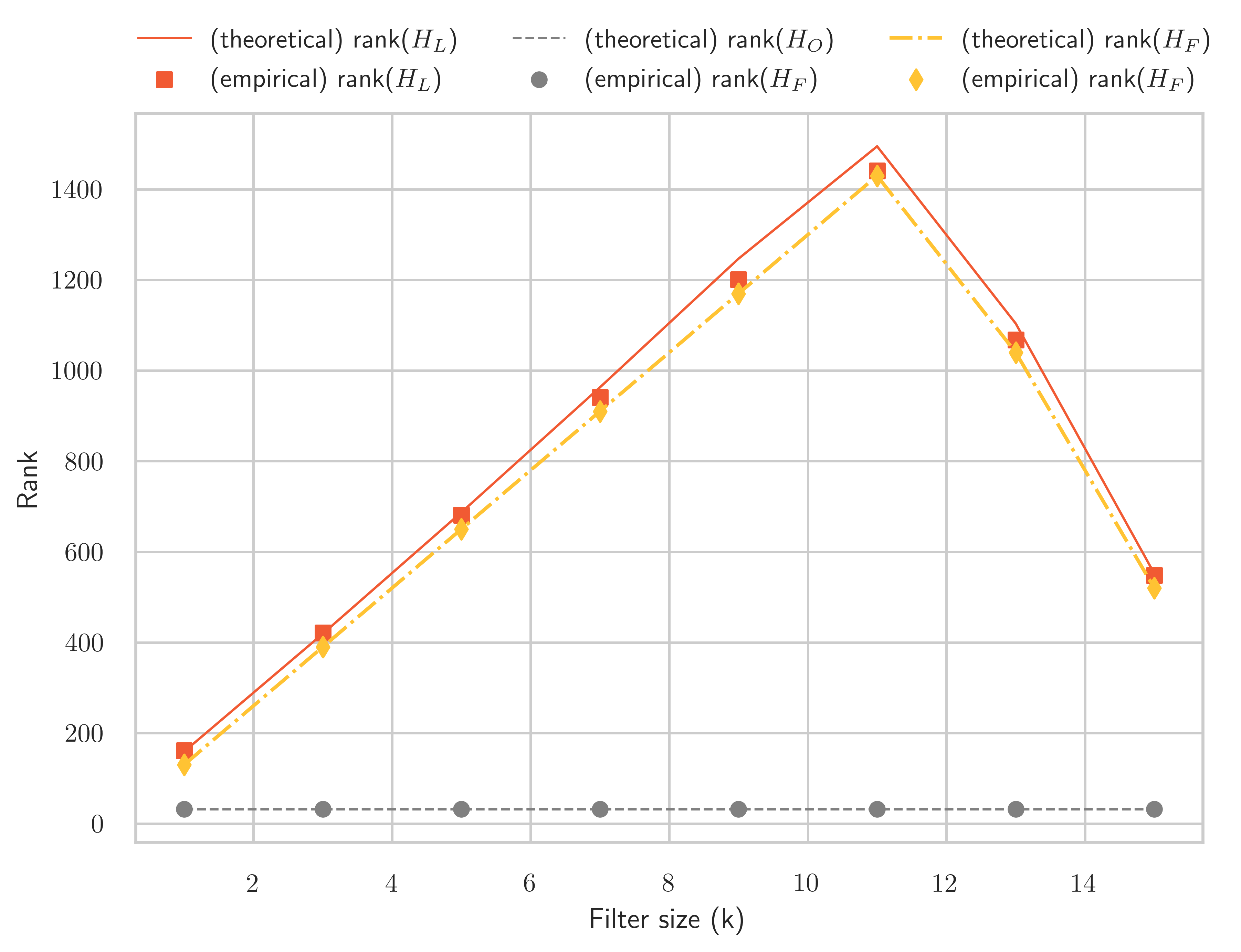}
	\caption{$m=65$}
\end{subfigure}
	\begin{subfigure}[b]{0.3\textwidth}
	
	\includegraphics[width=\textwidth]{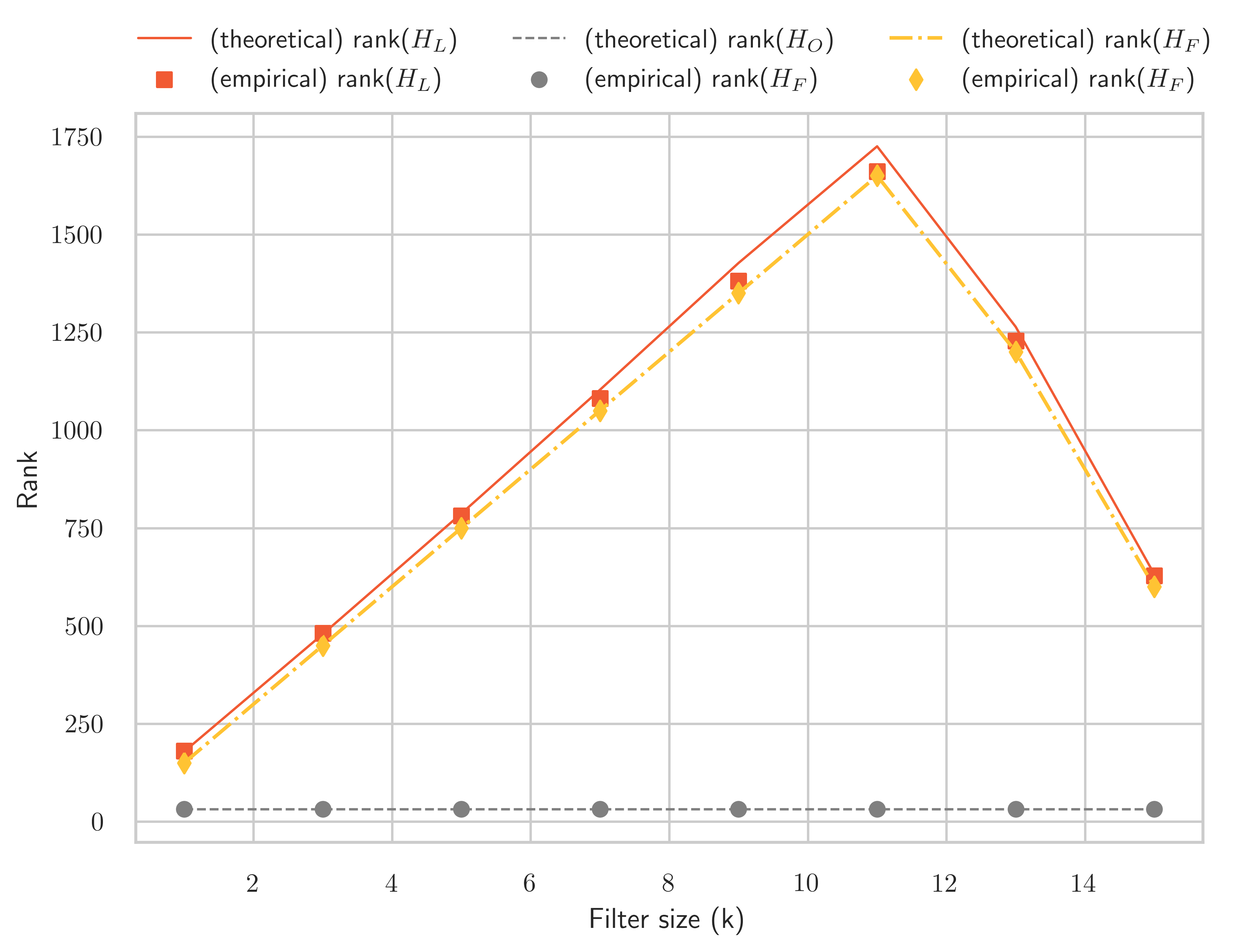}
	\caption{$m=75$}
\end{subfigure}
\begin{subfigure}[b]{0.3\textwidth}
	
	\includegraphics[width=\textwidth]{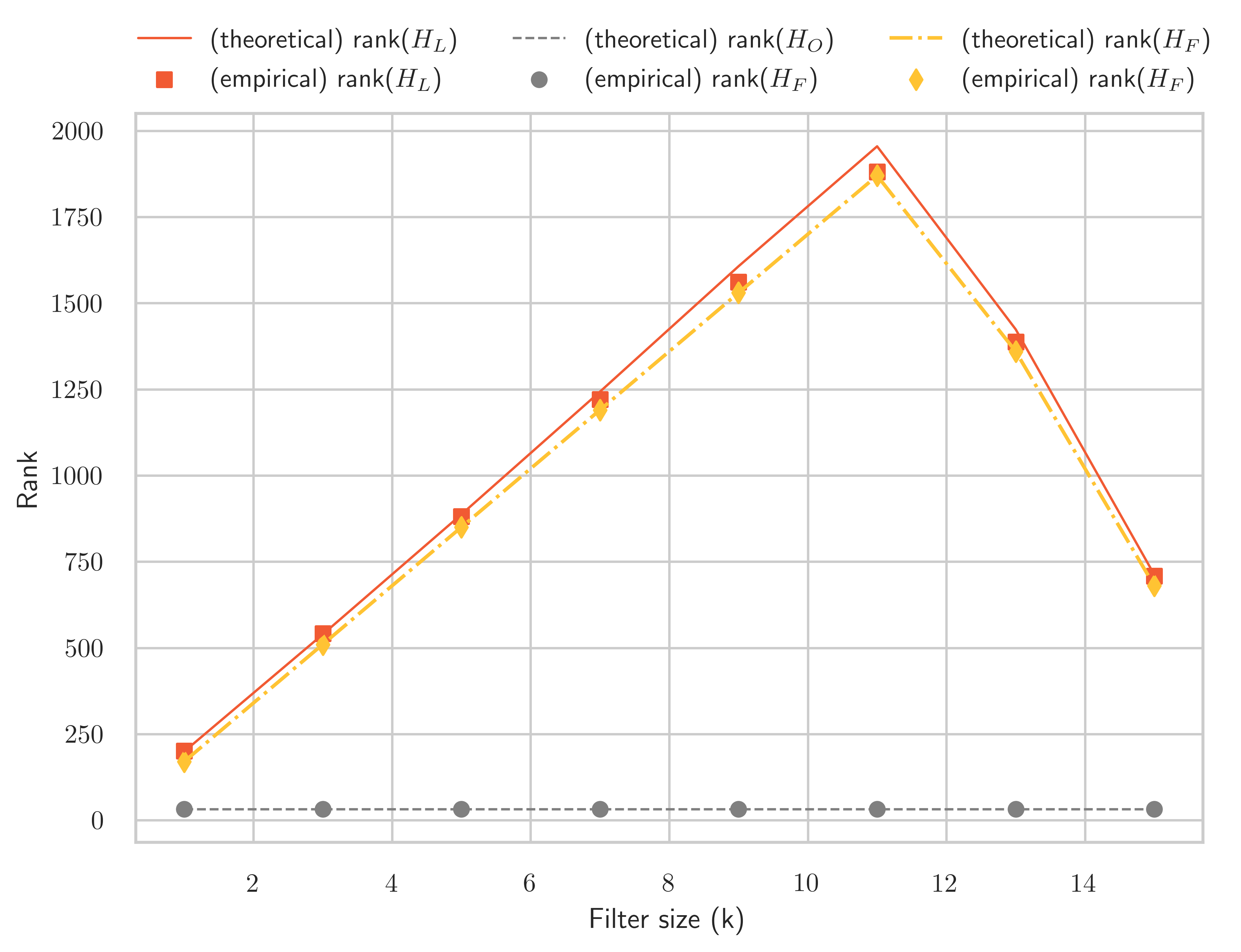}
	\caption{$m=85$}
\end{subfigure}
\begin{subfigure}[b]{0.3\textwidth}
	
	\includegraphics[width=\textwidth]{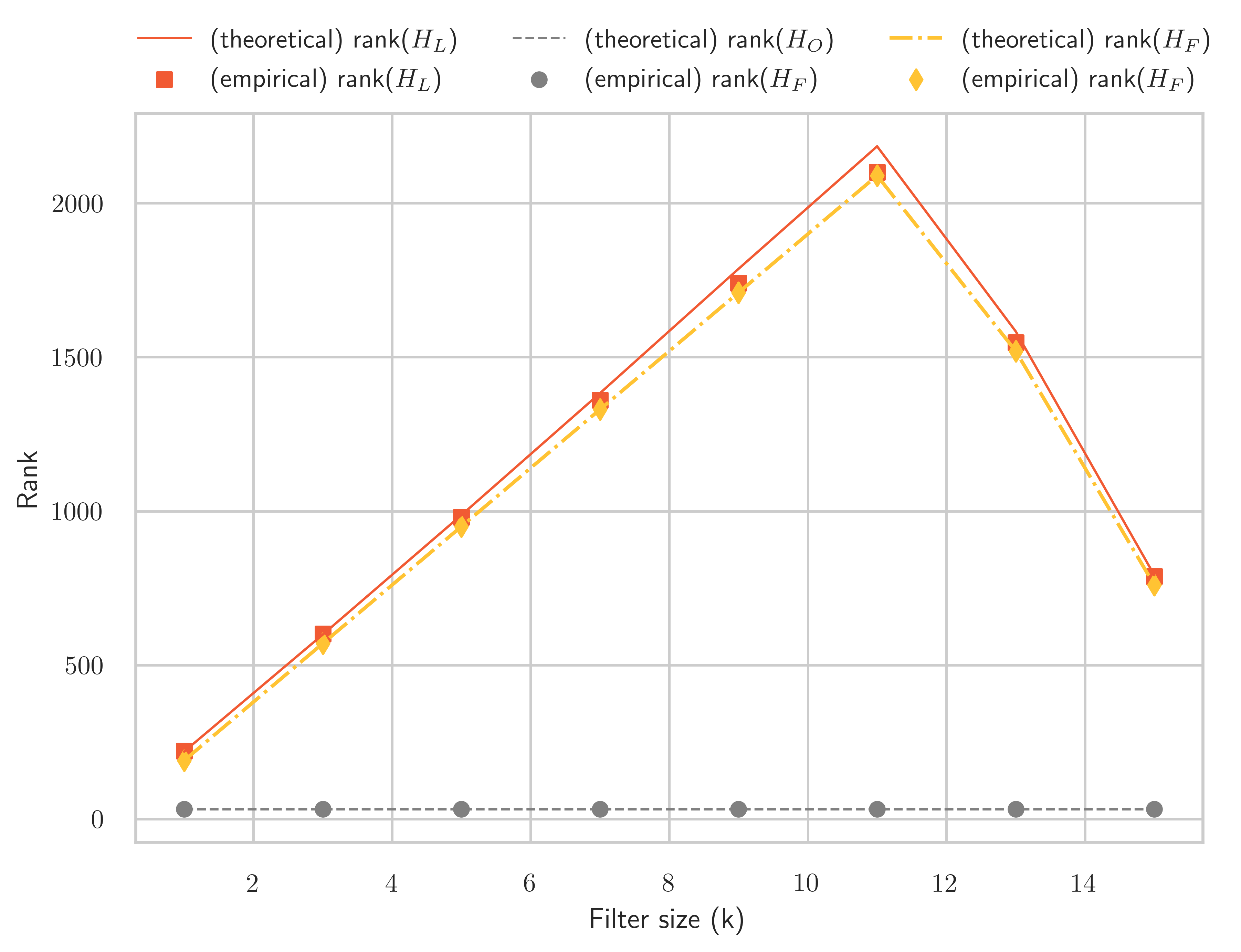}
	\caption{$m=95$}
\end{subfigure}
\begin{subfigure}[b]{0.3\textwidth}
	
	\includegraphics[width=\textwidth]{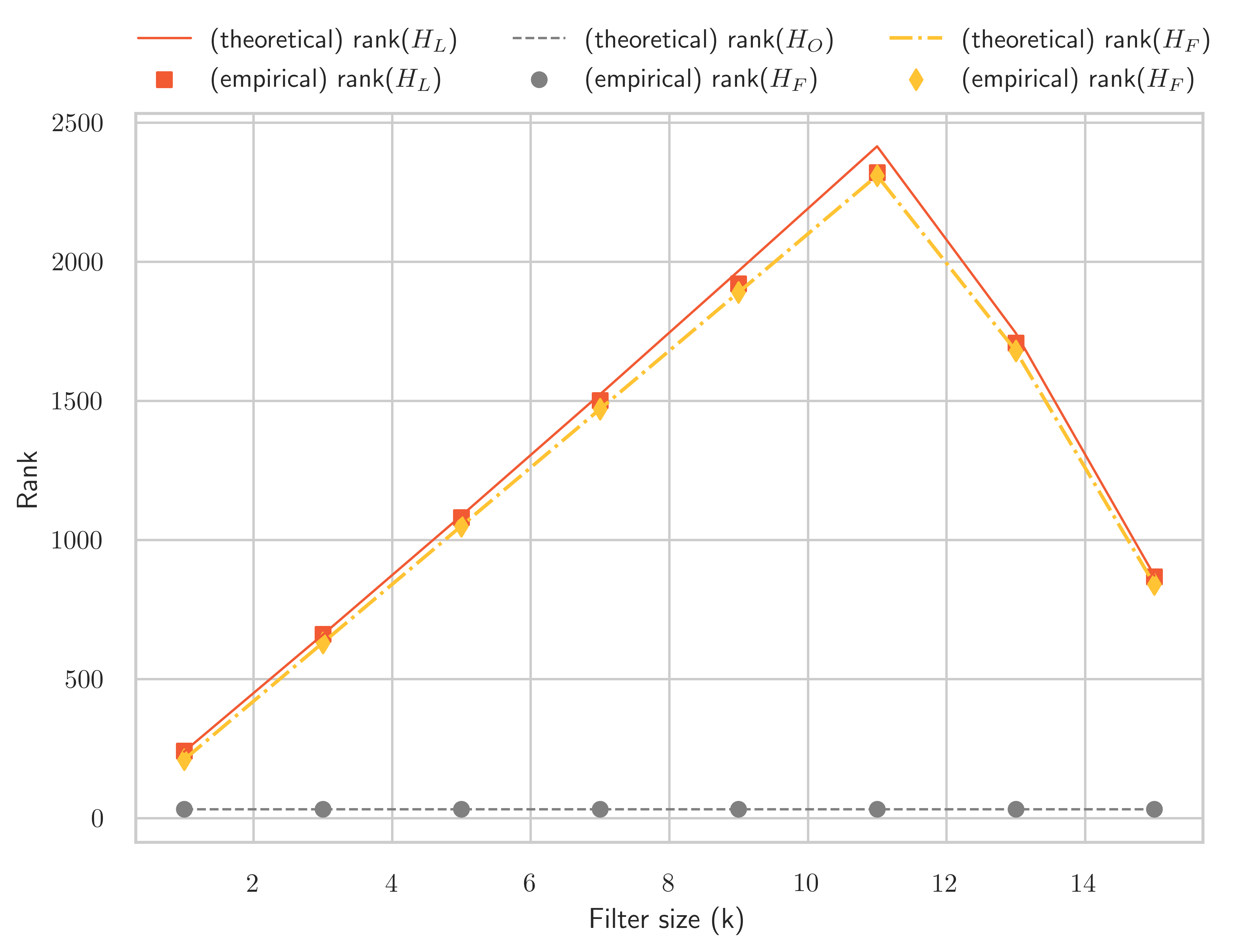}
	\caption{$m=105$}
\end{subfigure}
\begin{subfigure}[b]{0.3\textwidth}
	
	\includegraphics[width=\textwidth]{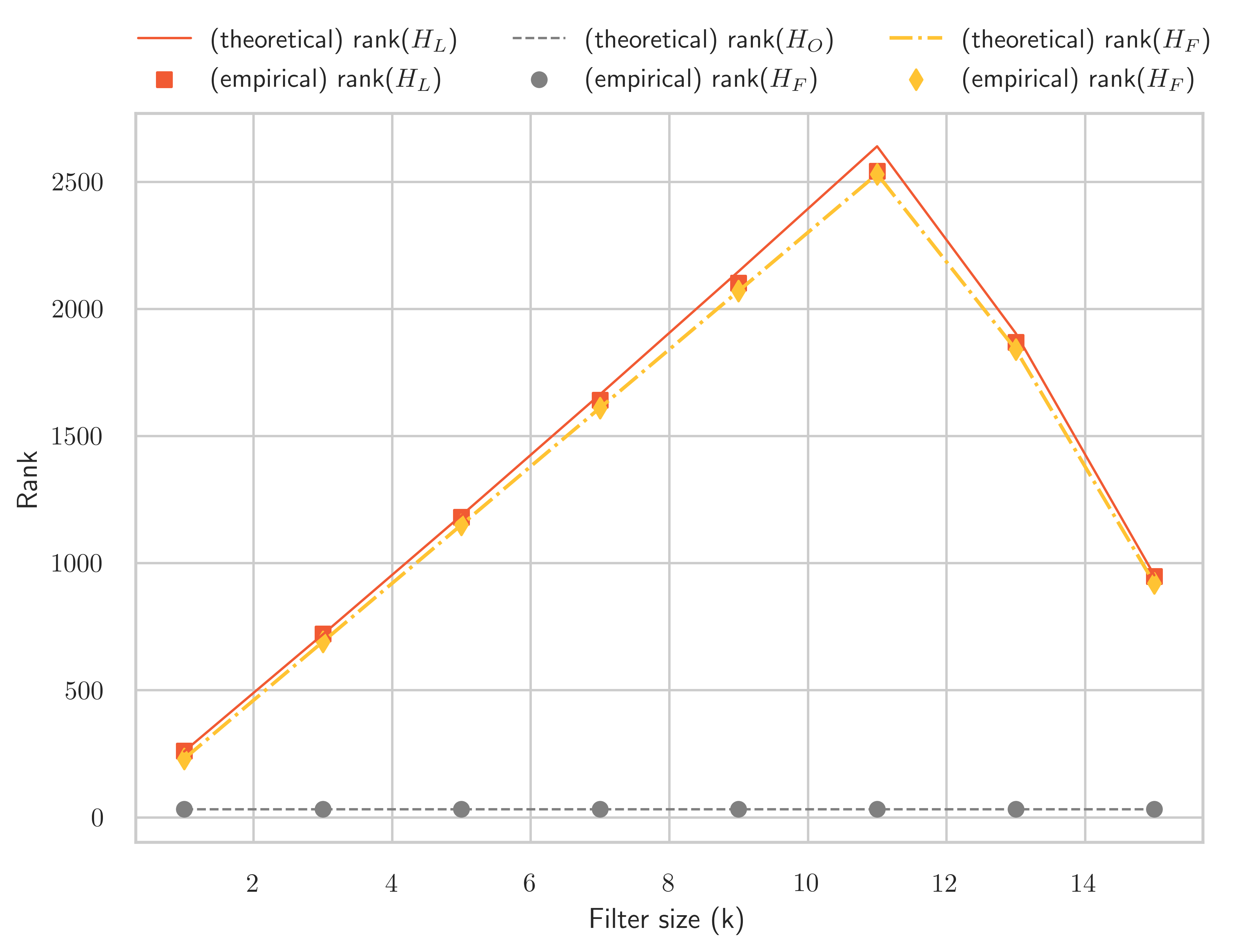}
	\caption{$m=115$}
\end{subfigure}
\begin{subfigure}[b]{0.3\textwidth}
	
	\includegraphics[width=\textwidth]{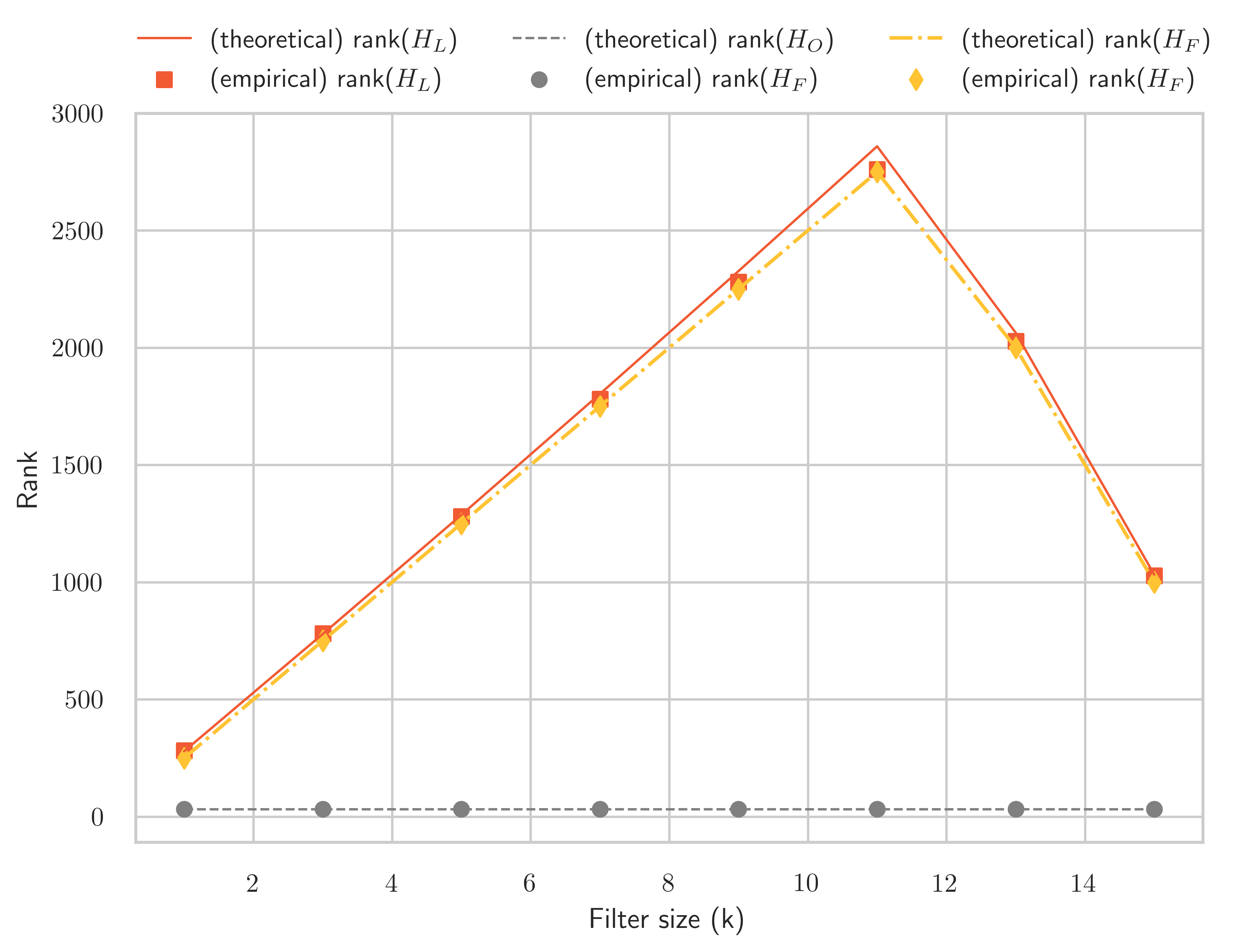}
	\caption{$m=125$}
\end{subfigure}
\begin{subfigure}[b]{0.3\textwidth}
	
	\includegraphics[width=\textwidth]{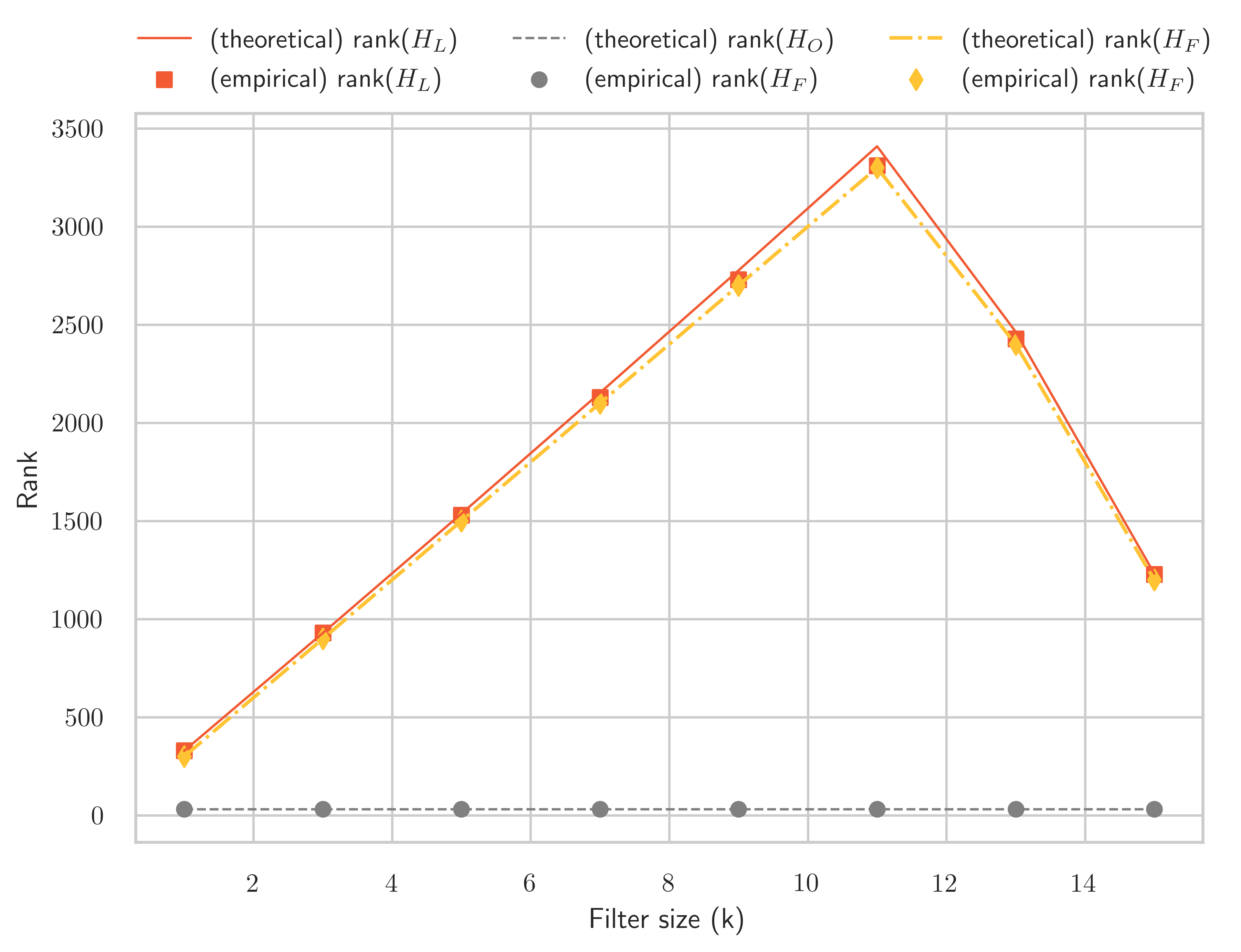}
	\caption{$m=150$}
\end{subfigure}
	\caption{Rank vs Filter Size for Linear CNN on CIFAR10 with Mean Squared Error loss. The upper bounds are reliable estimator of the true rank values. The rank of the functional and outer-product Hessian, as given by our upper bounds, are exact --- the lines and dots coincide perfectly in the plots. The rank of the loss Hessian upper bounded as the sum of the ranks of these two matrices is slightly loose, and this difference becomes negligible for $m\sim100$.}
\end{figure*}
\clearpage

\subsubsection{ReLU activations, MSE loss }

\begin{figure*}[!h]
	\centering
	\begin{subfigure}[b]{0.3\textwidth}
		
		\includegraphics[width=\textwidth]{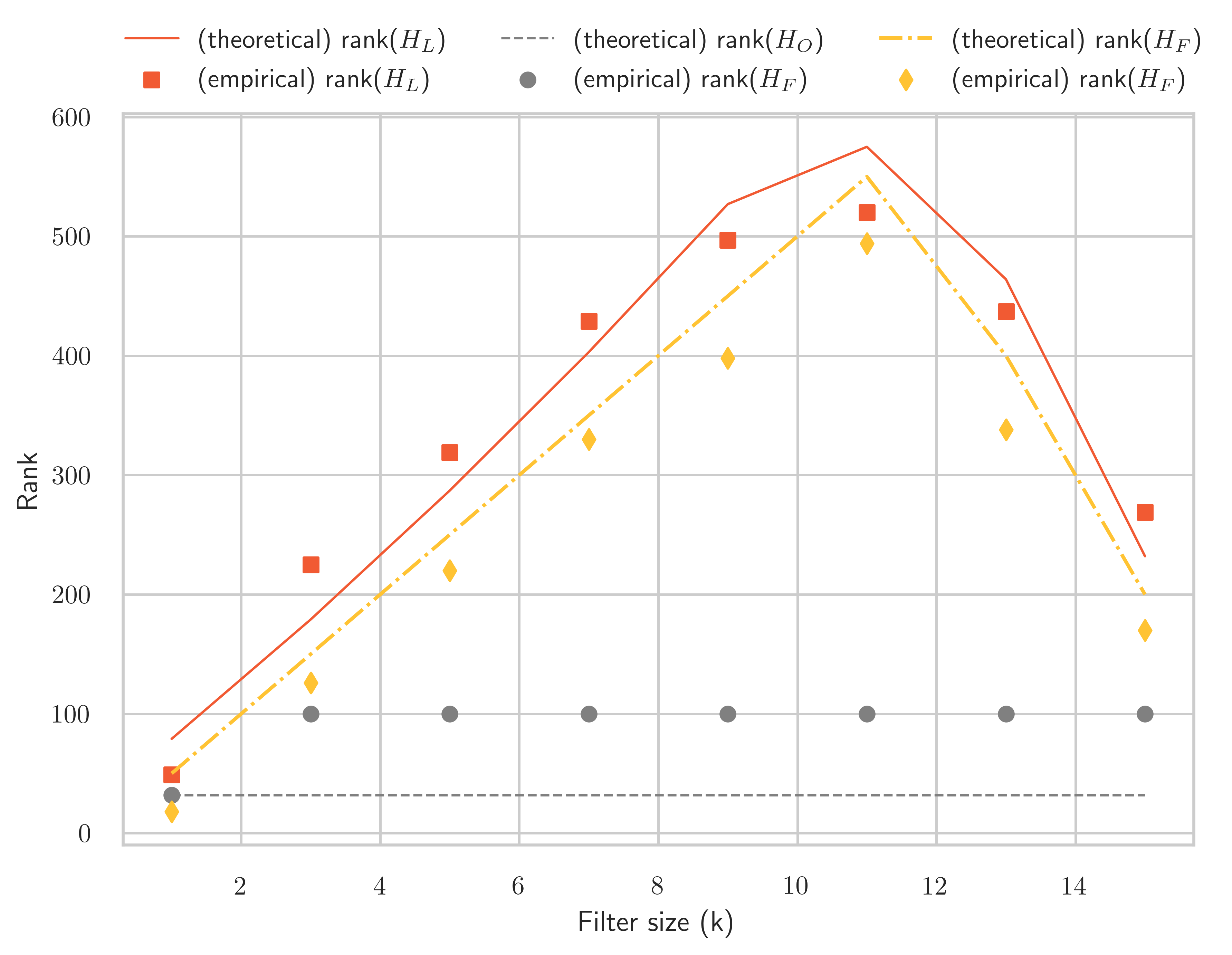}
		\caption{$m=25$}
	\end{subfigure}
	\begin{subfigure}[b]{0.3\textwidth}
		
		\includegraphics[width=\textwidth]{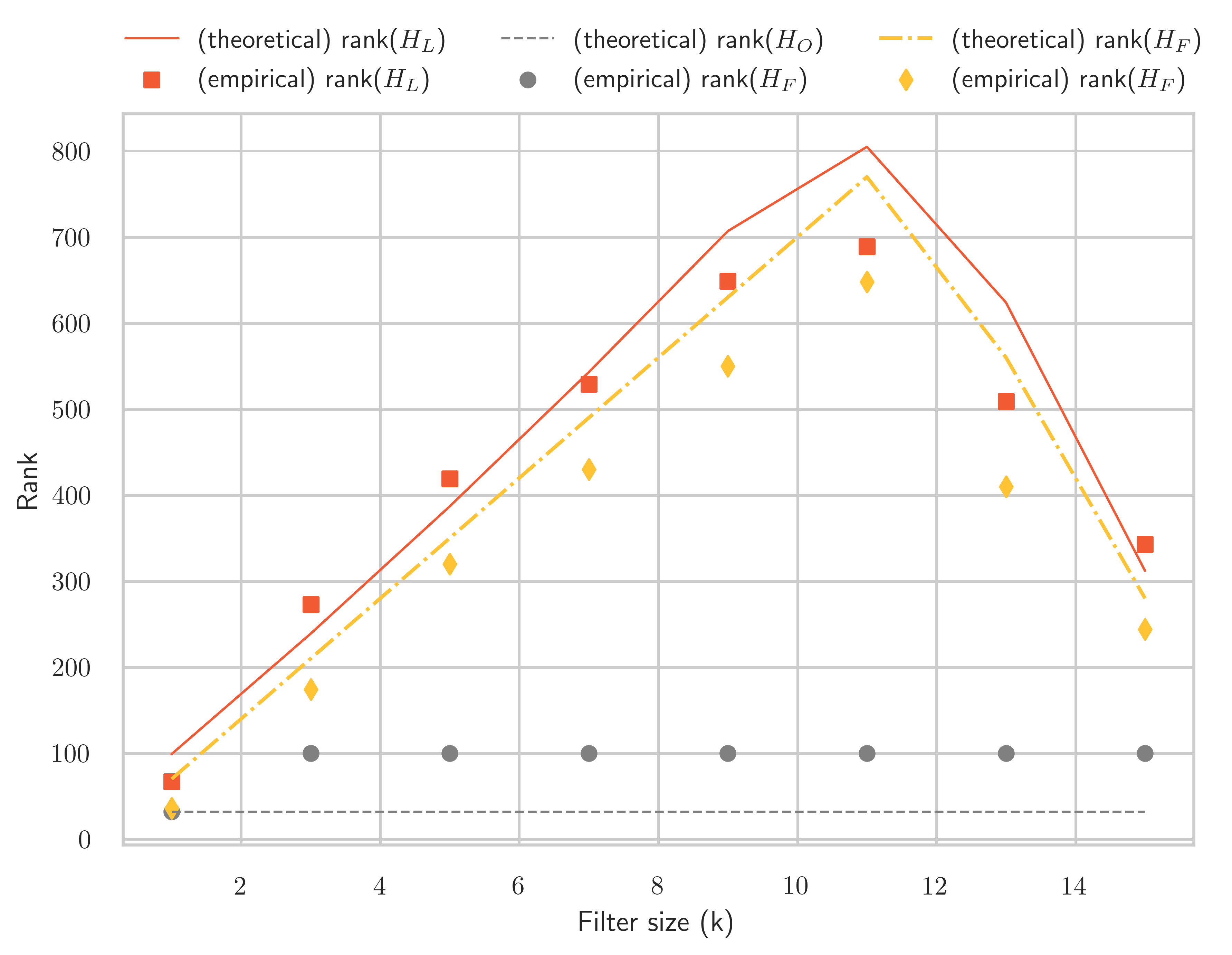}
		\caption{$m=35$}
	\end{subfigure}
	\begin{subfigure}[b]{0.3\textwidth}
		
		\includegraphics[width=\textwidth]{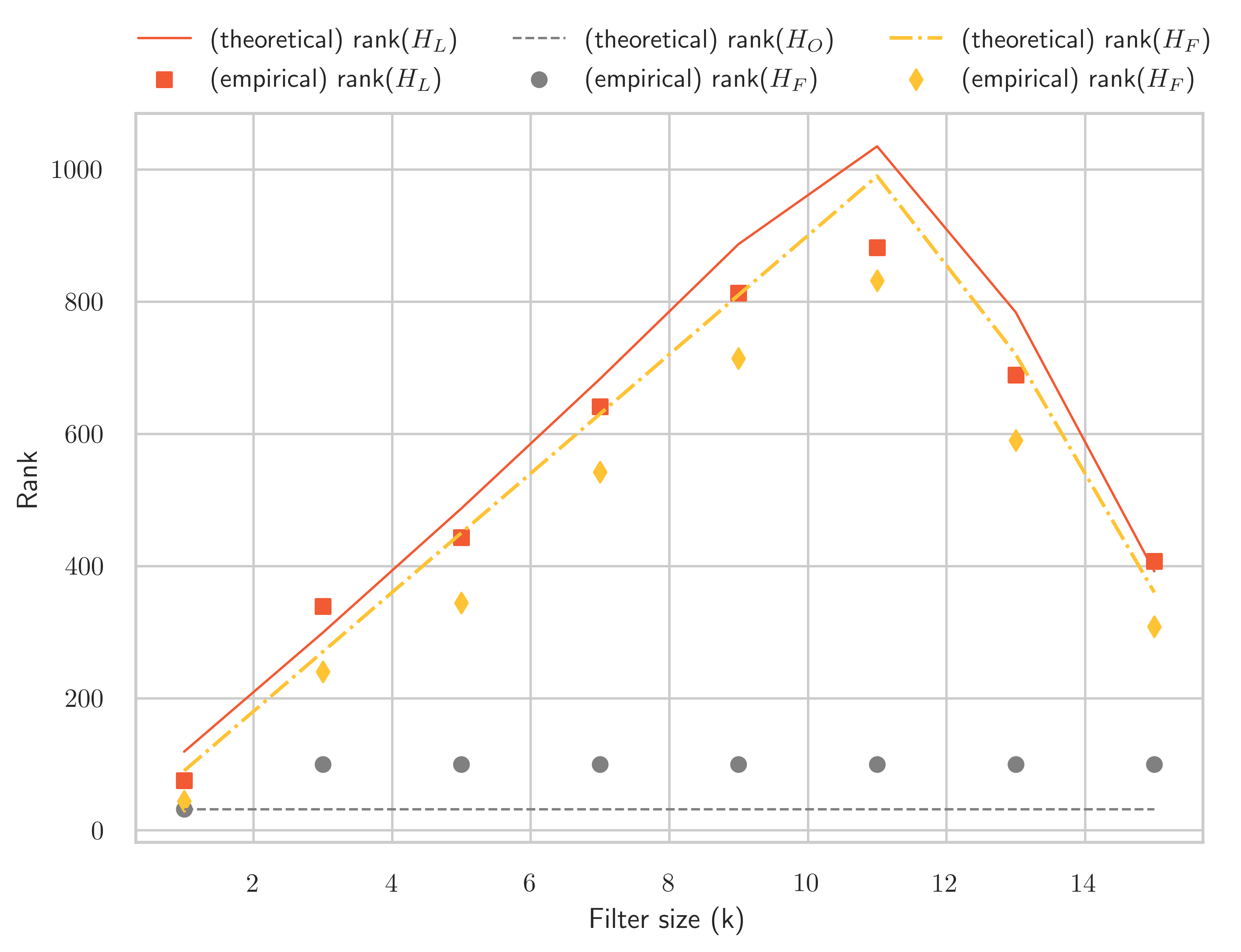}
		\caption{$m=45$}
	\end{subfigure}
	\begin{subfigure}[b]{0.3\textwidth}
		
		\includegraphics[width=\textwidth]{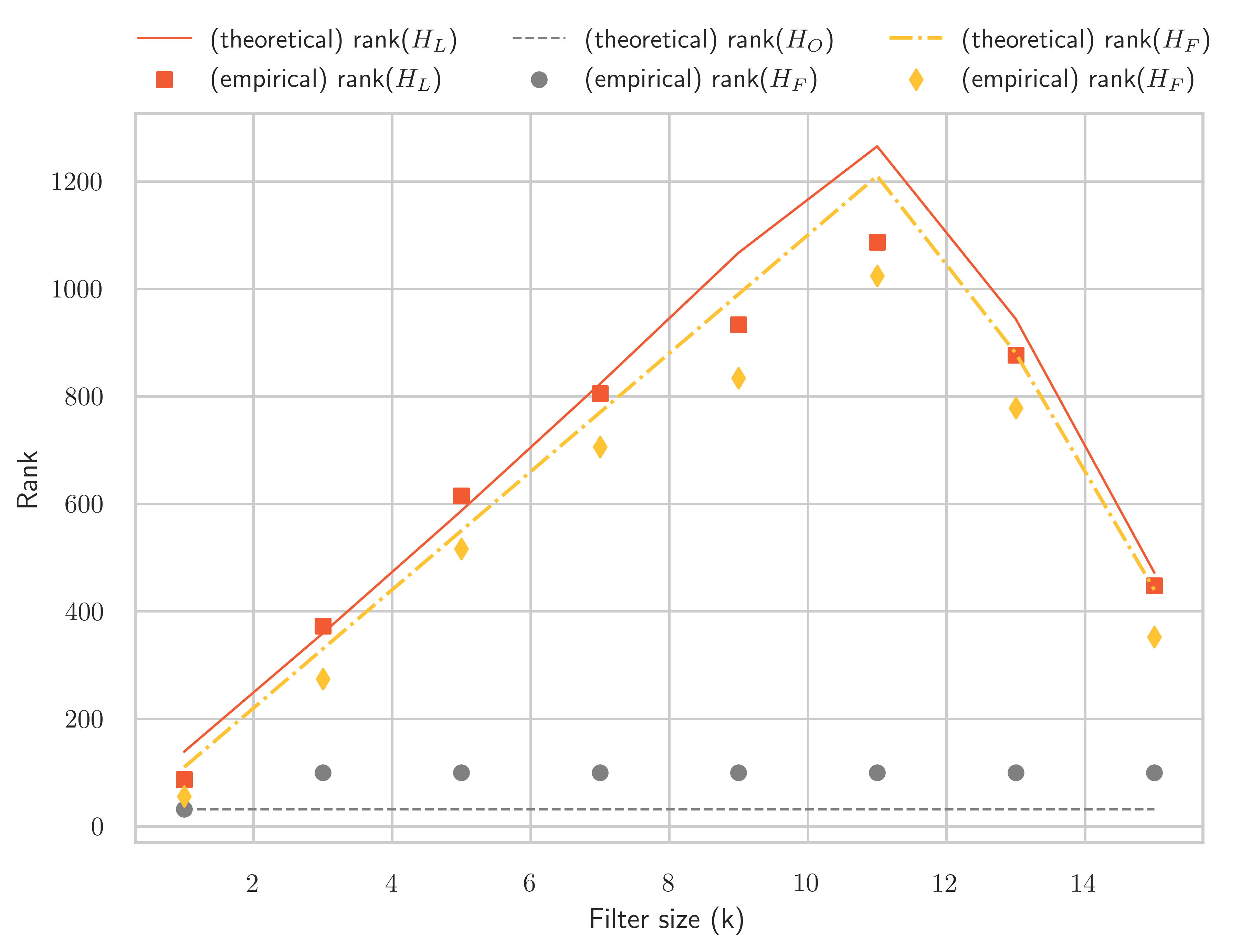}
		\caption{$m=55$}
	\end{subfigure}
	\begin{subfigure}[b]{0.3\textwidth}
		
		\includegraphics[width=\textwidth]{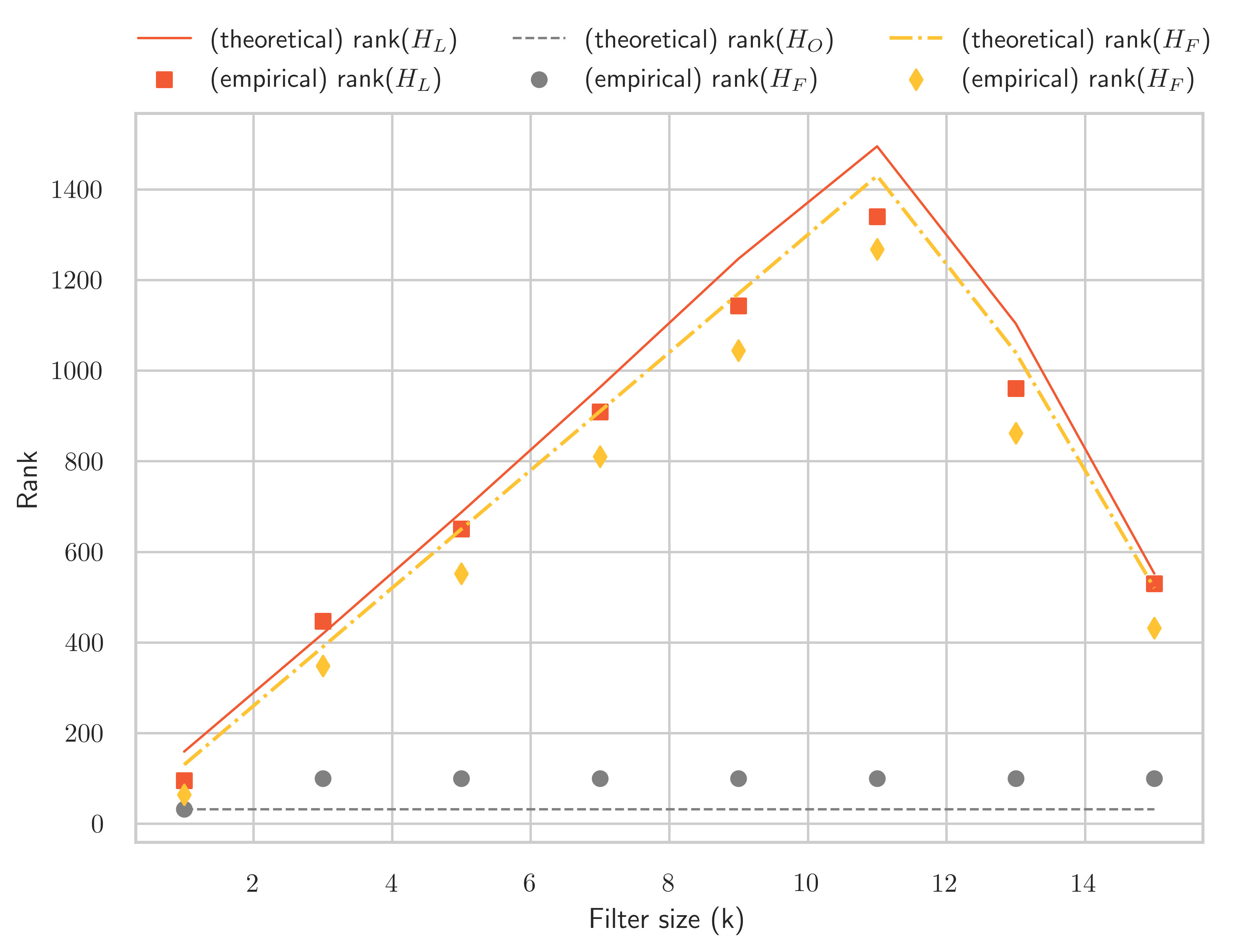}
		\caption{$m=65$}
	\end{subfigure}
	\begin{subfigure}[b]{0.3\textwidth}
		
		\includegraphics[width=\textwidth]{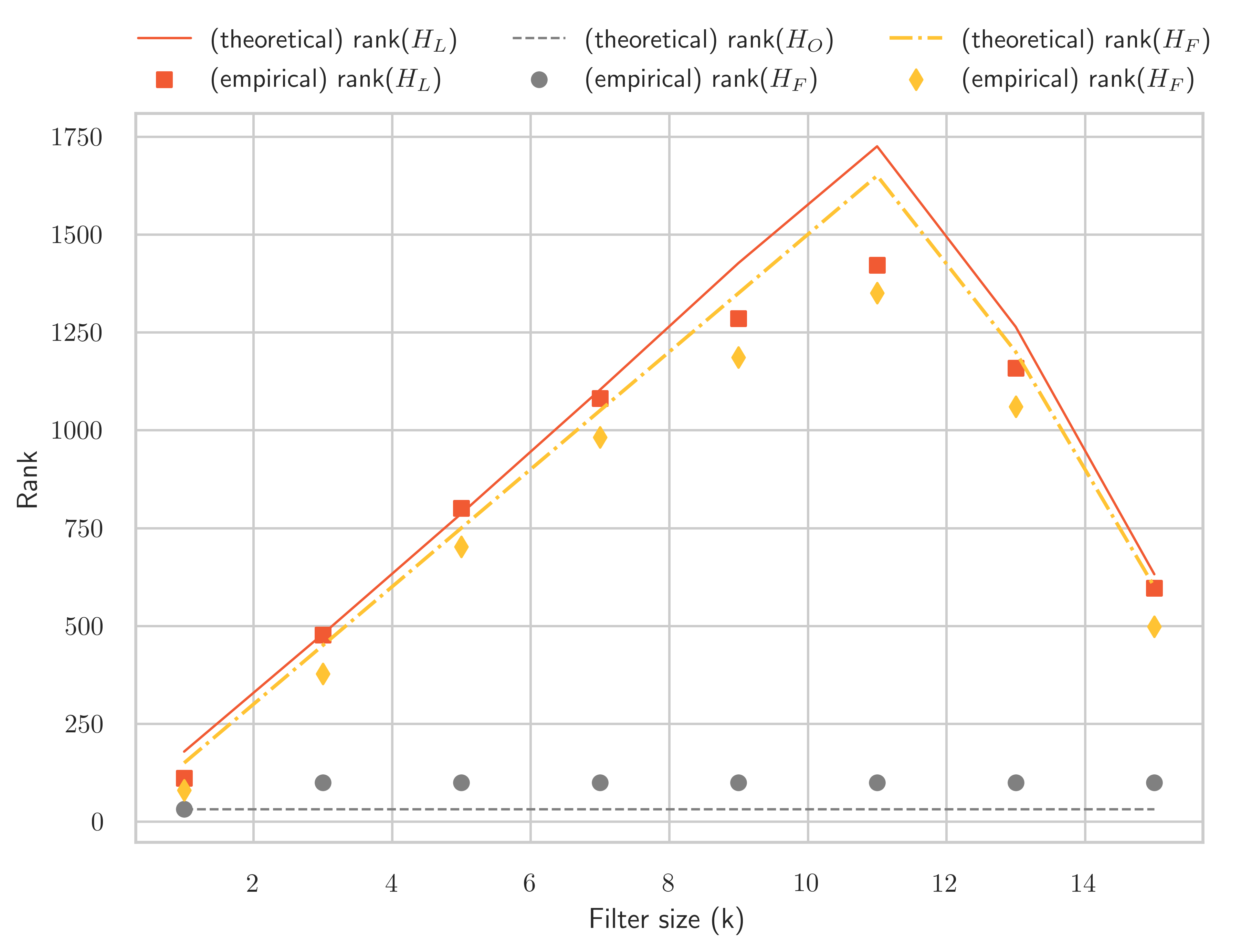}
		\caption{$m=75$}
	\end{subfigure}
	\begin{subfigure}[b]{0.3\textwidth}
		
		\includegraphics[width=\textwidth]{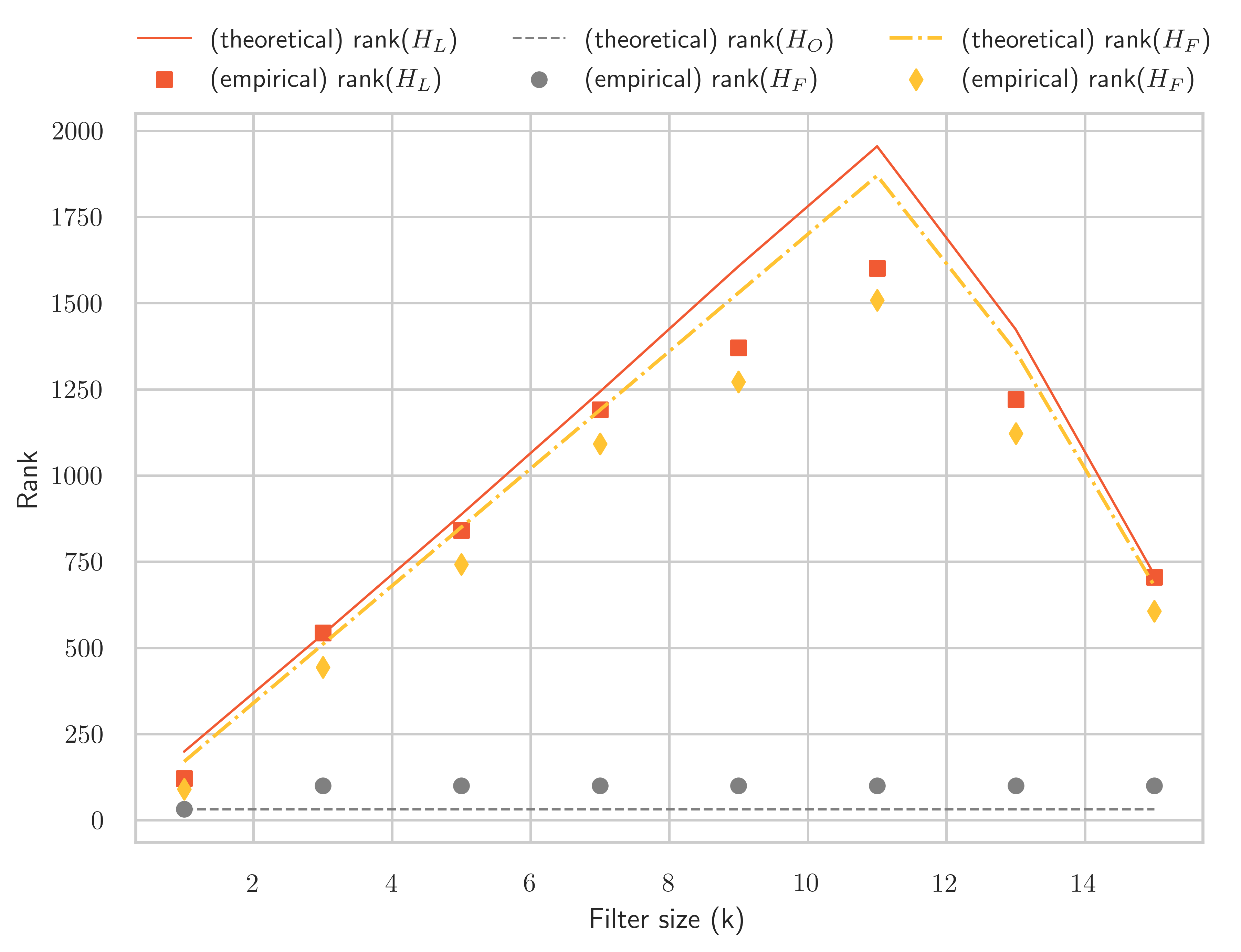}
		\caption{$m=85$}
	\end{subfigure}
	\begin{subfigure}[b]{0.3\textwidth}
		
		\includegraphics[width=\textwidth]{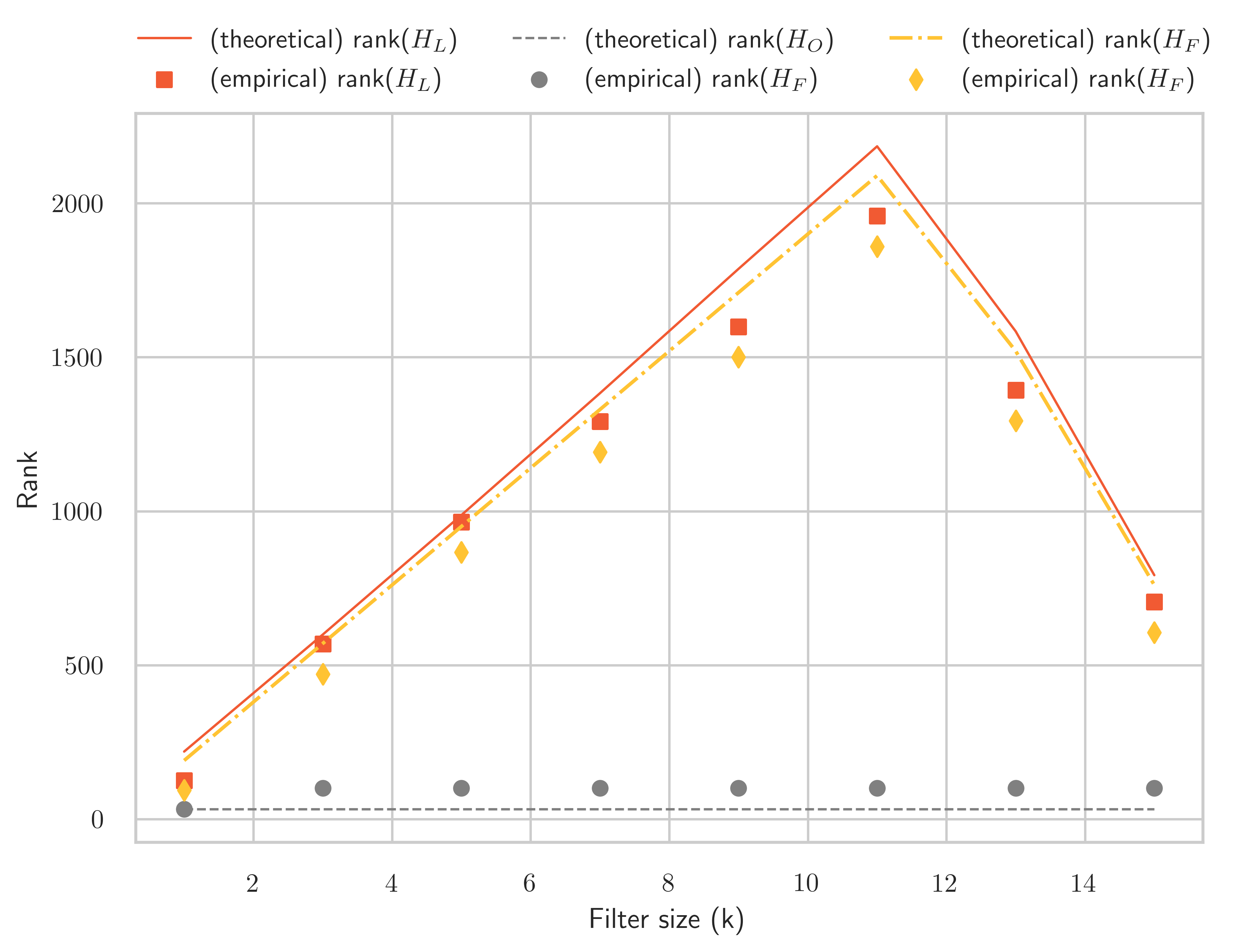}
		\caption{$m=95$}
	\end{subfigure}
	\begin{subfigure}[b]{0.3\textwidth}
		
		\includegraphics[width=\textwidth]{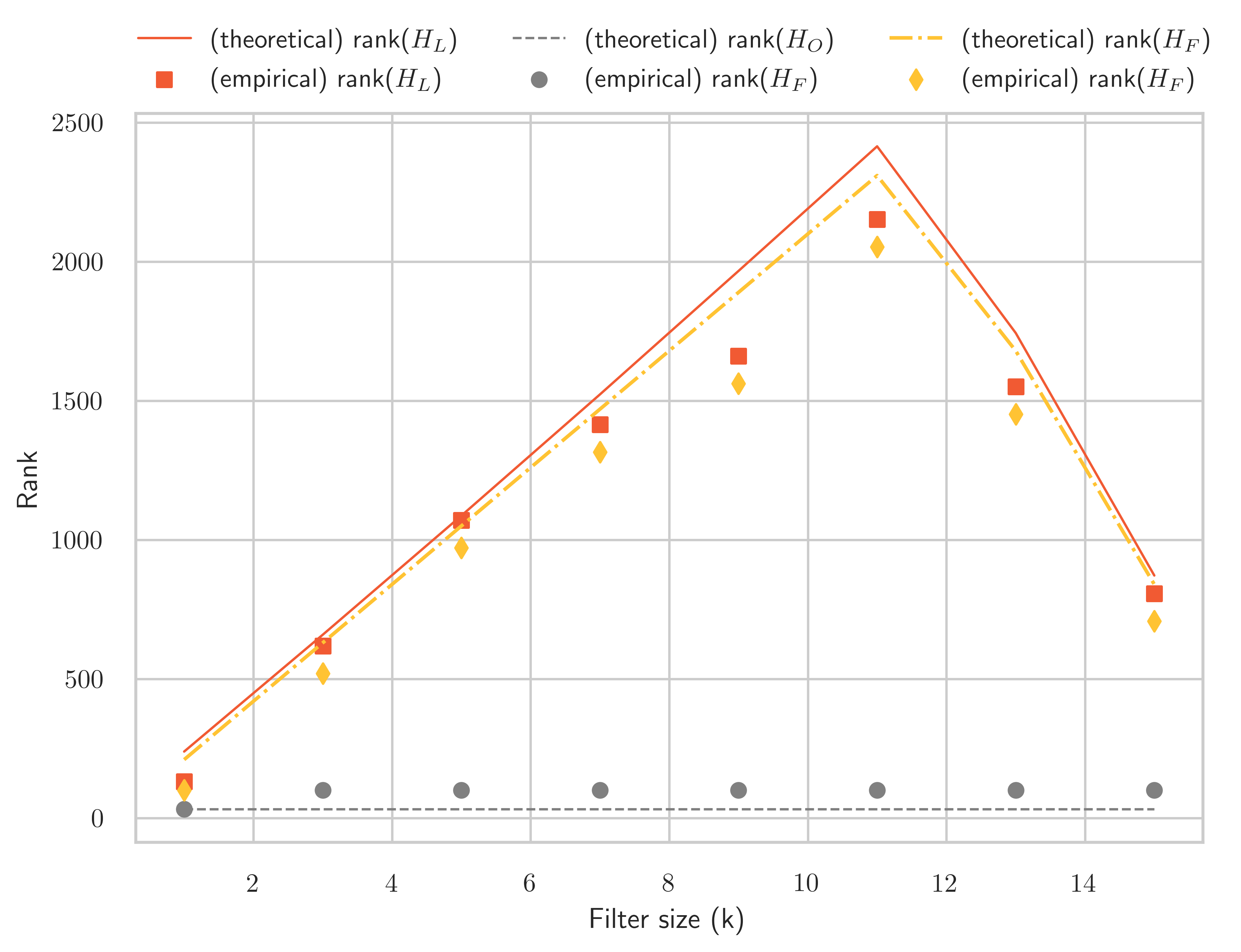}
		\caption{$m=105$}
	\end{subfigure}
	\begin{subfigure}[b]{0.3\textwidth}
		
		\includegraphics[width=\textwidth]{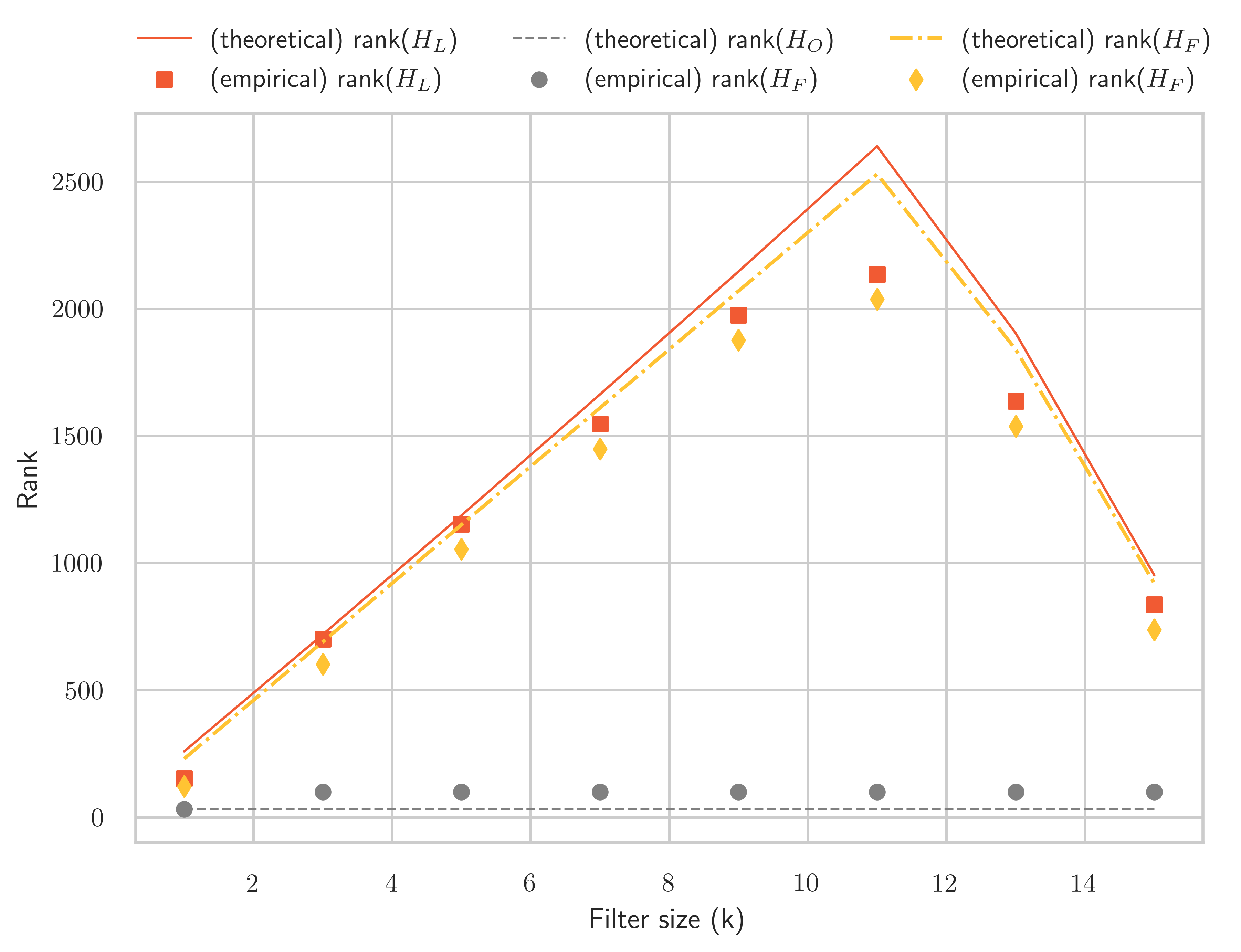}
		\caption{$m=115$}
	\end{subfigure}
	\begin{subfigure}[b]{0.3\textwidth}
		
		\includegraphics[width=\textwidth]{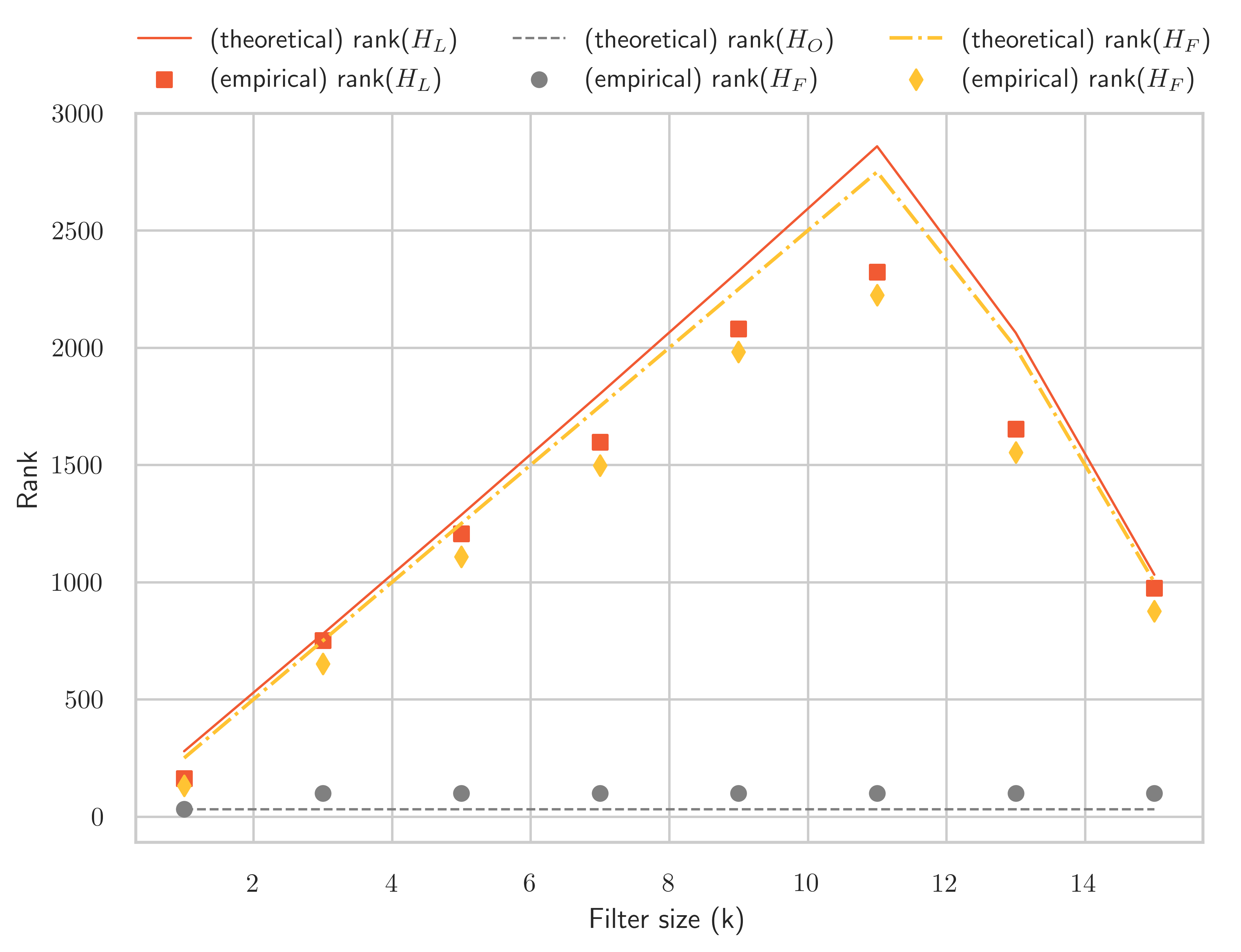}
		\caption{$m=125$}
	\end{subfigure}
	\begin{subfigure}[b]{0.3\textwidth}
		
		\includegraphics[width=\textwidth]{figures/filter_size/jaxcifar10_mse_relu_m-150_d-16/rank_vs_filtersize.png}
		\caption{$m=150$}
	\end{subfigure}
	\caption{Rank vs Filter Size for ReLU CNN on CIFAR10 with Mean Squared Error loss. The upper bounds are reliable estimator of the true rank values. The rank of the functional and outer-product Hessian, as given by our upper bounds, are exact --- the lines and dots coincide perfectly in the plots. The rank of the loss Hessian upper bounded as the sum of the ranks of these two matrices is slightly loose, and this difference becomes negligible for $m\sim100$.}
\end{figure*}
\clearpage
\subsubsection{ReLU  activations, CE loss }
\begin{figure*}[!h]
	\centering
	\begin{subfigure}[b]{0.3\textwidth}
		
		\includegraphics[width=\textwidth]{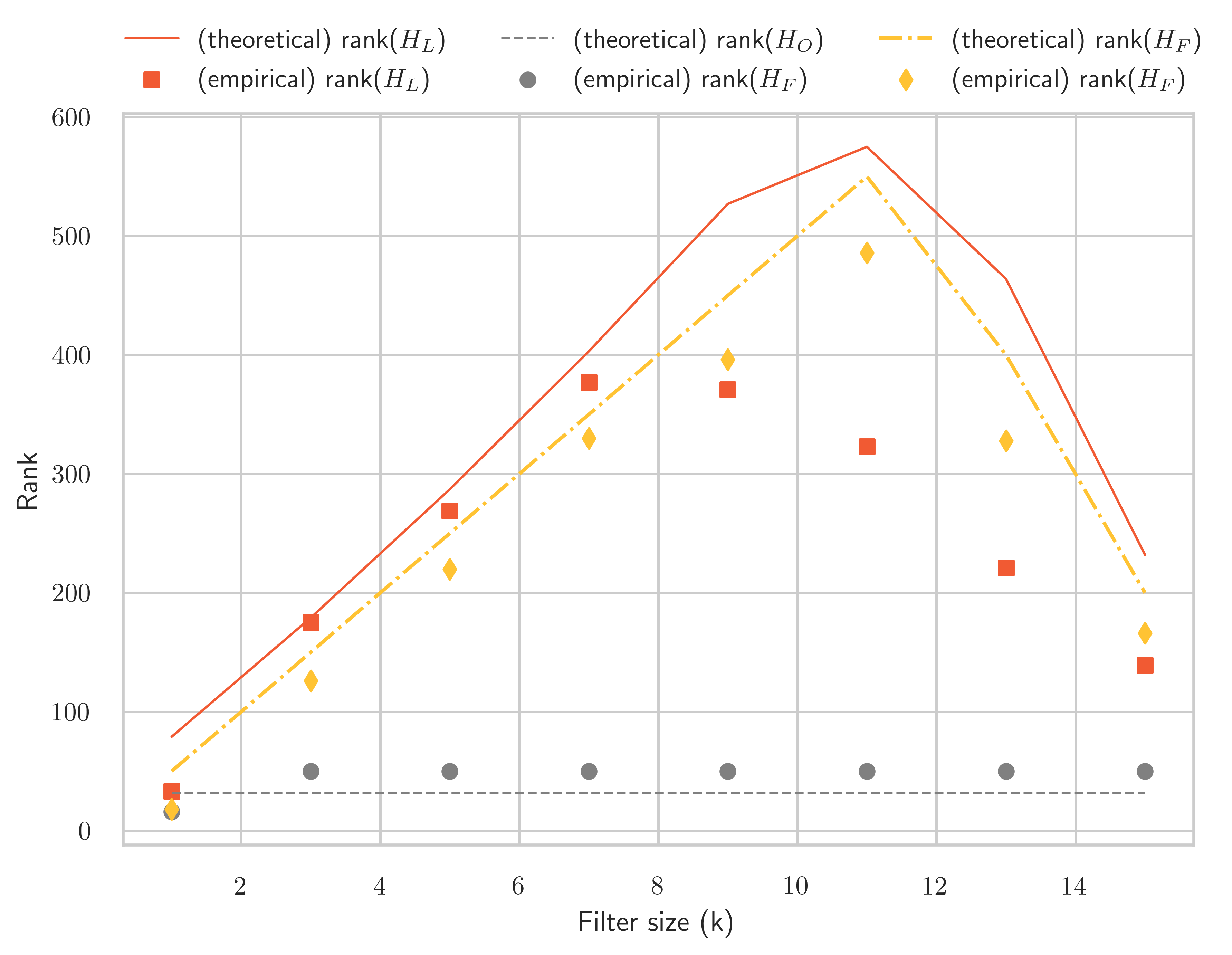}
		\caption{$m=25$}
	\end{subfigure}
	\begin{subfigure}[b]{0.3\textwidth}
		
		\includegraphics[width=\textwidth]{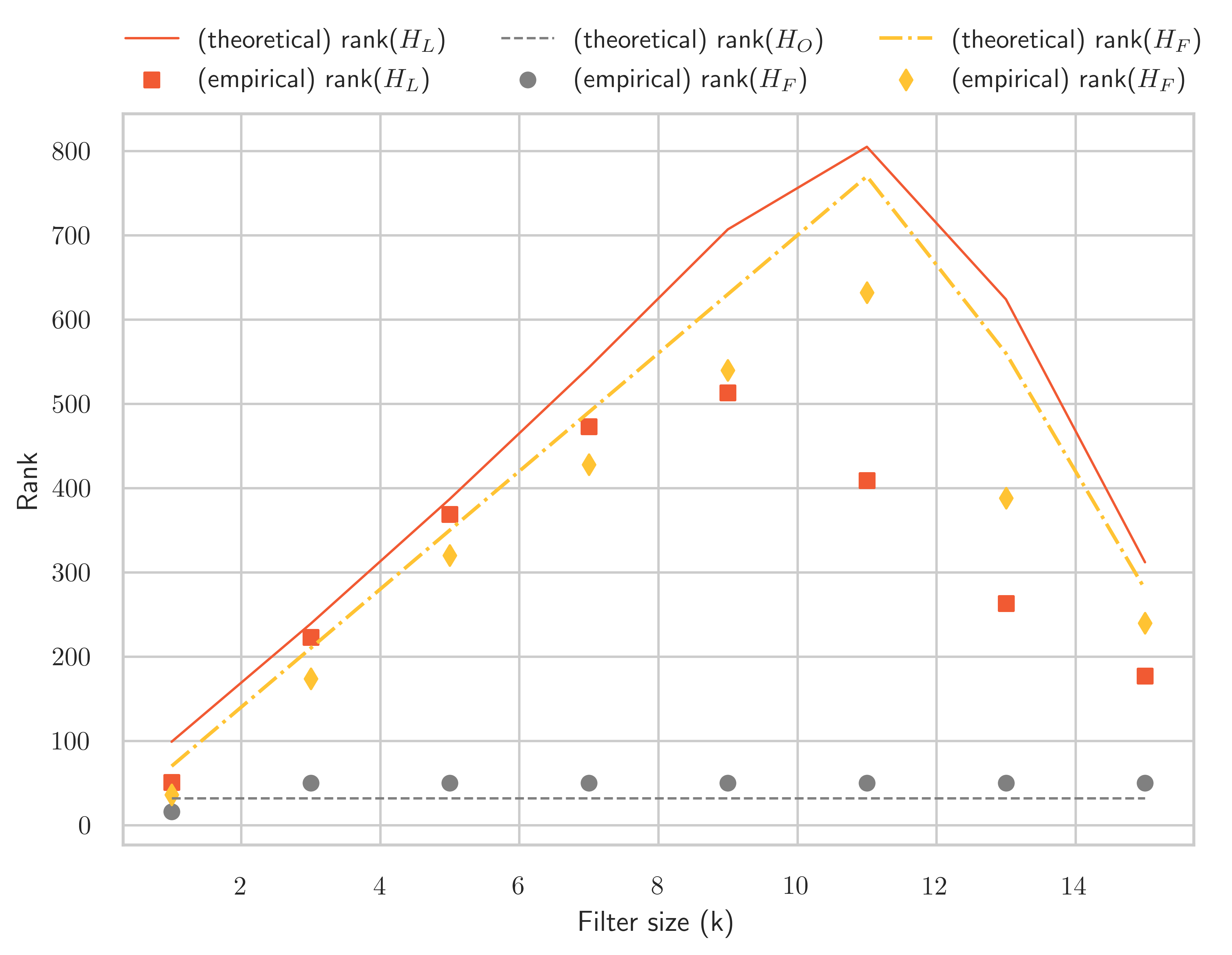}
		\caption{$m=35$}
	\end{subfigure}
	\begin{subfigure}[b]{0.3\textwidth}
		
		\includegraphics[width=\textwidth]{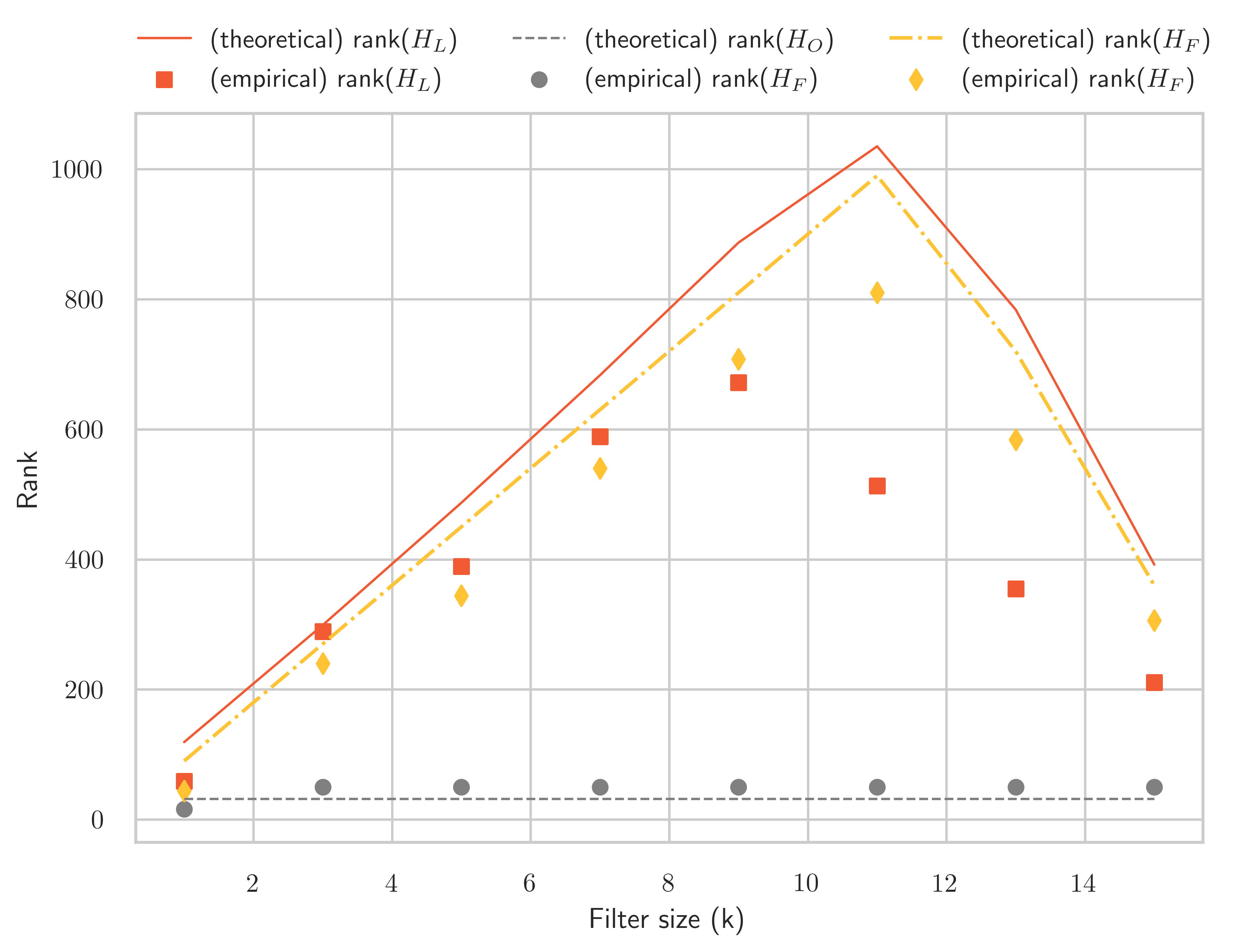}
		\caption{$m=45$}
	\end{subfigure}
	\begin{subfigure}[b]{0.3\textwidth}
		
		\includegraphics[width=\textwidth]{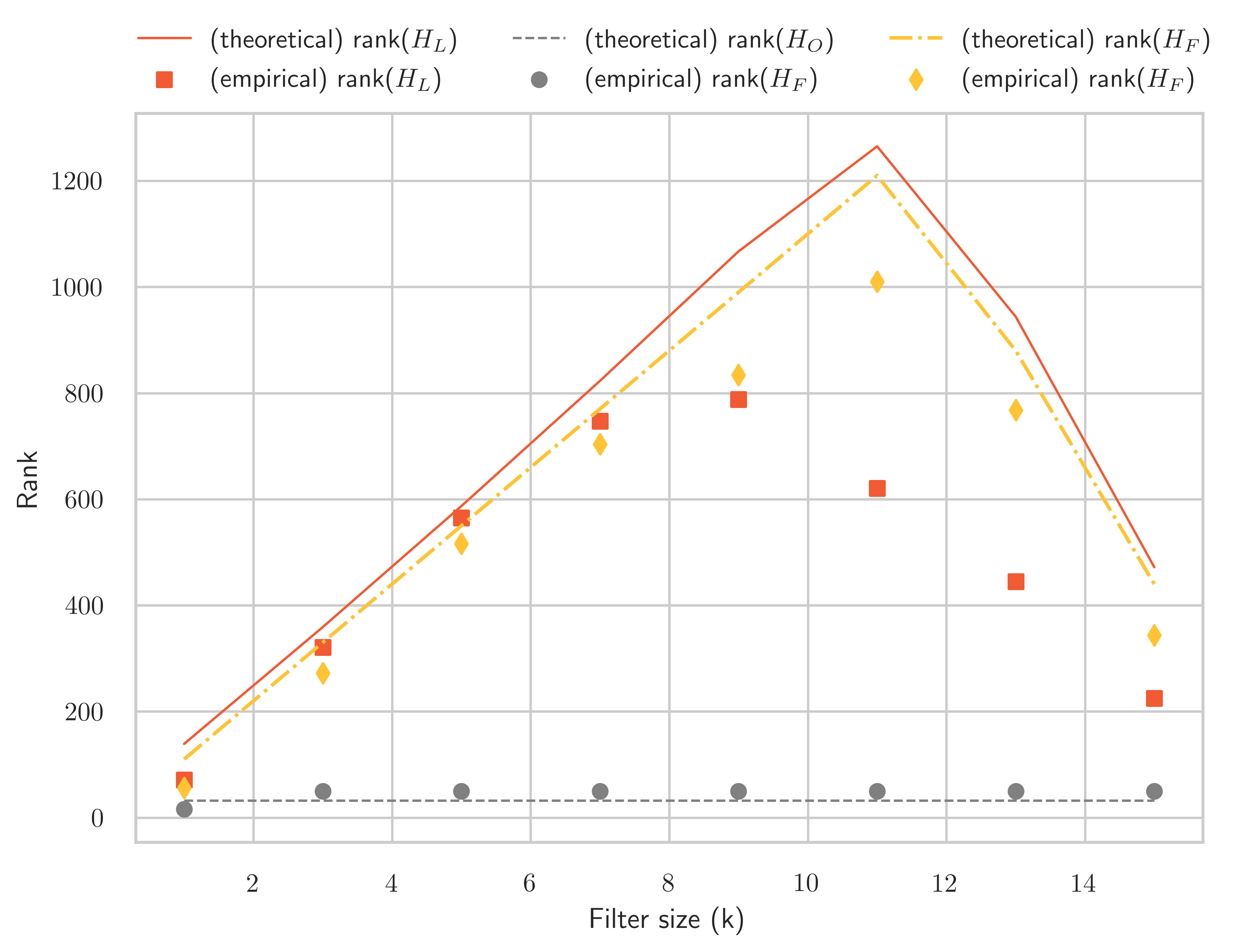}
		\caption{$m=55$}
	\end{subfigure}
	\begin{subfigure}[b]{0.3\textwidth}
		
		\includegraphics[width=\textwidth]{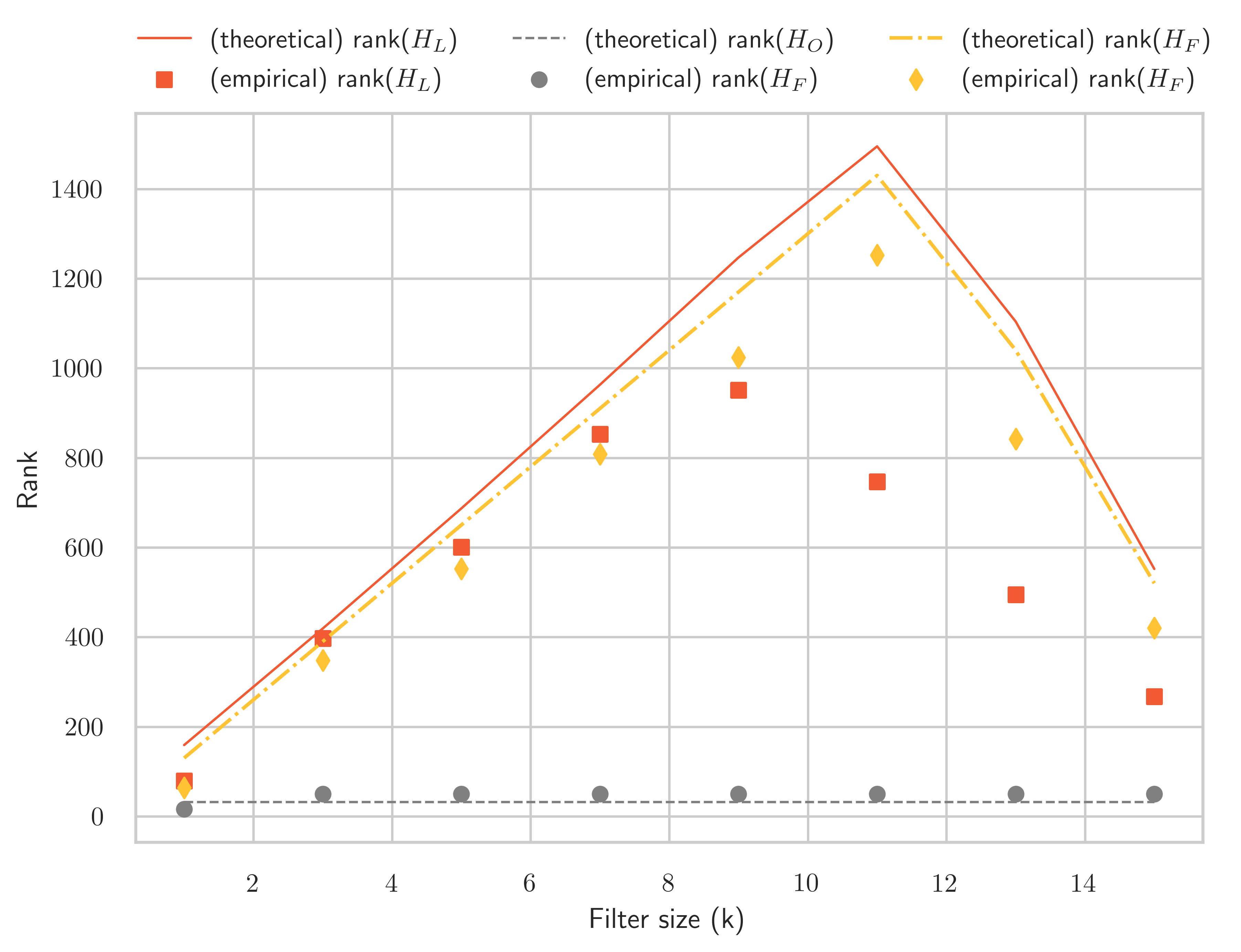}
		\caption{$m=65$}
	\end{subfigure}
	\begin{subfigure}[b]{0.3\textwidth}
		
		\includegraphics[width=\textwidth]{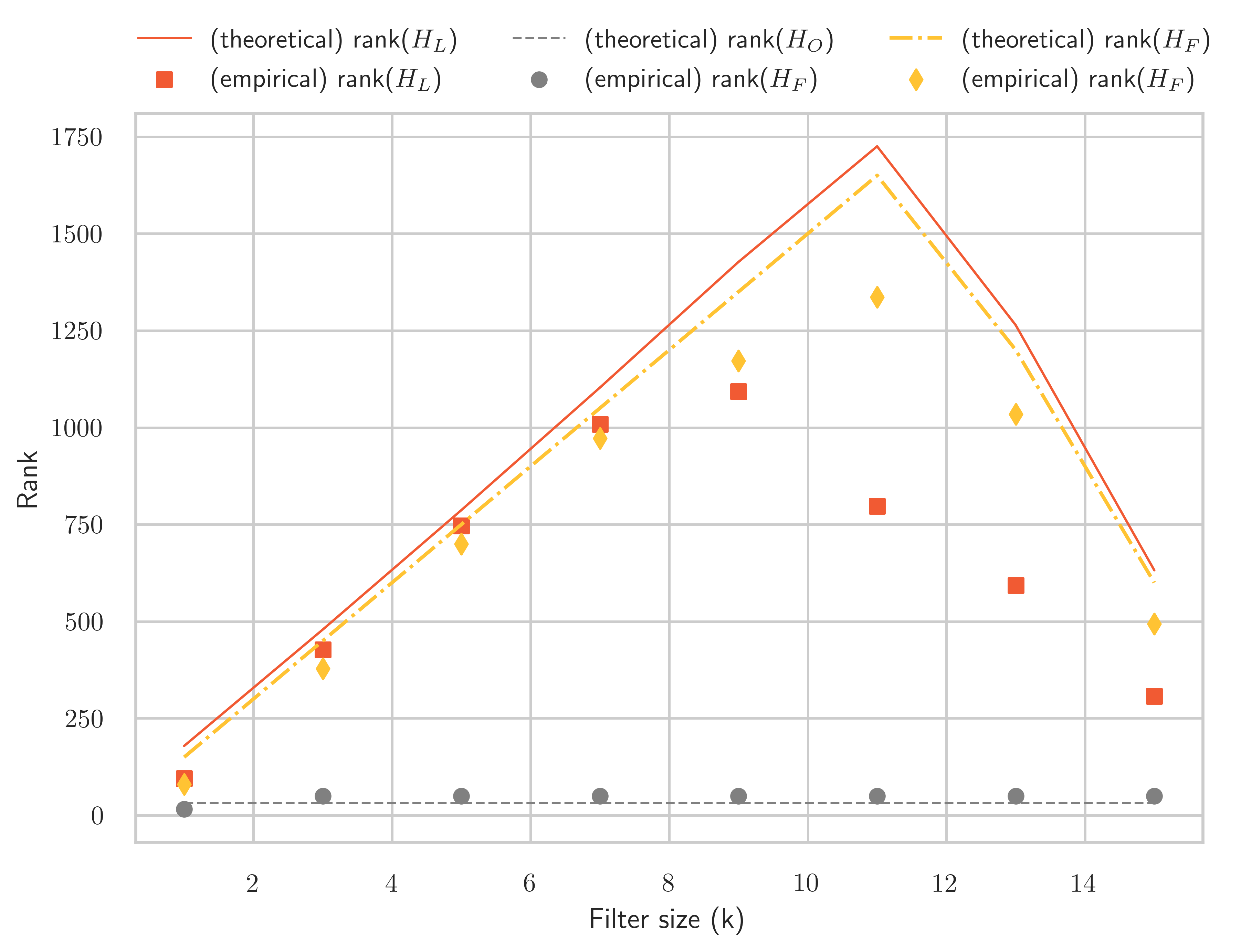}
		\caption{$m=75$}
	\end{subfigure}
	\begin{subfigure}[b]{0.3\textwidth}
		
		\includegraphics[width=\textwidth]{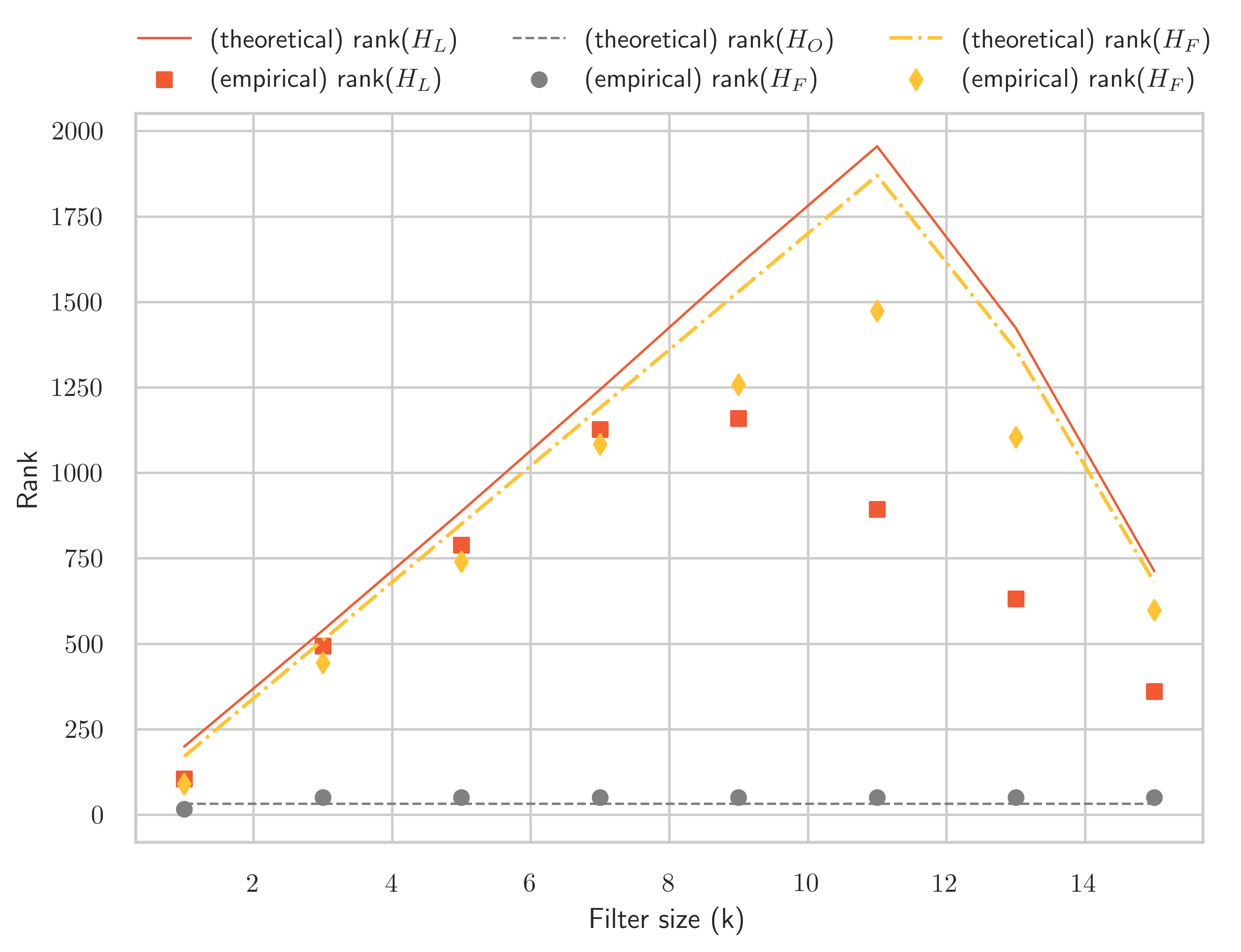}
		\caption{$m=85$}
	\end{subfigure}
	\begin{subfigure}[b]{0.3\textwidth}
		
		\includegraphics[width=\textwidth]{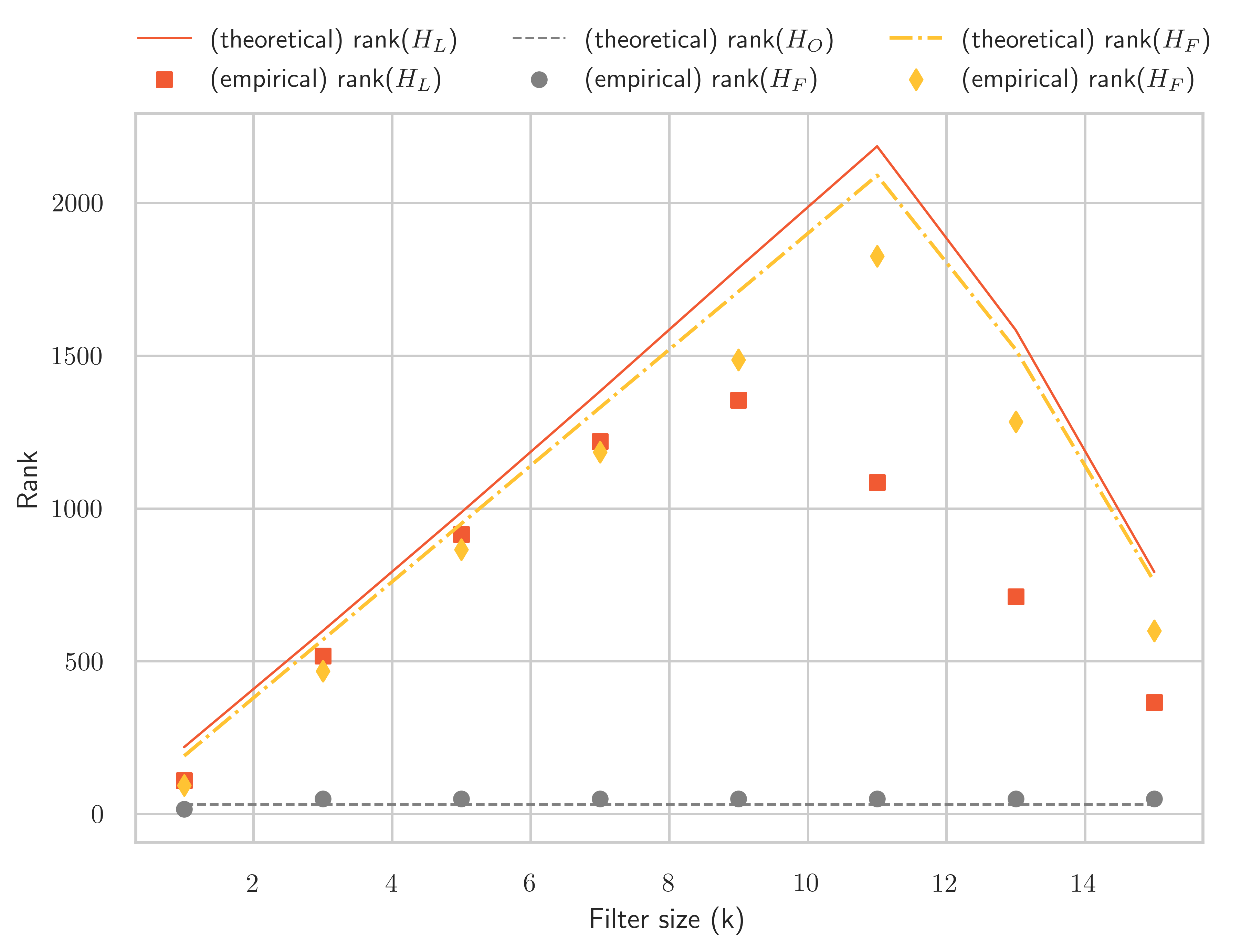}
		\caption{$m=95$}
	\end{subfigure}
	\begin{subfigure}[b]{0.3\textwidth}
		
		\includegraphics[width=\textwidth]{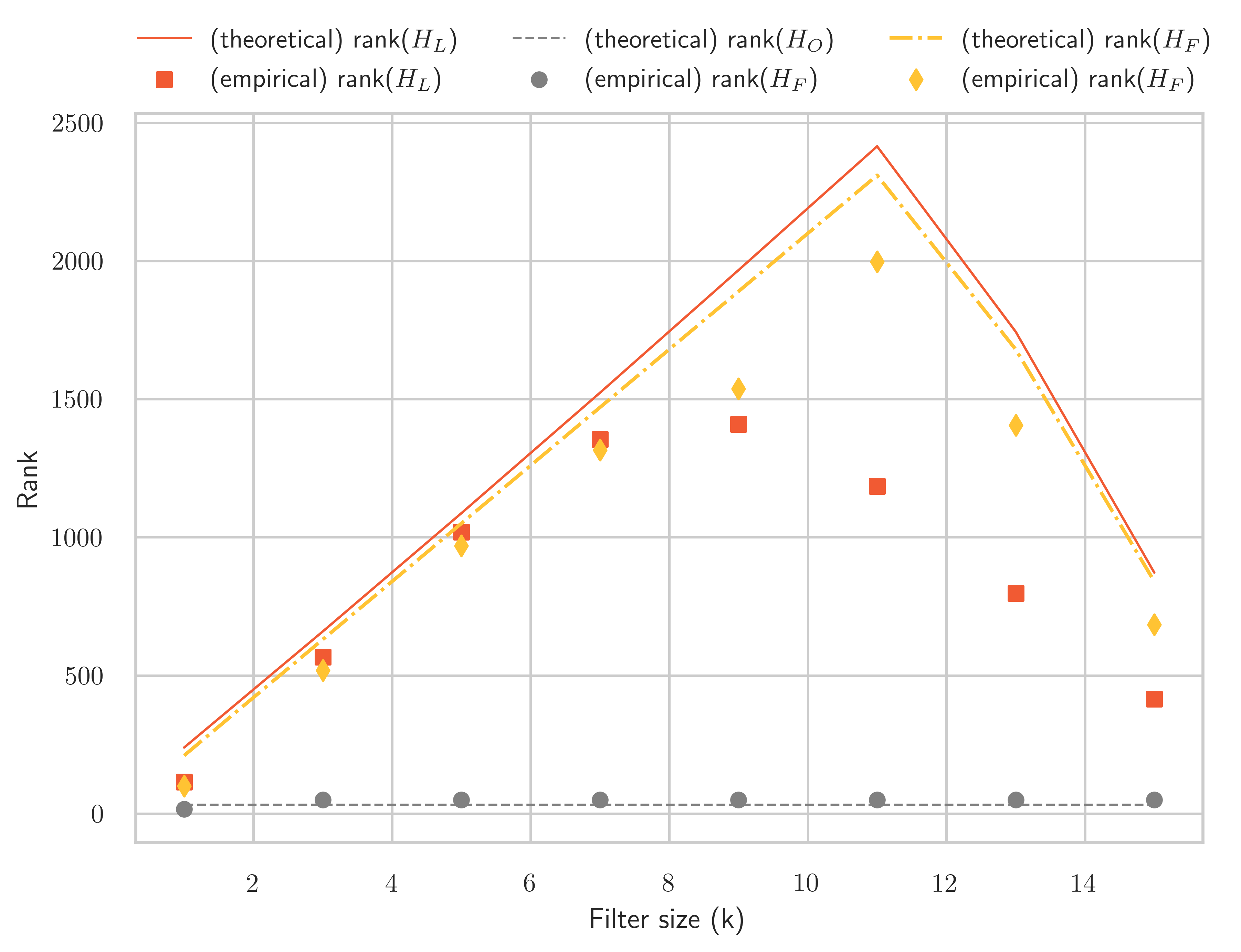}
		\caption{$m=105$}
	\end{subfigure}
	\begin{subfigure}[b]{0.3\textwidth}
		
		\includegraphics[width=\textwidth]{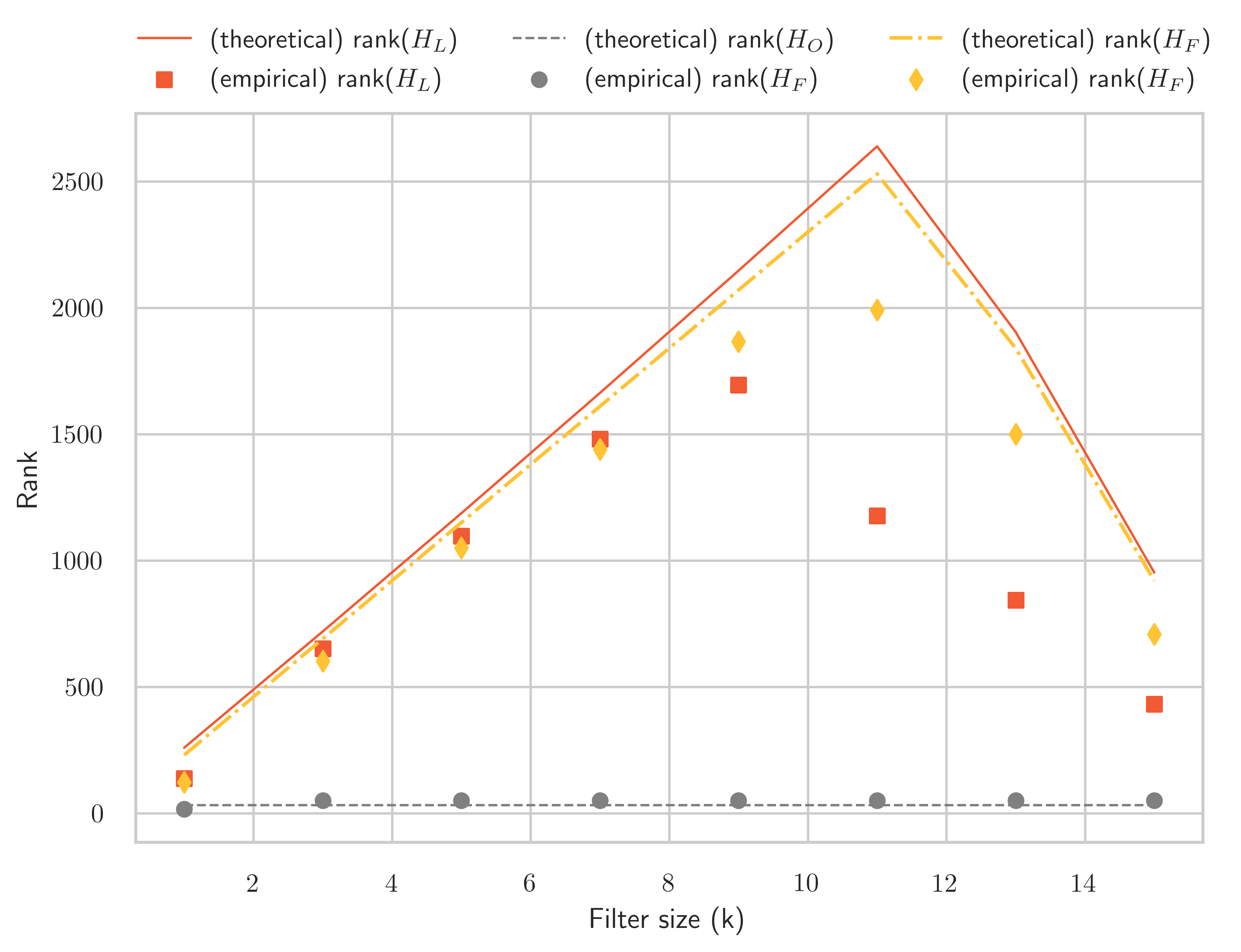}
		\caption{$m=115$}
	\end{subfigure}
	\begin{subfigure}[b]{0.3\textwidth}
		
		\includegraphics[width=\textwidth]{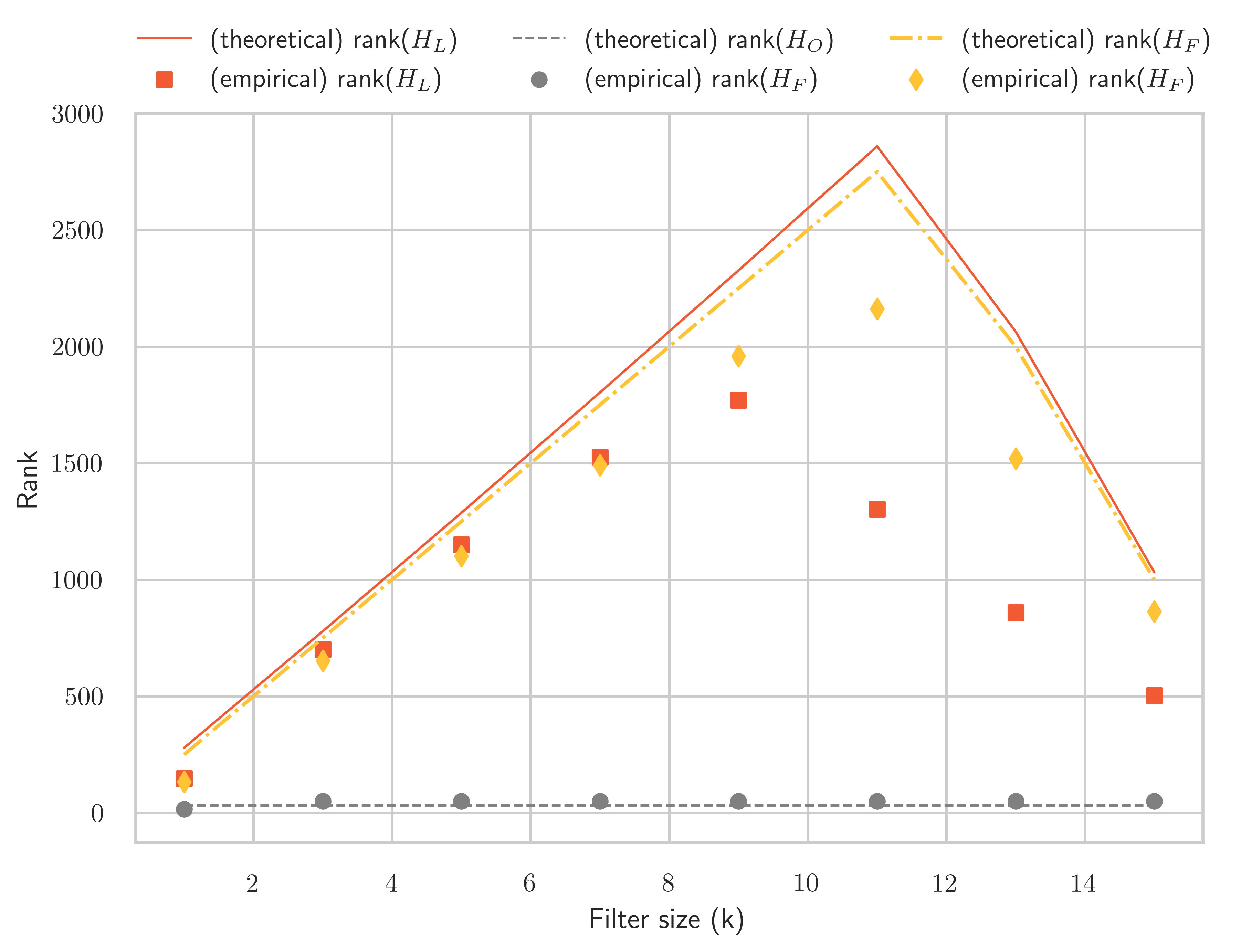}
		\caption{$m=125$}
	\end{subfigure}
	\begin{subfigure}[b]{0.3\textwidth}
		
		\includegraphics[width=\textwidth]{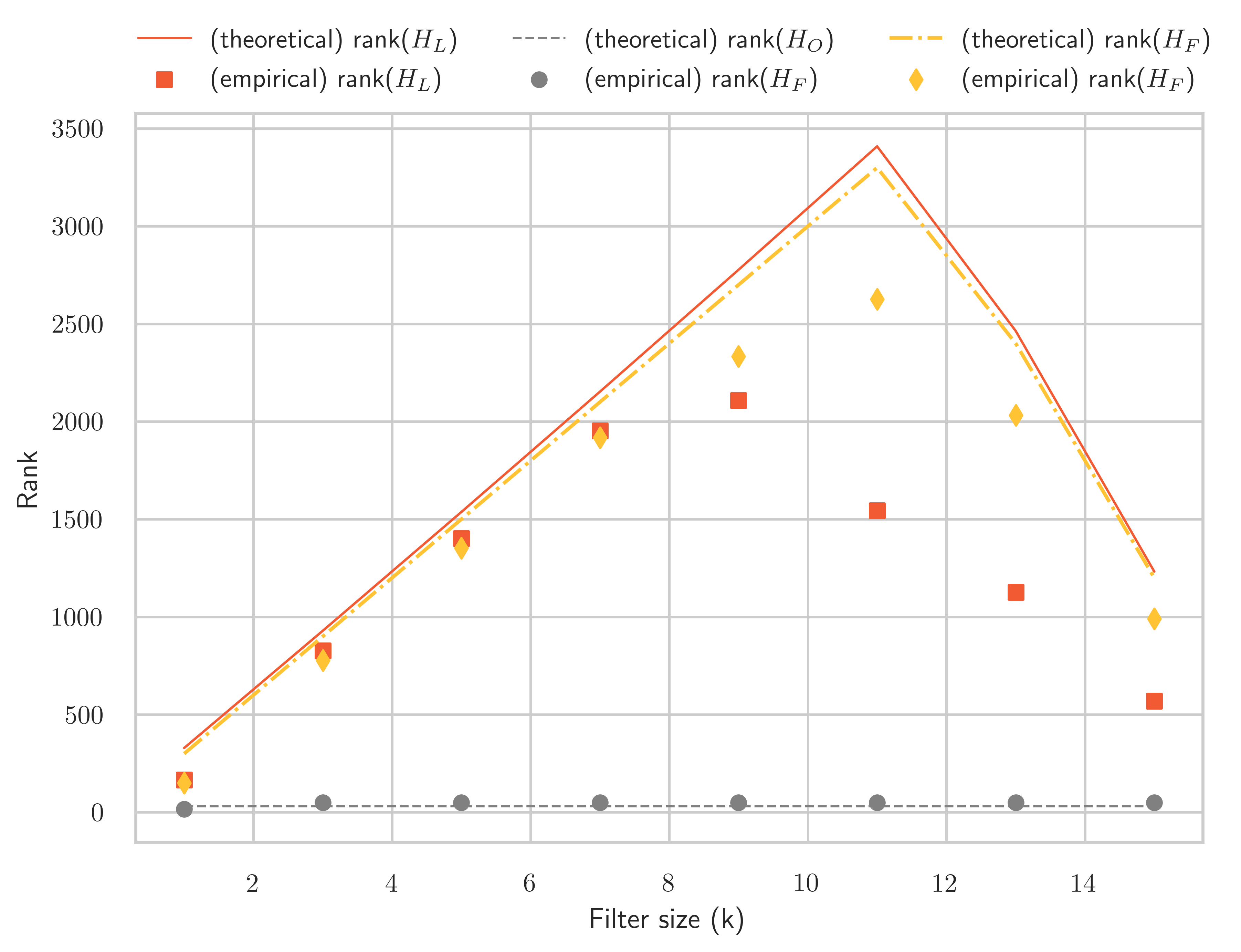}
		\caption{$m=150$}
	\end{subfigure}
	\caption{Rank vs Filter Size for ReLU CNN on CIFAR10 with Cross Entropy loss. The upper bounds are reliable estimator of the true rank values. The rank of the functional and outer-product Hessian, as given by our upper bounds, are exact --- the lines and dots coincide perfectly in the plots. The rank of the loss Hessian upper bounded as the sum of the ranks of these two matrices is slightly loose, and this difference becomes negligible for $m\sim100$.}
\end{figure*}
\clearpage
\subsection{LCN vs LCN + WS}
This comparison is done based on Linear activations with MSE loss as a verification of our theory and the dataset used is CIFAR10.
\begin{figure*}[!h]
	\centering
	\begin{subfigure}[b]{0.3\textwidth}
		
		\includegraphics[width=\textwidth]{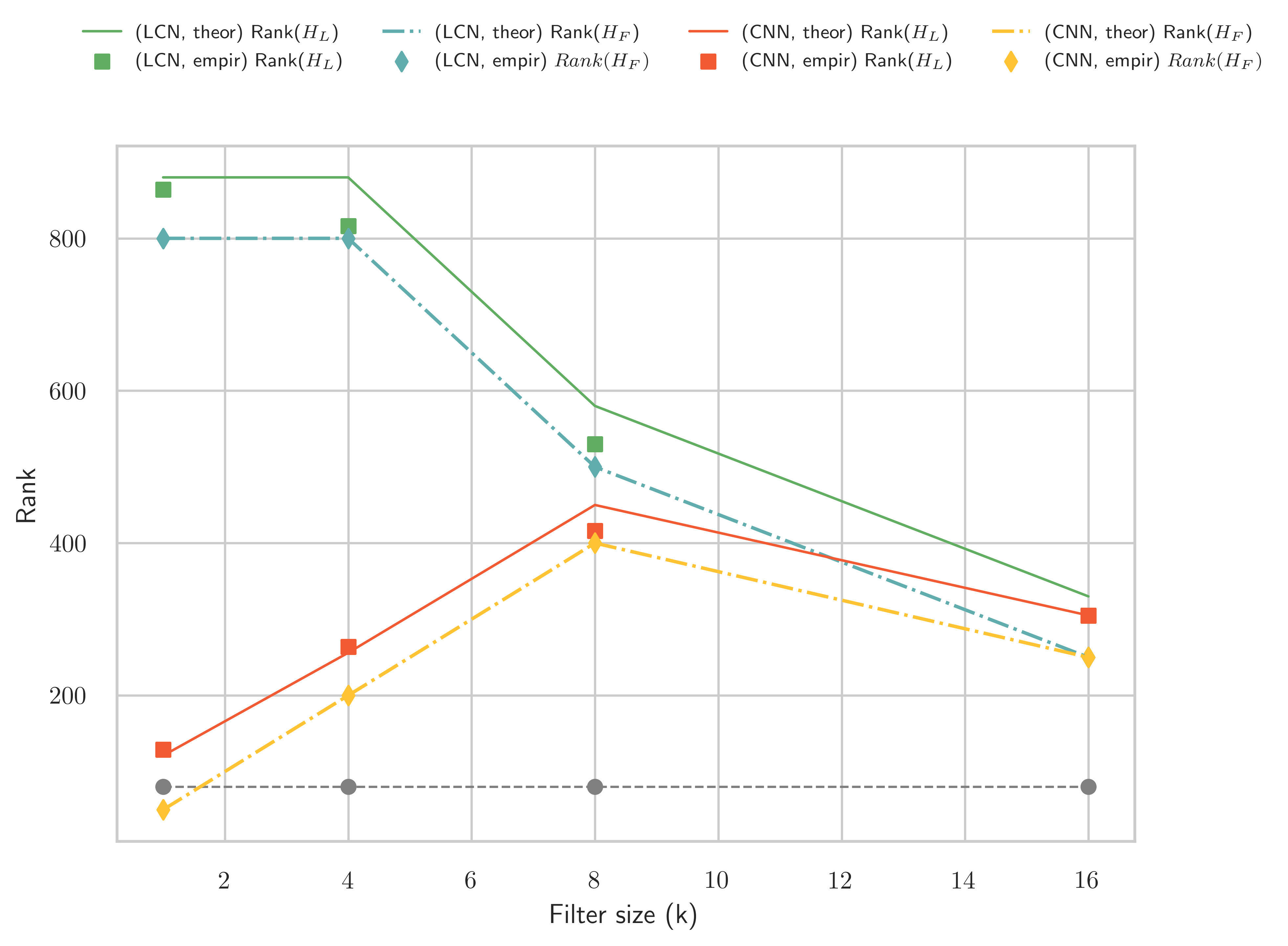}
		\caption{$m=25$}
	\end{subfigure}
	\begin{subfigure}[b]{0.3\textwidth}
		
		\includegraphics[width=\textwidth]{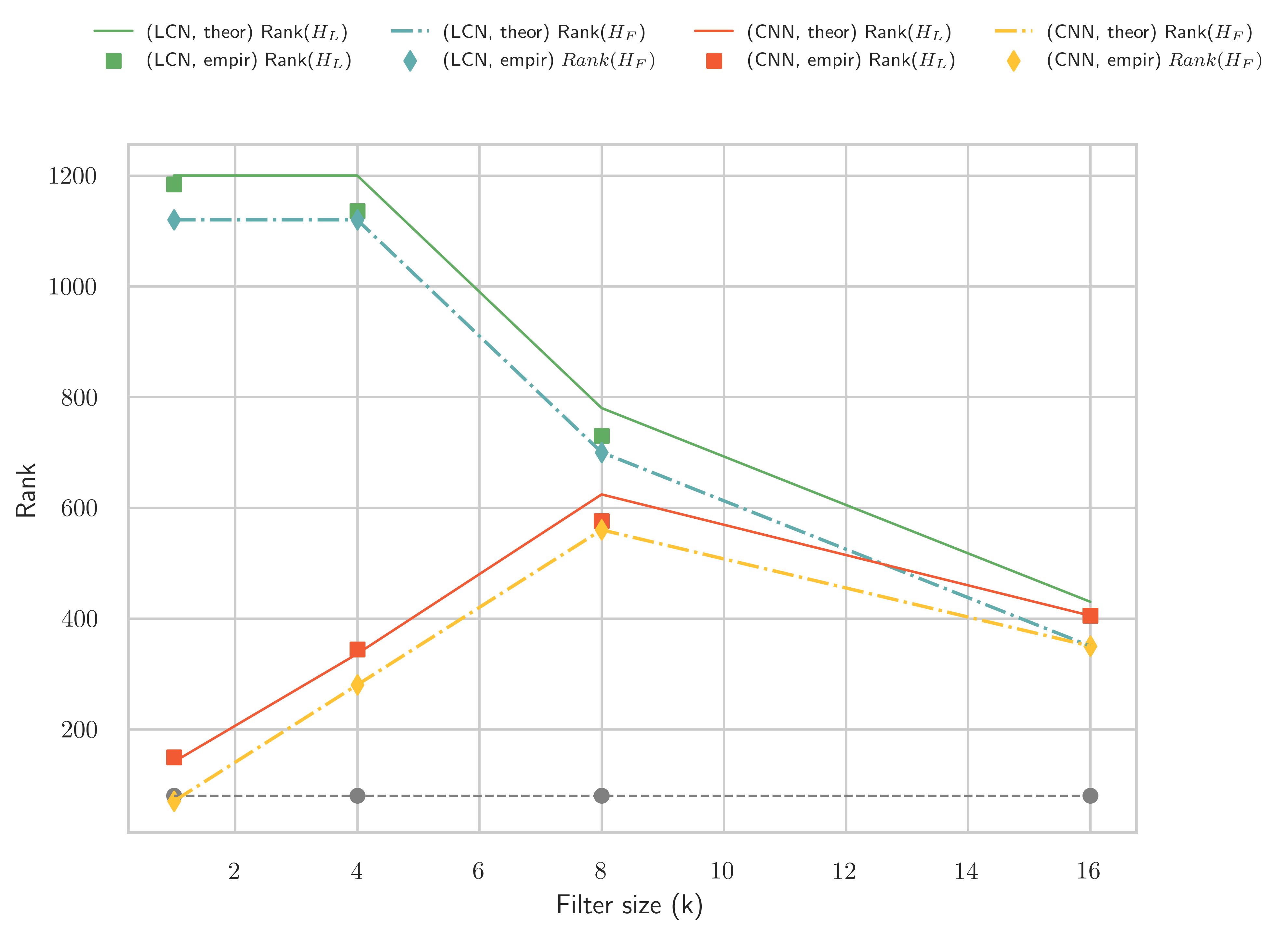}
		\caption{$m=35$}
	\end{subfigure}
	\begin{subfigure}[b]{0.3\textwidth}
		
		\includegraphics[width=\textwidth]{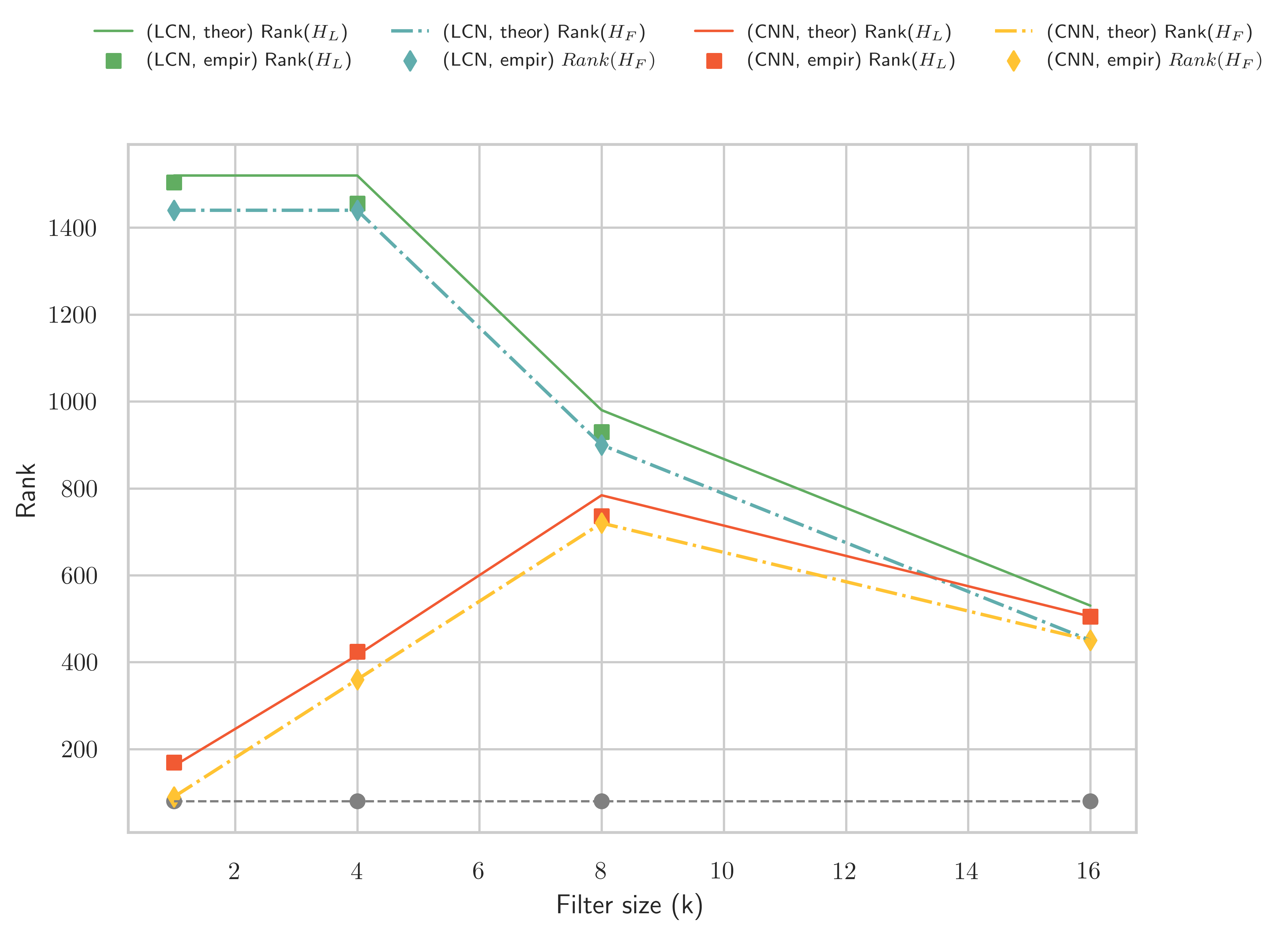}
		\caption{$m=45$}
	\end{subfigure}
	\begin{subfigure}[b]{0.3\textwidth}
		
		\includegraphics[width=\textwidth]{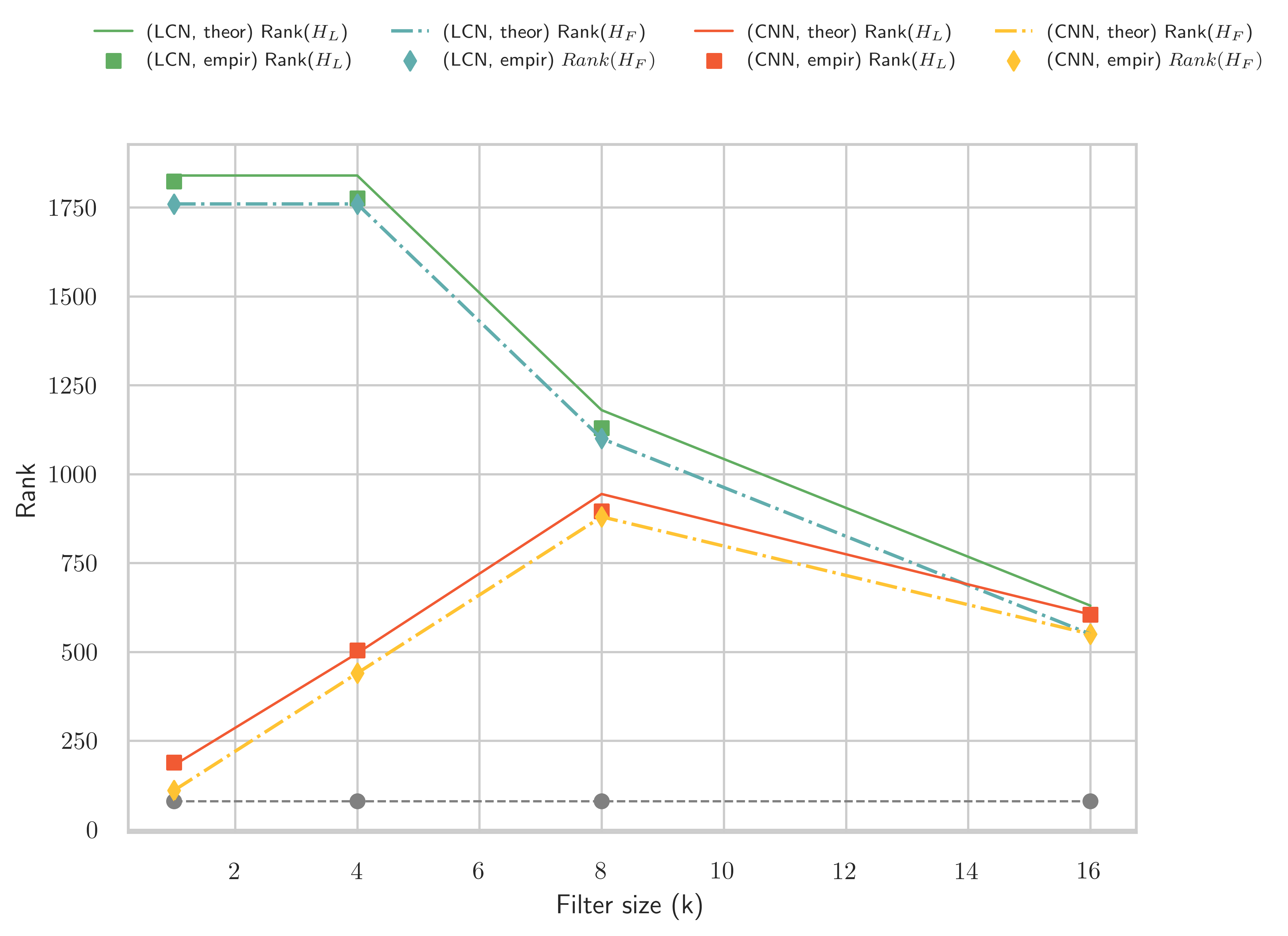}
		\caption{$m=55$}
	\end{subfigure}
	\begin{subfigure}[b]{0.3\textwidth}
		
		\includegraphics[width=\textwidth]{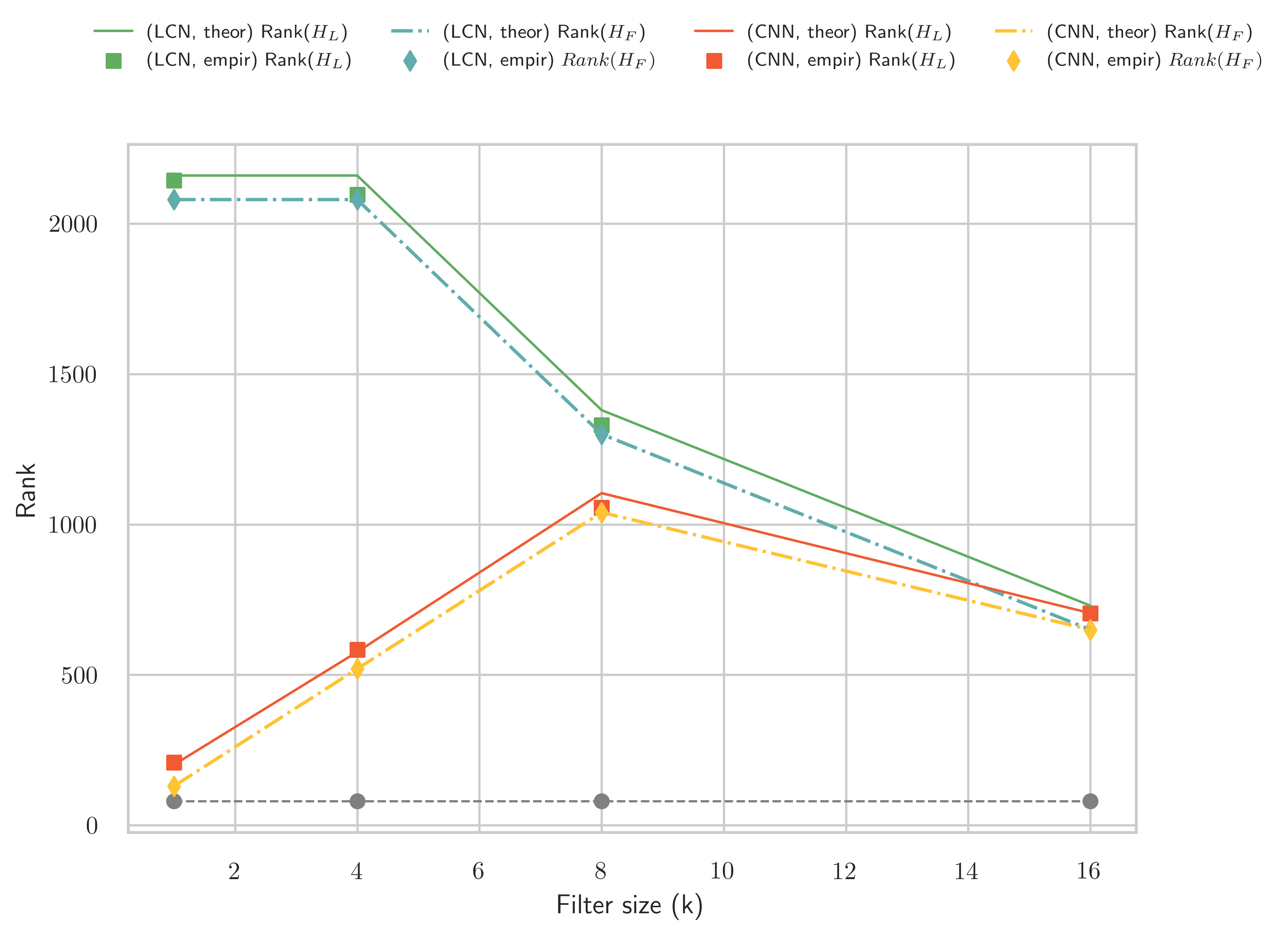}
		\caption{$m=65$}
	\end{subfigure}
	\begin{subfigure}[b]{0.3\textwidth}
		
		\includegraphics[width=\textwidth]{figures/lcn_cnn/jaxcifar10_mse_linear_m-75_d-16_cleaned/lcn_vs_cnn.png}
		\caption{$m=75$}
	\end{subfigure}
	\begin{subfigure}[b]{0.3\textwidth}
		
		\includegraphics[width=\textwidth]{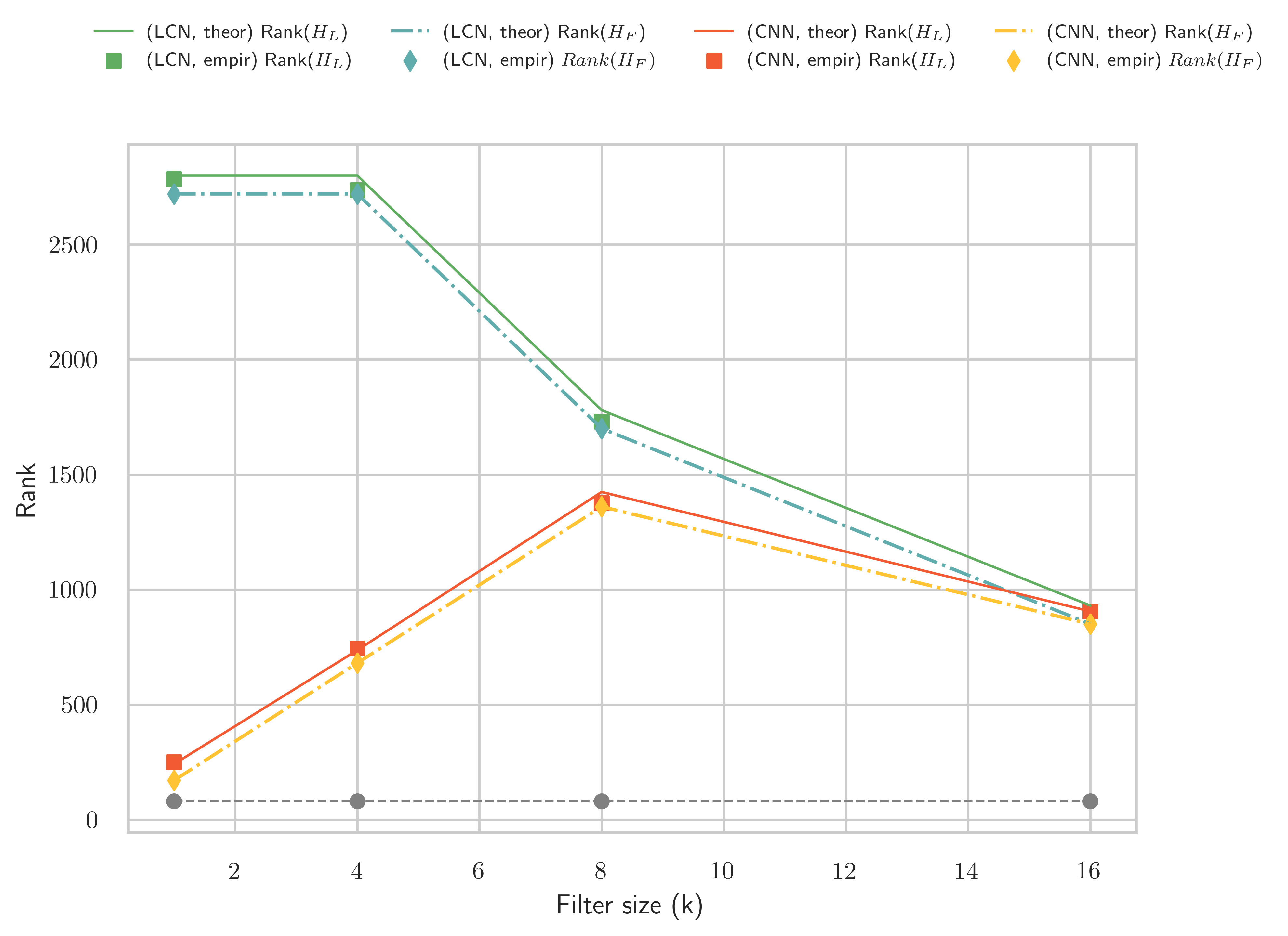}
		\caption{$m=85$}
	\end{subfigure}
%
%
%
%
%
	\caption{Rank vs Filter Size for Linear LCN and LCN + WS on CIFAR10 with Mean Squared Error loss. .}
\end{figure*}
\clearpage
\subsection{ReLU-based LCNs}

\begin{figure*}[!h]
	\centering
	\begin{subfigure}[b]{0.3\textwidth}
		
		\includegraphics[width=\textwidth]{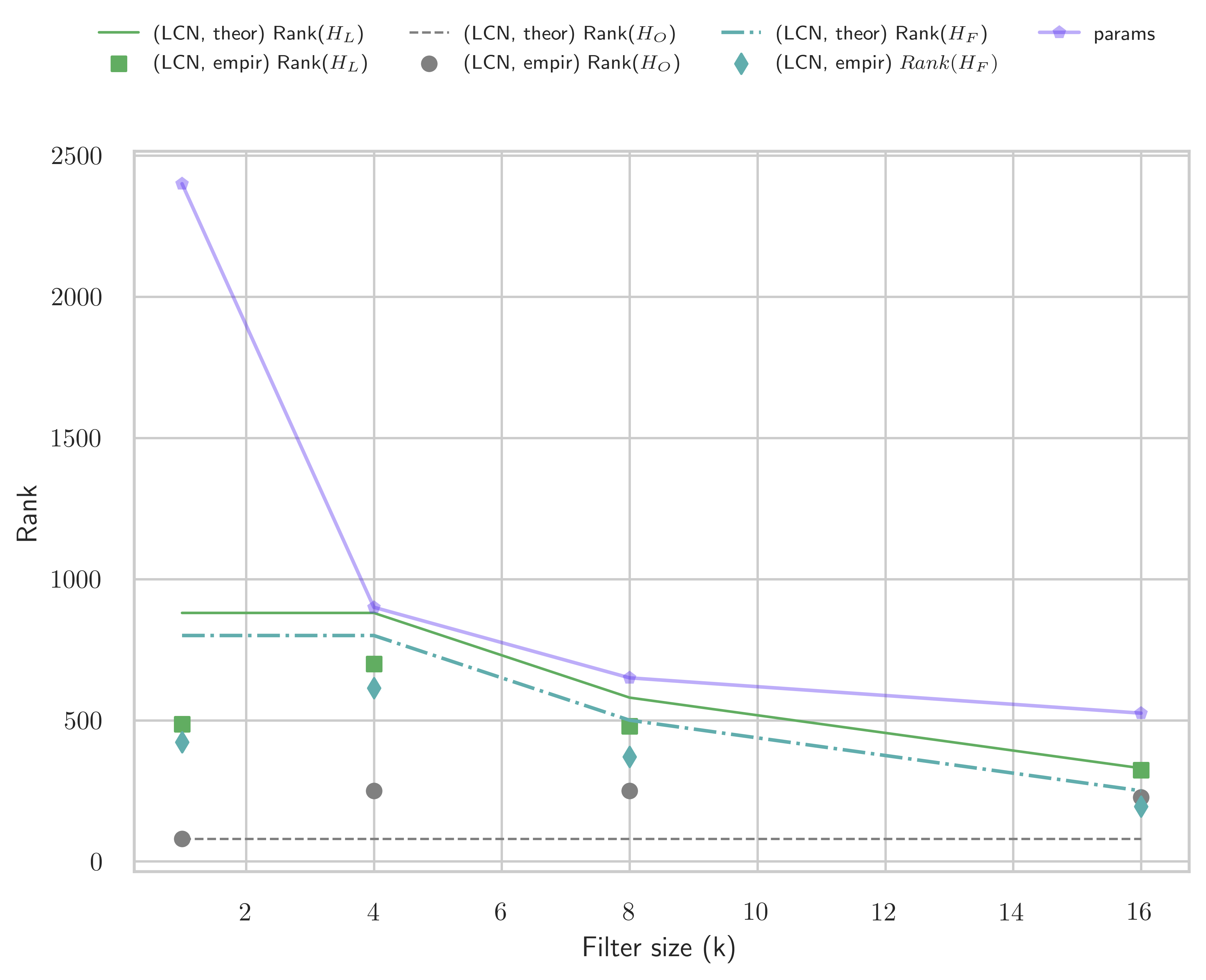}
		\caption{$m=25$}
	\end{subfigure}
	\begin{subfigure}[b]{0.3\textwidth}
		
		\includegraphics[width=\textwidth]{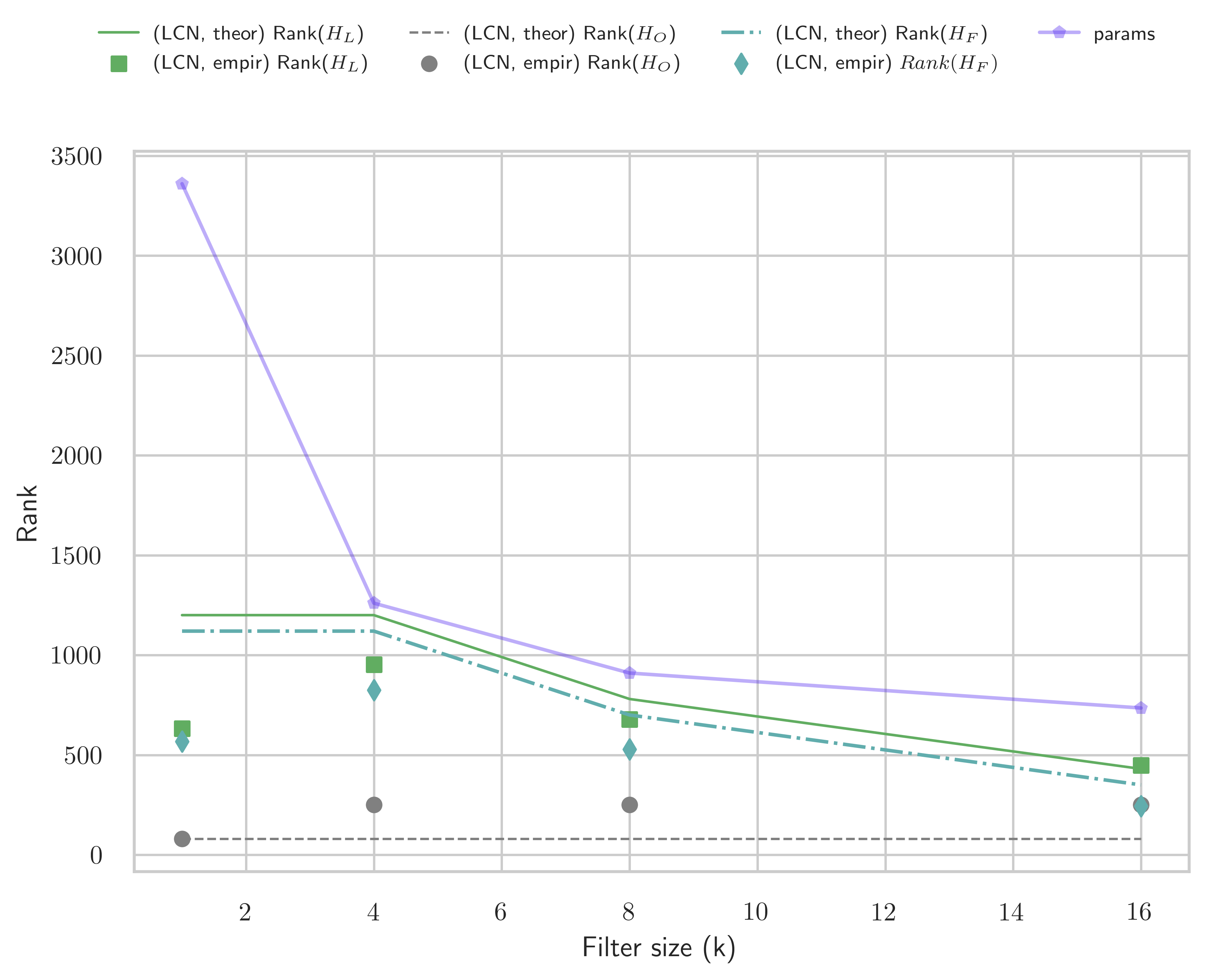}
		\caption{$m=35$}
	\end{subfigure}
	\begin{subfigure}[b]{0.3\textwidth}
		
		\includegraphics[width=\textwidth]{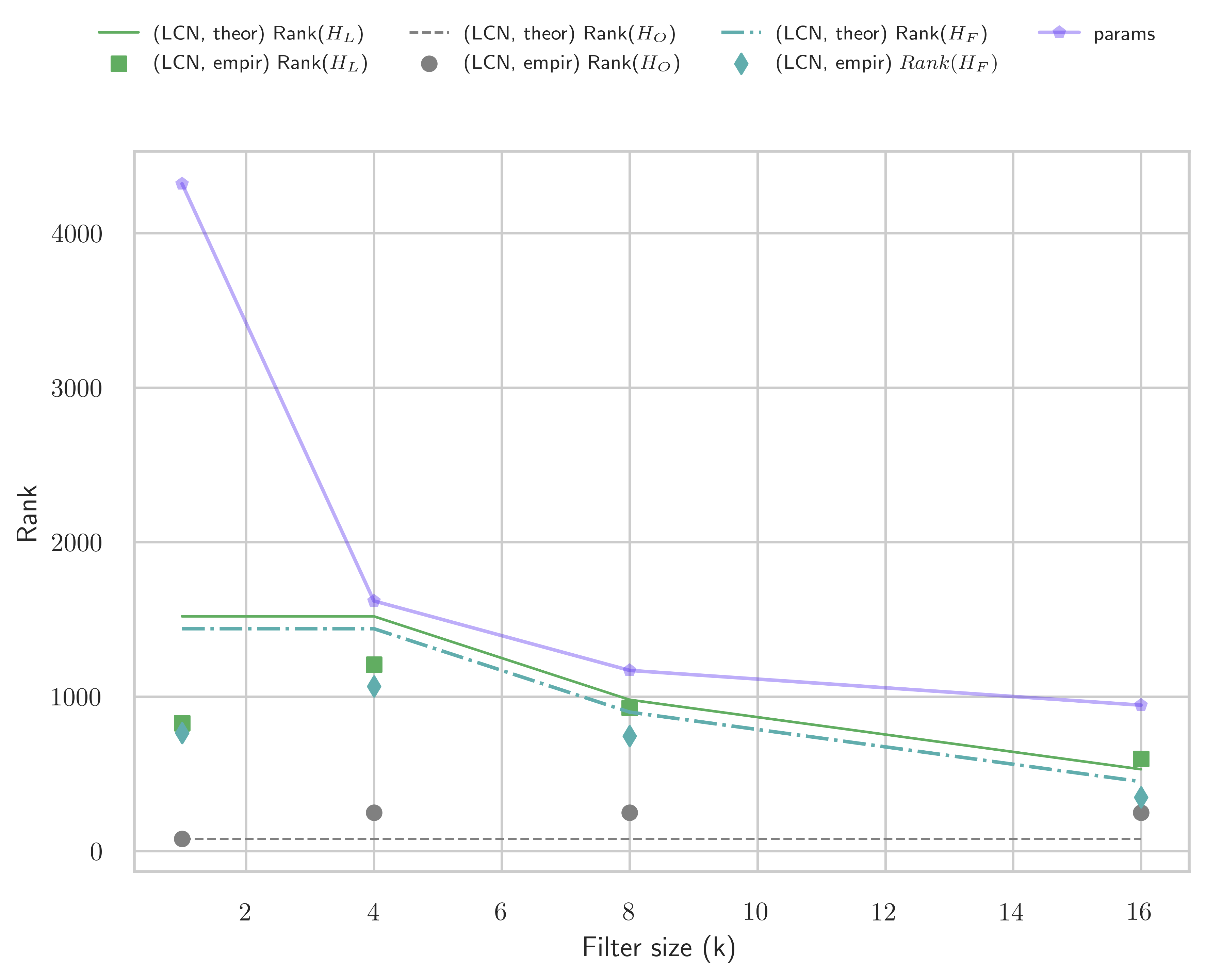}
		\caption{$m=45$}
	\end{subfigure}
	\begin{subfigure}[b]{0.3\textwidth}
		
		\includegraphics[width=\textwidth]{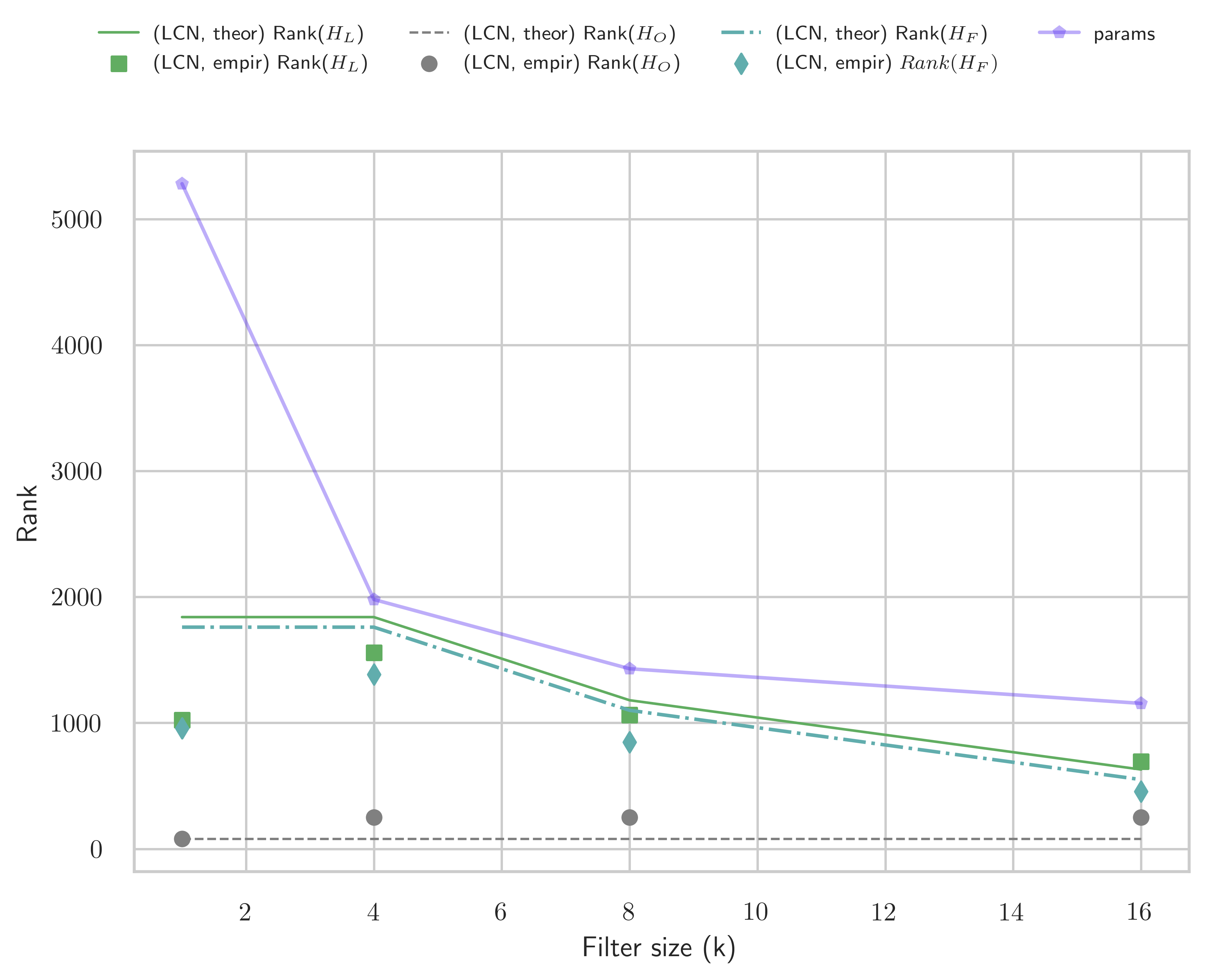}
		\caption{$m=55$}
	\end{subfigure}
	\begin{subfigure}[b]{0.3\textwidth}
		
		\includegraphics[width=\textwidth]{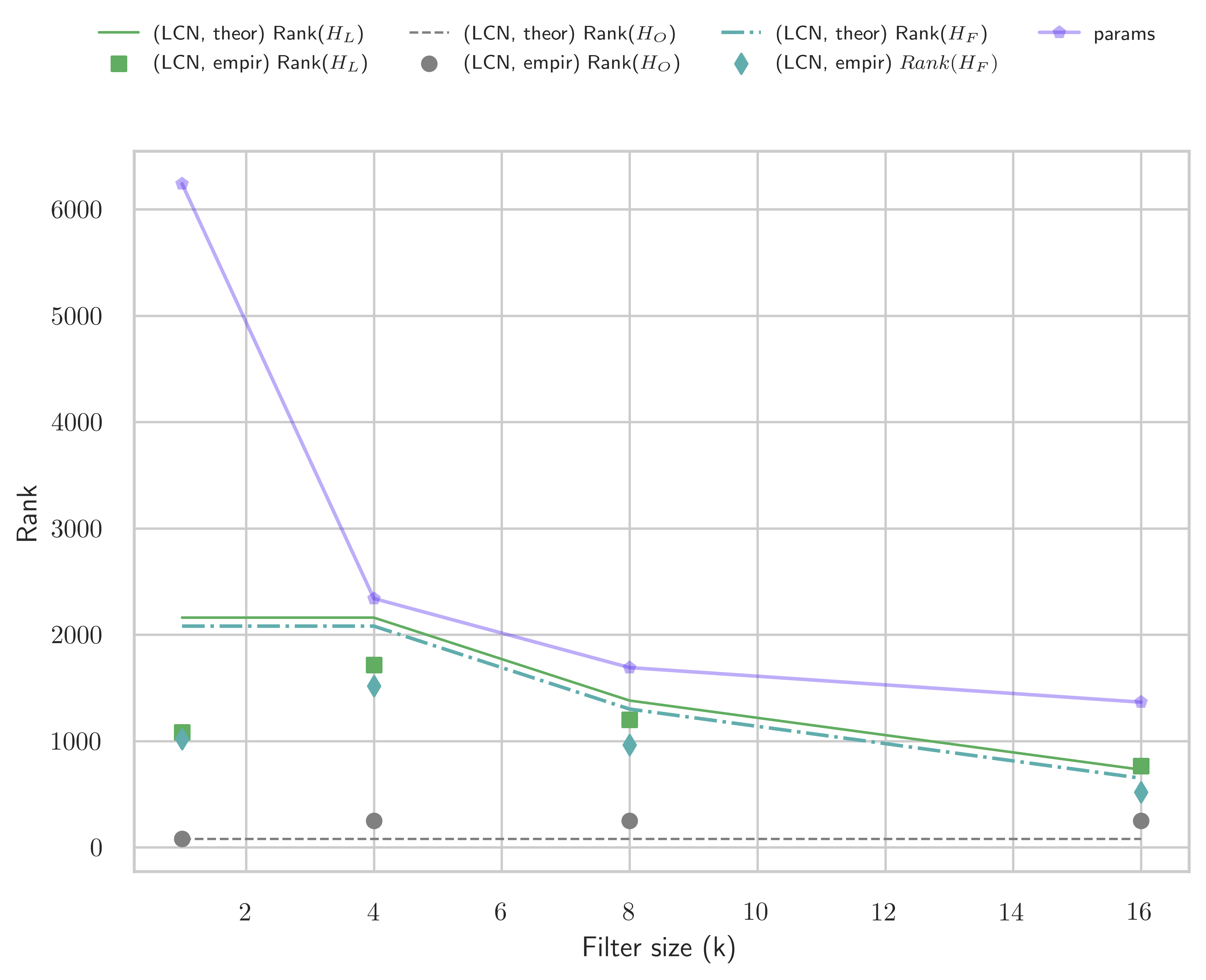}
		\caption{$m=65$}
	\end{subfigure}
	\begin{subfigure}[b]{0.3\textwidth}
		
		\includegraphics[width=\textwidth]{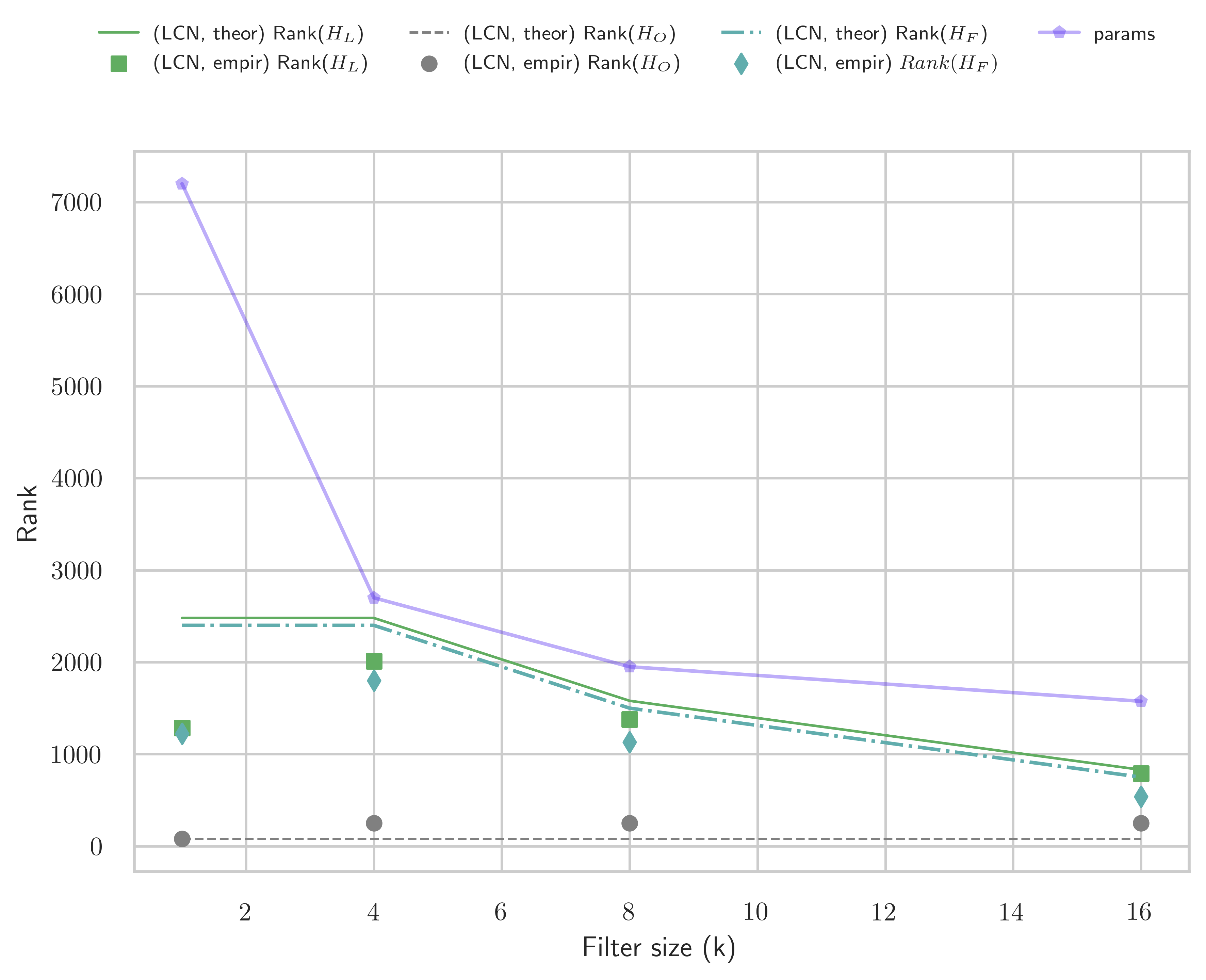}
		\caption{$m=75$}
	\end{subfigure}
	\begin{subfigure}[b]{0.3\textwidth}
		
		\includegraphics[width=\textwidth]{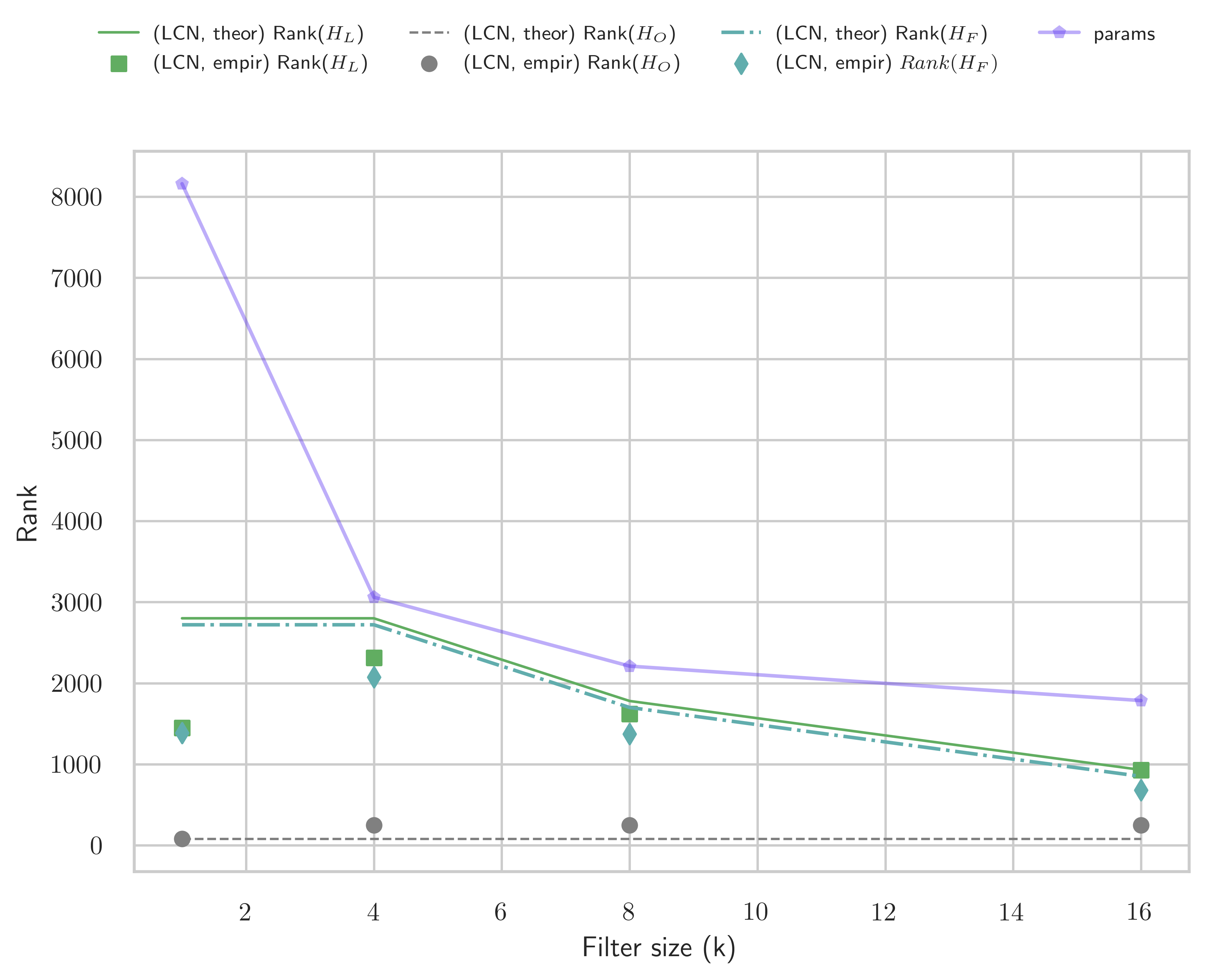}
		\caption{$m=85$}
	\end{subfigure}
	%
	%
	%
	%
	%
	\caption{Rank vs Filter Size for ReLU LCNs on CIFAR10 with Mean Squared Error loss. .}
\end{figure*}
\clearpage
\subsection{ReLU-based LCNs + Weight Sharing}

\begin{figure*}[!h]
	\centering
	\begin{subfigure}[b]{0.3\textwidth}
		
		\includegraphics[width=\textwidth]{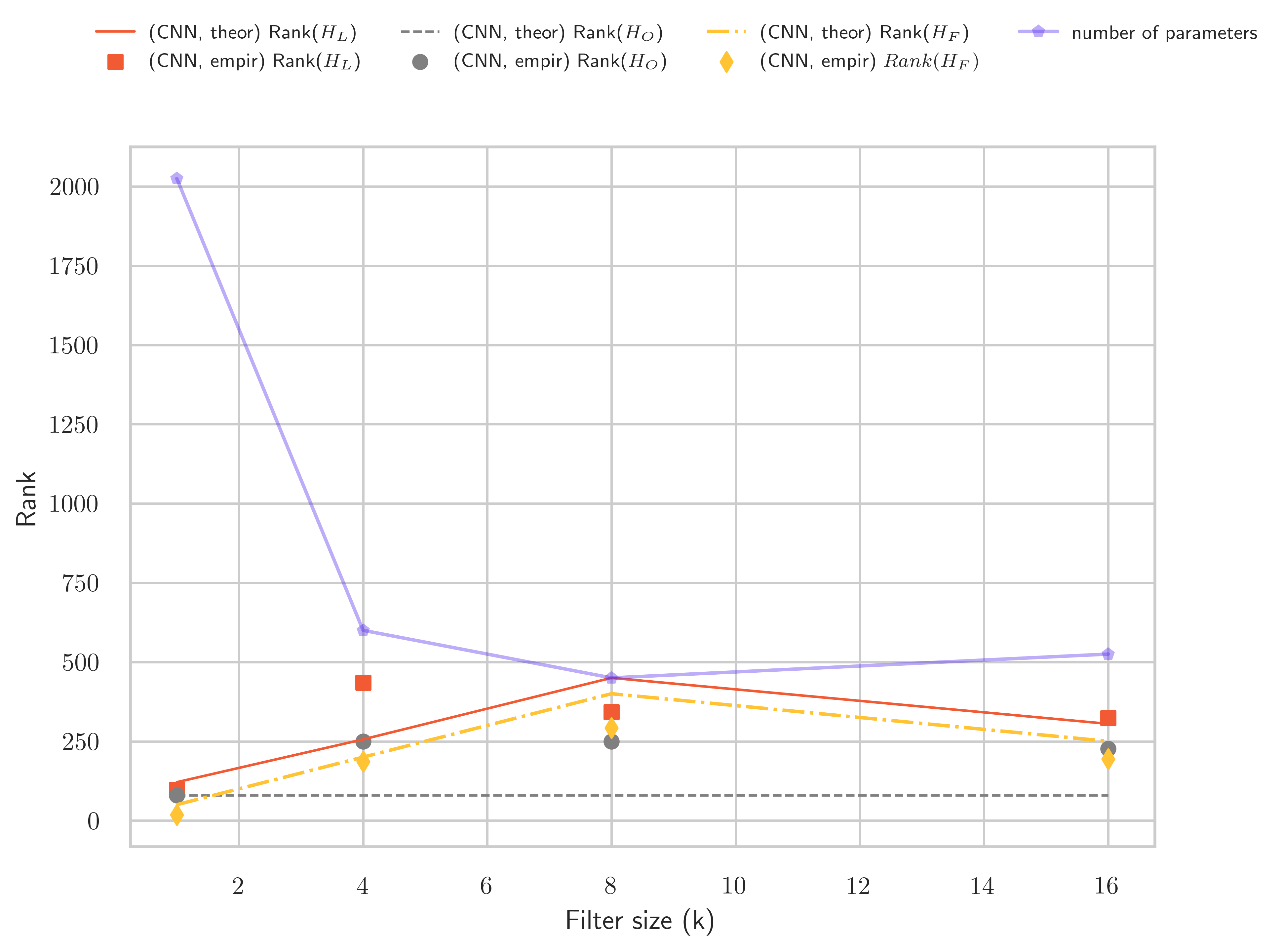}
		\caption{$m=25$}
	\end{subfigure}
	\begin{subfigure}[b]{0.3\textwidth}
		
		\includegraphics[width=\textwidth]{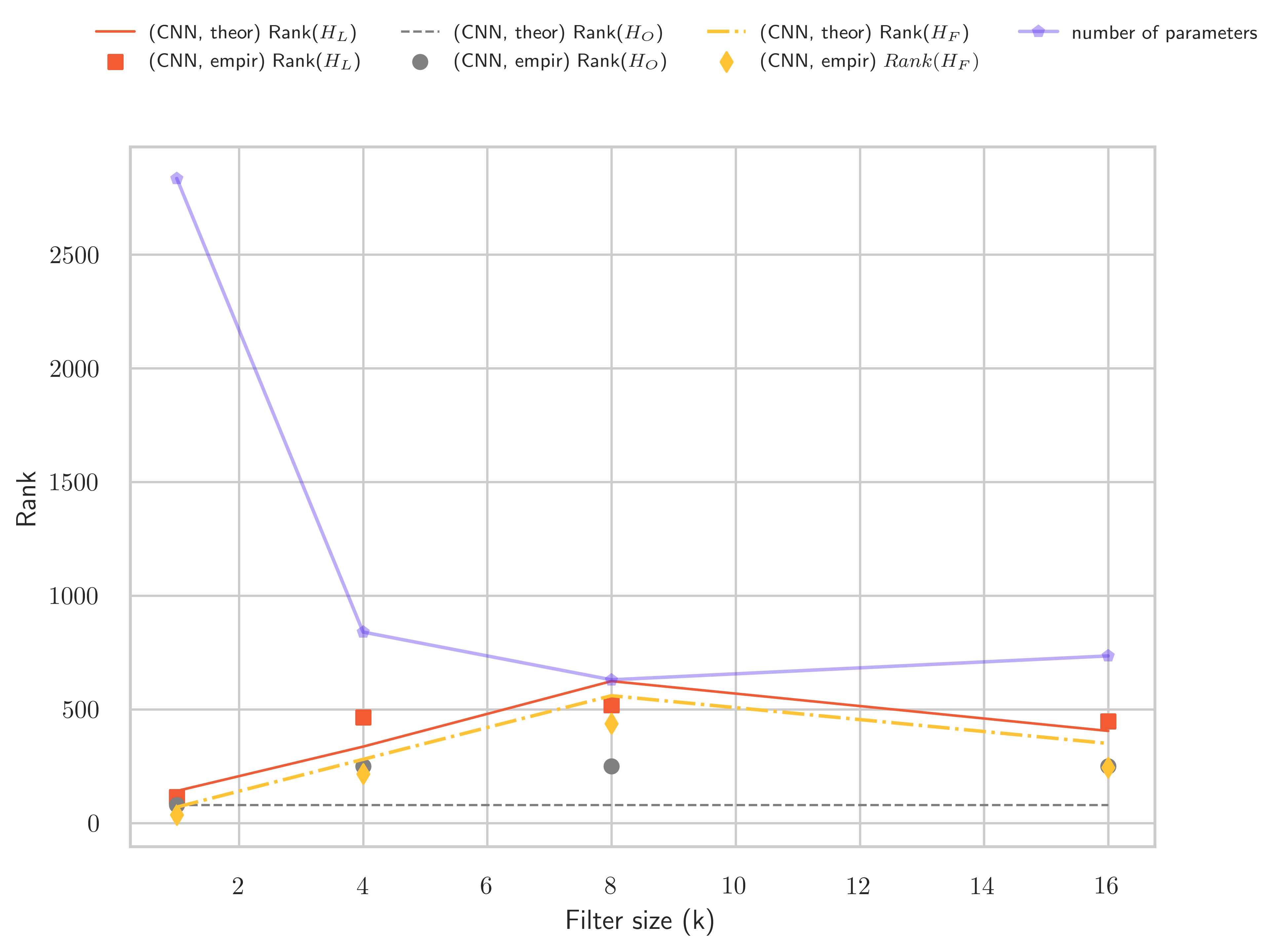}
		\caption{$m=35$}
	\end{subfigure}
	\begin{subfigure}[b]{0.3\textwidth}
		
		\includegraphics[width=\textwidth]{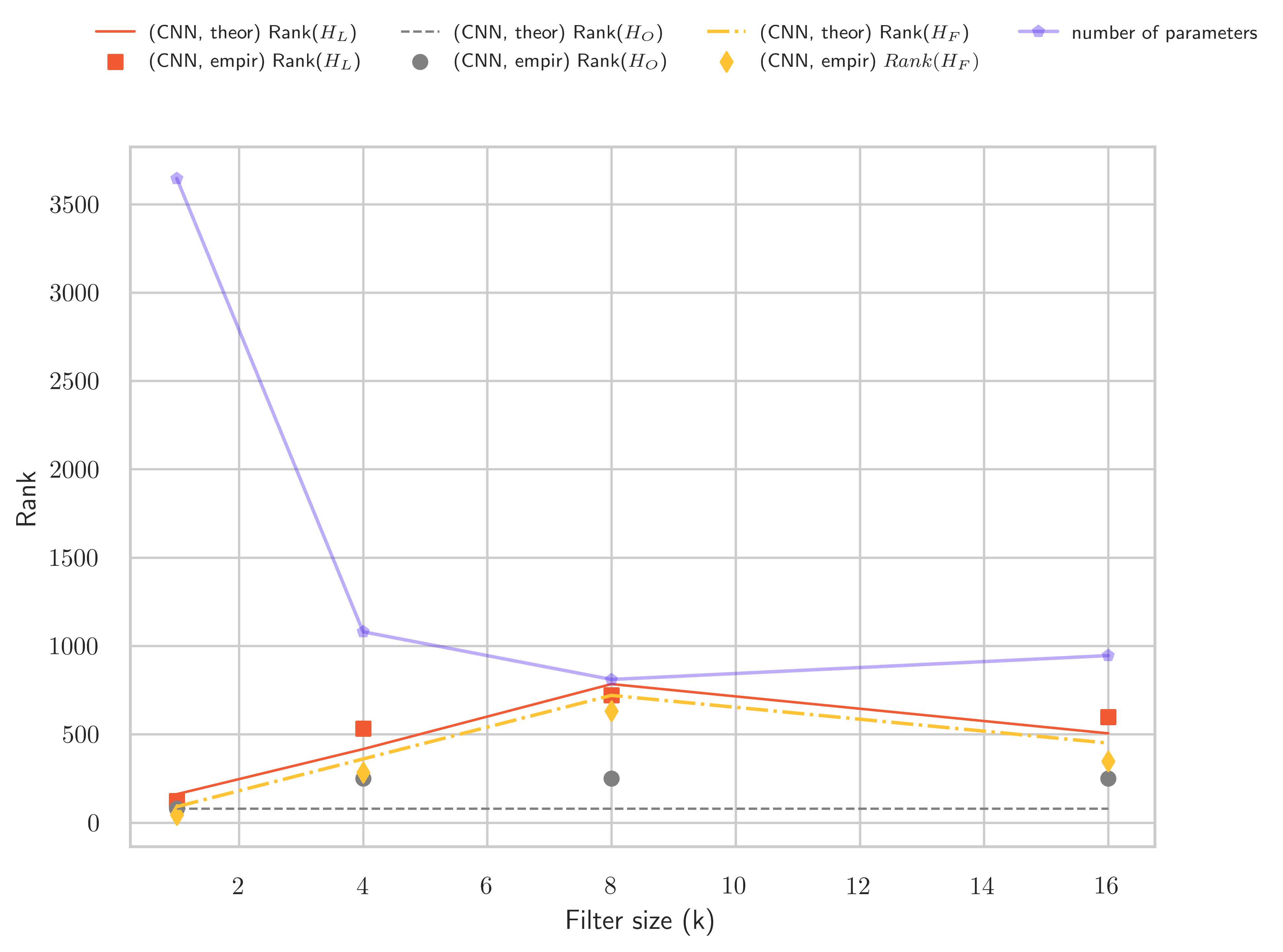}
		\caption{$m=45$}
	\end{subfigure}
	\begin{subfigure}[b]{0.3\textwidth}
		
		\includegraphics[width=\textwidth]{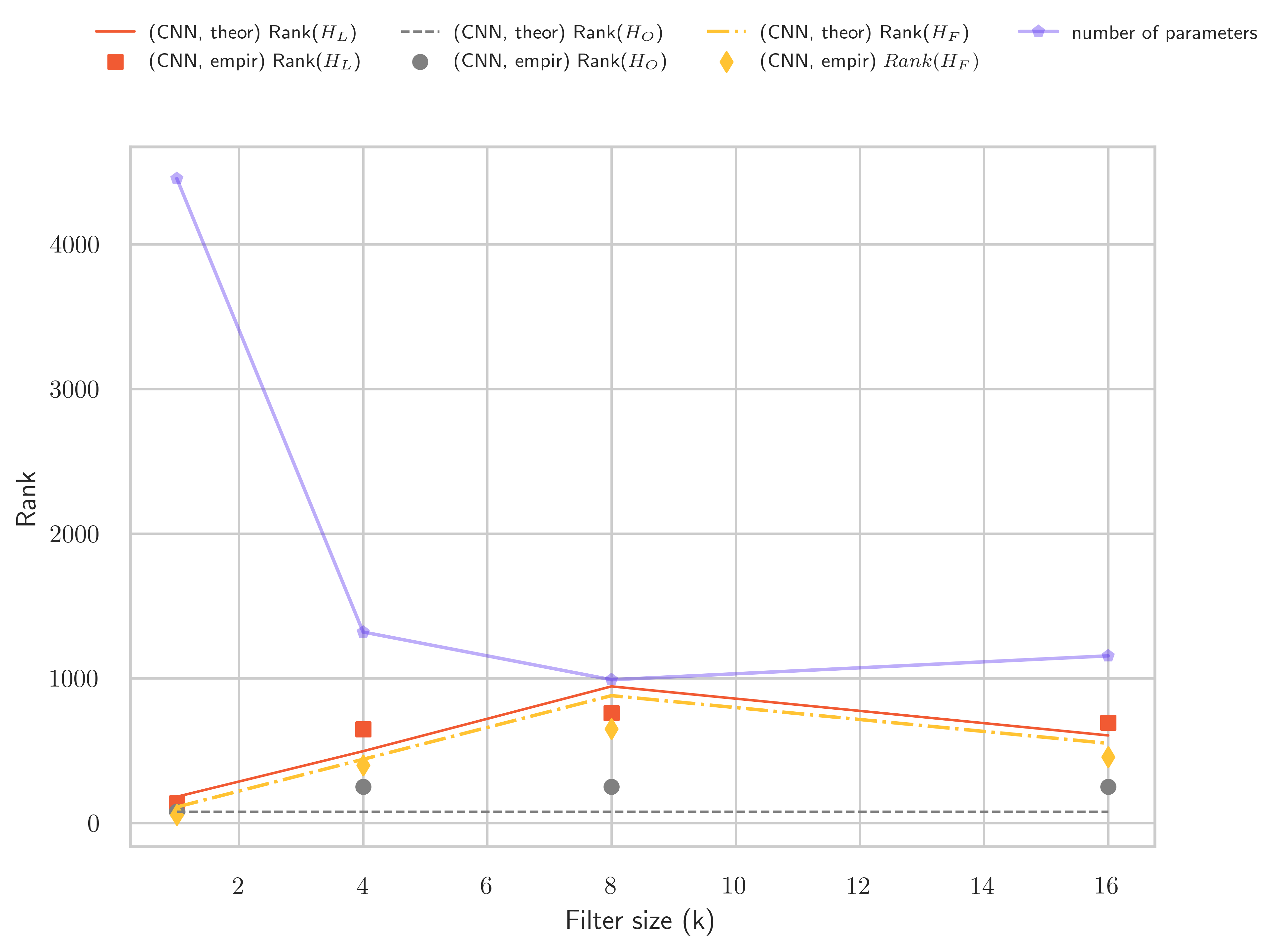}
		\caption{$m=55$}
	\end{subfigure}
	\begin{subfigure}[b]{0.3\textwidth}
		
		\includegraphics[width=\textwidth]{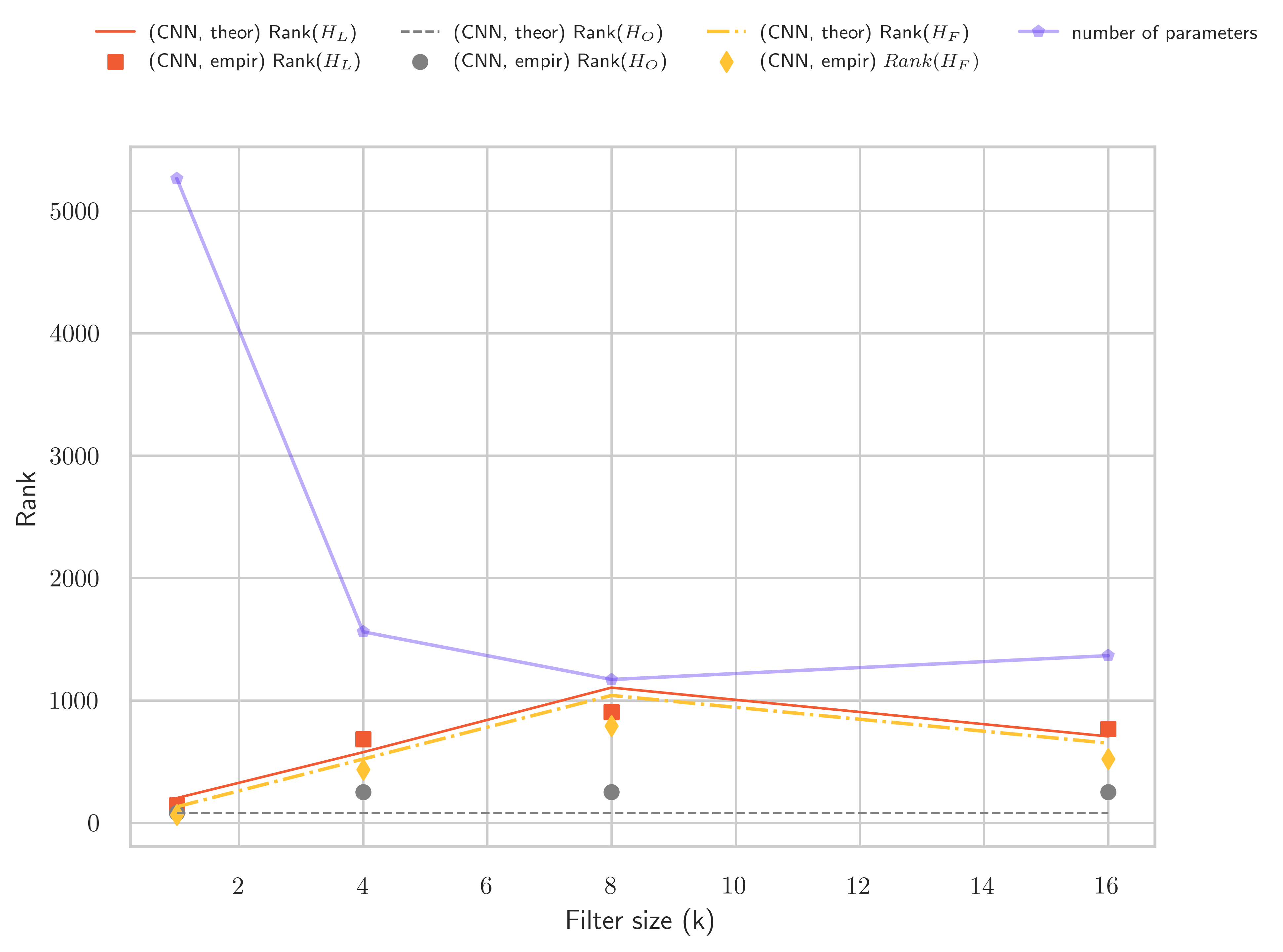}
		\caption{$m=65$}
	\end{subfigure}
	\begin{subfigure}[b]{0.3\textwidth}
		
		\includegraphics[width=\textwidth]{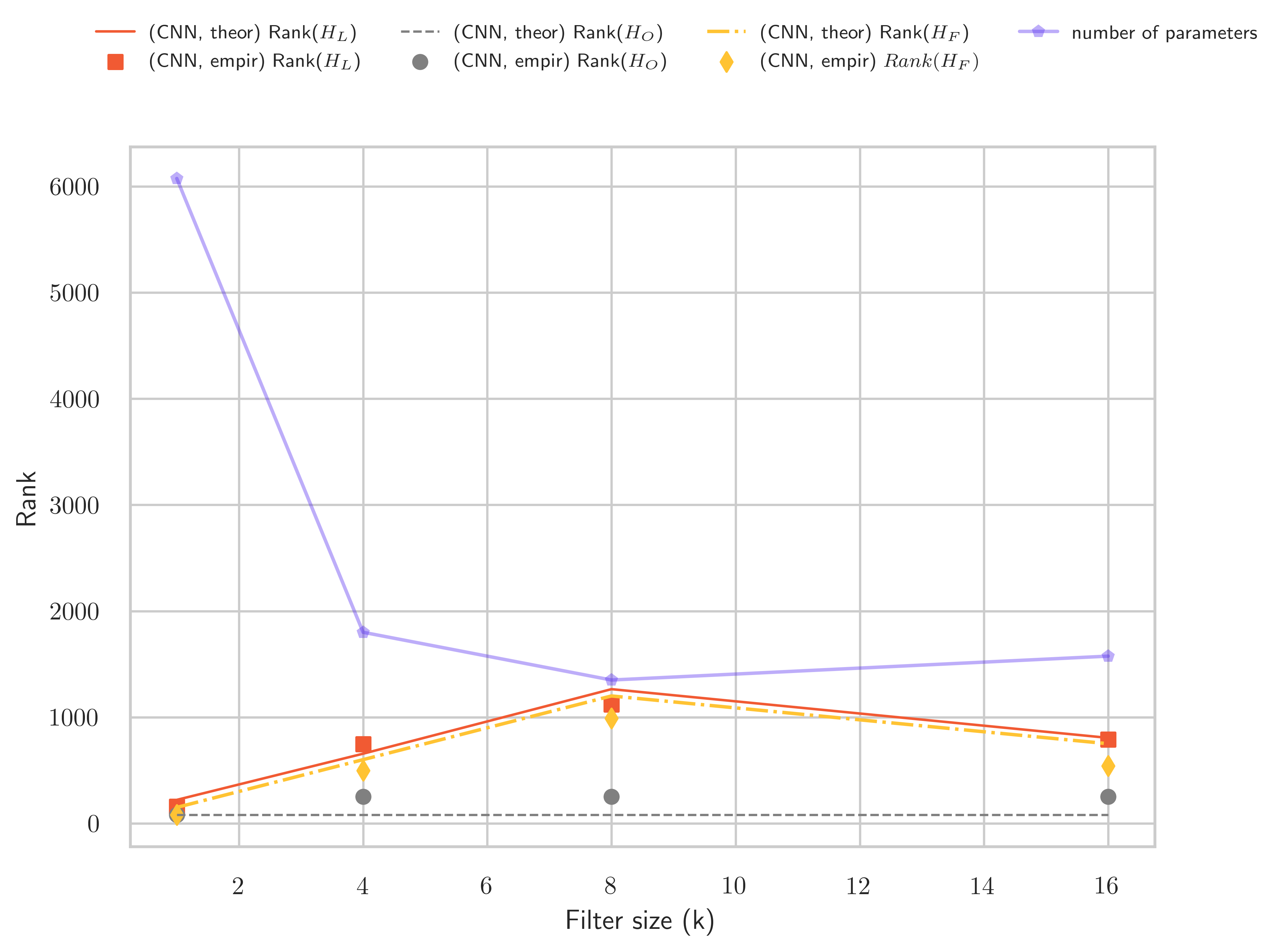}
		\caption{$m=75$}
	\end{subfigure}
	\begin{subfigure}[b]{0.3\textwidth}
		
		\includegraphics[width=\textwidth]{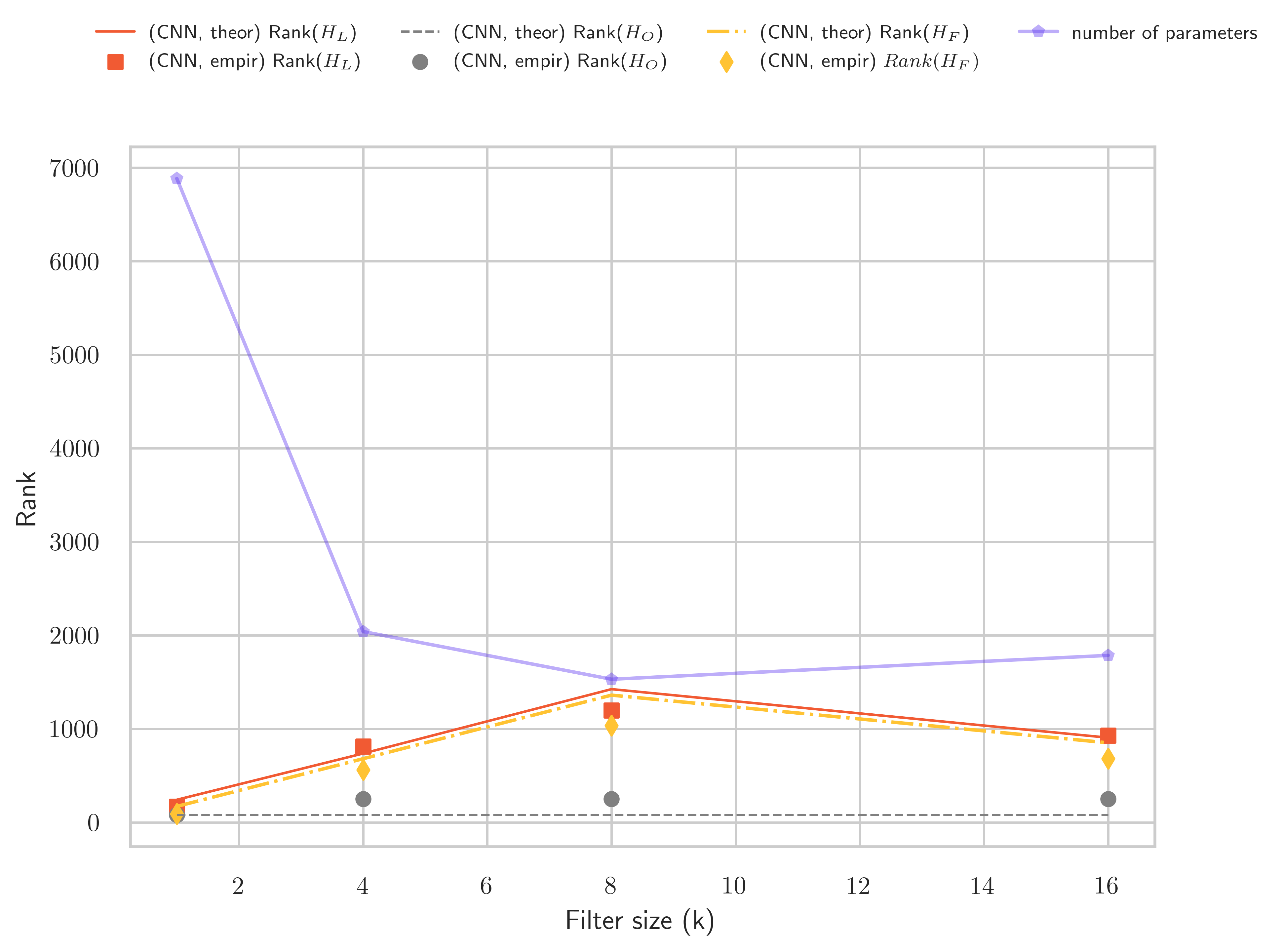}
		\caption{$m=85$}
	\end{subfigure}
	%
	%
	%
	%
	%
	\caption{Rank vs Filter Size for ReLU-based LCN + WS on CIFAR10 with Mean Squared Error loss. .}
\end{figure*}
\subsection{Correlation of Rank with Generalization Performance}

\begin{figure}[!h]
    \centering
    
    \includegraphics[width=0.45\textwidth]{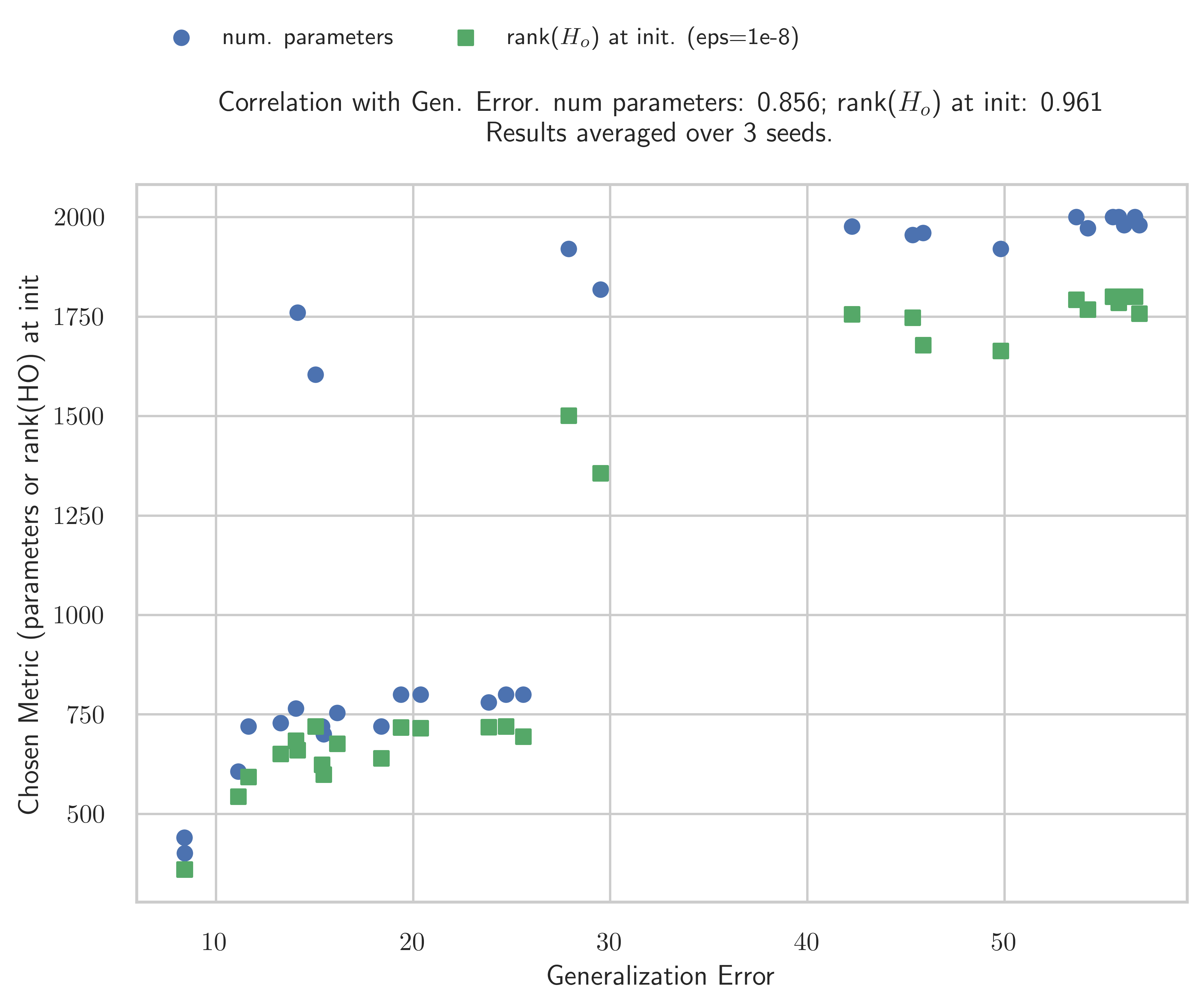}
    %
    \caption{Correlation of Hessian rank at initialization with generalization error for a sweep over filter sizes and number of channels in a one-hidden layer CNN}
    \label{fig:gen}
\end{figure}

\end{document}